\documentclass[10pt,twocolumn,letterpaper]{article}

\usepackage{iccv}
\usepackage{times}
\usepackage{epsfig}
\usepackage{graphicx}
\usepackage{amsmath}
\usepackage{amssymb}
\usepackage{makecell}
\usepackage{multirow}
\usepackage{subcaption}
\usepackage{pbox}

\graphicspath{ {figures/} }

\usepackage[pagebackref=true,breaklinks=true,letterpaper=true,colorlinks,bookmarks=false]{hyperref}

\iccvfinalcopy 

\def\iccvPaperID{946} 

\ificcvfinal\fi
\usepackage[skip=2pt]{caption}
\begin{document}

\title{Shape Inpainting using 3D Generative Adversarial Network and Recurrent Convolutional Networks}

\author{Weiyue Wang$^1$
 \qquad Qiangui Huang$^1$ \qquad Suya You$^2$ \qquad Chao Yang$^1$ \qquad Ulrich Neumann$^1$\\
$^1$University of Southern California \hspace{20mm} $^2$US Army Research Laboratory\\
\hspace{5mm}Los Angeles, California \hspace{35mm} Playa Vista, California\\
\hspace{-15mm}{\tt\small \{weiyuewa,qianguih,chaoy,uneumann\}@usc.edu}\hspace{20mm}{\tt\small suya.you.civ@mail.mil}\qquad
}

\maketitle

\begin{abstract}
Recent advances in convolutional neural networks have shown promising results in 3D shape completion. But due to GPU memory limitations, these methods can only produce low-resolution outputs. To inpaint 3D models with semantic plausibility and contextual details, we introduce a hybrid framework that combines a 3D Encoder-Decoder Generative Adversarial Network (3D-ED-GAN) and a Long-term Recurrent Convolutional Network (LRCN). The 3D-ED-GAN is a 3D convolutional neural network trained with a generative adversarial paradigm to fill missing 3D data in low-resolution. LRCN adopts a recurrent neural network architecture to minimize GPU memory usage and incorporates an Encoder-Decoder pair into a Long Short-term Memory Network. By handling the 3D model as a sequence of 2D slices, LRCN transforms a coarse 3D shape into a more complete and higher resolution volume. While 3D-ED-GAN captures global contextual structure of the 3D shape, LRCN localizes the fine-grained details. Experimental results on both real-world and synthetic data show reconstructions from corrupted models result in complete and high-resolution 3D objects.\let\thefootnote\relax\footnote{We thank Alex Tong Lin for discussions and proofreading.}
\end{abstract}

\section{Introduction}

\begin{figure*}[t]
    \includegraphics[width=\textwidth]{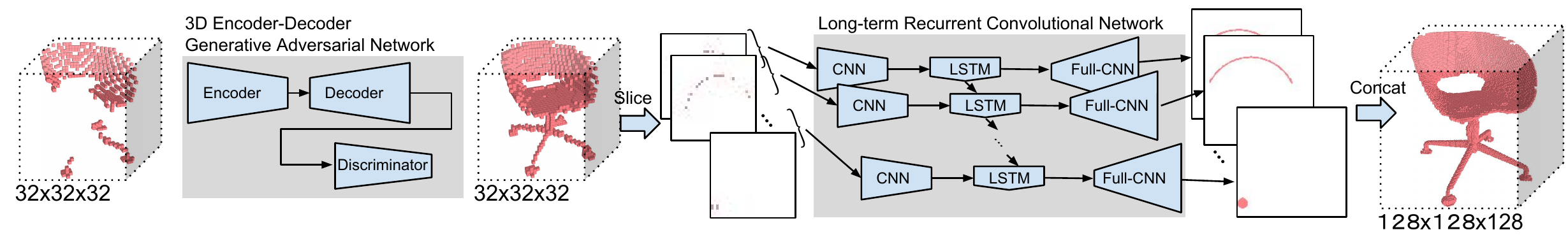}
    \caption{Our method completes a corrupted 3D scan using a convolutional Encoder-Decoder generative adversarial network in low resolution. The outputs are then sliced into a sequence of 2D images and a recurrent convolutional network is further introduced to produce high-resolution completion prediction.   }
    \label{fig:framework}
\end{figure*}

Data collected by 3D sensors (e.g. LiDAR, Kinect) are often impacted by occlusion, sensor noise, and illumination, leading to incomplete and noisy 3D models. For example, a building scan occluded by a tree leads to a hole or gap in the 3D building model. However, a human can comprehend and describe the geometry of the complete building based on the corrupted 3D model. Our 3D inpainting method attempts to mimic this ability to reconstruct complete 3D models from incomplete data.

Convolutional Neural Network (CNN) based methods~\cite{gan,dcgan,contextencoder,neuralpatch} yield impressive results for 2D image generation and image inpainting. Generating and inpainting 3D models is a new and more challenging problem due to its higher dimensionality. The availability of large 3D CAD datasets~\cite{shapenet,modelnet} and CNNs for voxel (spatial occupancy) models~\cite{3dgan,vconvdae,3depn} enabled progress in learning 3D representation, shape generation and completion. Despite their encouraging results, artifacts still persists in their generated shapes. Moreover, their methods are all based on 3D CNN, which impedes their ability to handle higher resolution data due to limited GPU memory.

In this paper, a new system for 3D object inpainting is introduced to overcome the aforementioned limitations. Given a 3D object with holes, we aim to (1) fill the missing or damaged portions and reconstruct a complete 3D structure, and (2) further predict high-resolution shapes with fine-grained details. We propose a hybrid network structure based on 3D CNN that leverages the generalization power of a Generative Adversarial model and the memory efficiency of Recurrent Neural Network (RNN) to handle 3D data sequentially. The framework is illustrated in Figure~\ref{fig:framework}.

More specifically, a 3D Encoder-Decoder Generative Adversarial Network (3D-ED-GAN) is firstly proposed to generalize geometric structures and map corrupted scans to complete shapes in low resolution. Like a variational autoencoder (VAE)~\cite{VAE,vconvdae}, 3D-ED-GAN utilizes an encoder to map voxelized 3D objects into a probabilistic latent space, and a Generative Adversarial Network (GAN) to help the decoder predict the complete volumetric objects from the latent feature representation. We train this network by minimizing both contextual loss and an adversarial loss. Using GAN, we can not only preserve contextual consistency of the input data, but also inherit information from data distribution.

Secondly, a Long-term Recurrent Convolutional Network (LRCN) is further introduced to obtain local geometric details and produce much higher resolution volumes. 3D CNN requires much more GPU memory than 2D CNN, which impedes volumetric network analysis of high-resolution 3D data. To overcome this limitation, we model the 3D objects as sequences of 2D slices. By utilizing the long-range learning capability from a series of conditional distributions of RNN, our LRCN is a Long Short-term Memory Network (LSTM) where each cell has a CNN encoder and a fully-convolutional decoder. The outputs of 3D-ED-GAN are sliced into 2D images, which are then fed into the LRCN, which gives us a sequence of high-resolution images.

Our hybrid network is an end-to-end trainable network which takes corrupted low resolution 3D structures and outputs complete and high-resolution volumes. We evaluate the proposed method qualitatively and quantitatively on both synthesized and real 3D scans in challenging scenarios. To further evaluate the ability of our model to capture shape features during 3D inpainting, we test our network for 3D object classification tasks and further explore the encoded latent vector to demonstrate that this embedded representation contains abundant semantic shape information. 

The main contributions of this paper are:
\begin{enumerate}
\item a 3D Encoder-Decoder Generative Adversarial Convolutional Neural Network that inpaints holes in 3D models, which can further help 3D shape feature learning and help object recognition.
\item a Long-term Recurrent Convolutional Network that produces high resolution 3D volumes with fine-grained details by modeling volumetric data as sequences of 2D images to overcome GPU memory limitation.
\item an end-to-end network that combines the above two ideas and completes corrupted 3D models, while also producing high resolution volumes. 
\end{enumerate}

\section{Related Work}

%

\subsection{Generative models}
Generative Adversarial Network (GAN)~\cite{gan} generates images by jointly training a generator and a discriminator. Following this pioneering work, a series of GAN models~\cite{dcgan,lapgan} were developed for image generation tasks. Pathak et al.~\cite{contextencoder} developed a context encoder in an unsupervised learning algorithm for image inpainting. Generative adversarial loss in their autoencoder-like network architecture achieves impressive performance for image inpainting.

With the introduction of 3D CAD model datasets~\cite{modelnet,shapenet}, recent developments in 3D generative models use data-driven methods to synthesize new objects. CNN is used to learn embedded object representations. Bansal et al.~\cite{marr} introduced a skip-network model to retrieve 3D models for objects depicted in 2D images of CAD data. Choy et al.~\cite{r2n2} used a recurrent network with multi-view images for 3D model reconstruction. Girdhar~\cite{tl} proposed a TL-embedding network to learn an embedding space that can be generative in 3D and predicative from 2D rendered images. Wu et al.~\cite{3dgan} showed that the learned latent vector by 3D GAN can generate high-quality 3D objects and improve object recognition accuracy as a shape descriptor. They also added an image encoder to 3D GAN to generate 3D model from 2D images. Yan et al.~\cite{honglak} formulated an encoder-decoder network with a loss by perspective transformation for predicting 3D models from a single-view 2D image.
\subsection{3D Completion}

Recent advances in deep learning have shown promising results in 3D completion. Wu et al.~\cite{modelnet} built a generative model with Convolutional Deep Belief Network by learning a probabilistic distribution from 3D volumes for shape completion from 2.5D depth maps. Sharma~\cite{vconvdae} introduced a fully convolutional autoencoder that learns volumetric representation from noisy data by estimating voxel occupancy grids. This is the state of the art for 3D volumetric occupancy grid inpainting to the best of our knowledge. An important benefit of our 3D-ED-GAN over theirs is that we introduce GAN to inherit information from the data distribution. Dai et al.~\cite{3depn} introduced a 3D-Encoder-Predictor Network to predict and fill missing data for 3D distance field and proposed a 3D synthesis procedure to obtain high-resolution objects. This is the state-of-the-art method for high-resolution object completion. However, instead of an end-to-end network, their shape synthesis procedure requires iterating every sample from the dataset. Since we are using occupancy grids to represent 3D shapes, we do not compare with them in our experiment. Song et al.~\cite{ssn} synthesized a 3D scenes dataset and proposed a semantic scene completion network to produce complete 3D volumes and semantic labels for a scene from single-view depth map.  Despite the encouraging results of the works mentioned above, these methods are mostly based on 3D CNN, which requires much more GPU memory than 2D convolution and impedes handling high-resolution data. 

\subsection{Recurrent Neural Networks}
RNNs have been shown to excel at hard sequence problems ranging from natural language translation~\cite{recurrenttranslation}, to video analysis~\cite{LRCNvideo}. By implicitly conditioning on all previous variables and preserving long-range contextual dependencies, RNNs are also suitable for dense prediction tasks such as semantic segmentation~\cite{reseg,scenelabelinglstm}, and image completion~\cite{pixelrnn}. Donahue et al.~\cite{LRCNvideo} applied 2D CNN and LSTM on 3D data (video) and developed a recurrent convolutional architecture for video recognition. Oord et al.~\cite{pixelrnn} presented a deep network that sequentially predicts the pixels in an image along two spatial dimensions. Choy et al.~\cite{r2n2} used a recurrent network and a CNN to reconstruct 3D models from a sequence of multi-view images. Followed by these pioneer works, we apply RNN on 3D object data and predict dense volume as sequences of 2D pixels. 
\section{Methods}

The goal of this paper is to take a corrupted 3D object in low resolution as input and produce a complete high-resolution model as output. The 3D model is represented as volumetric occupancy grids. To fill the missing data requires an approach that can make conceivable predictions from data distributions as well as preserve structural context of the imperfect input. 

We introduce an 3D Encoder-Decoder CNN by extending a 3D Generative Adversarial Network~\cite{3dgan}, namely 3D Encoder-Decoder Generative Adversarial Network (3D-ED-GAN), to accomplish the 3D inpainting task. Since 3D CNN is memory consuming and applying 3D-ED-GAN on a high-resolution volume is improbable, we only use 3D-ED-GAN to operate low-resolution voxels (say $32^3$). Then we treat 3D volume output of 3D-ED-GAN as a sequence of 2D images and reconstruct the object slice by slice. A Long-term Recurrent Convolutional Network (LRCN) based on LSTM is proposed to recover fine-grained details and produce high-resolution results. LRCN functions as an upsampling network while completing details by learning from the dataset. 

We now describe our network structure of 3D-ED-GAN and LRCN respectively and the details of the training procedure.

\begin{figure}[t]
    \includegraphics[width=.5\textwidth]{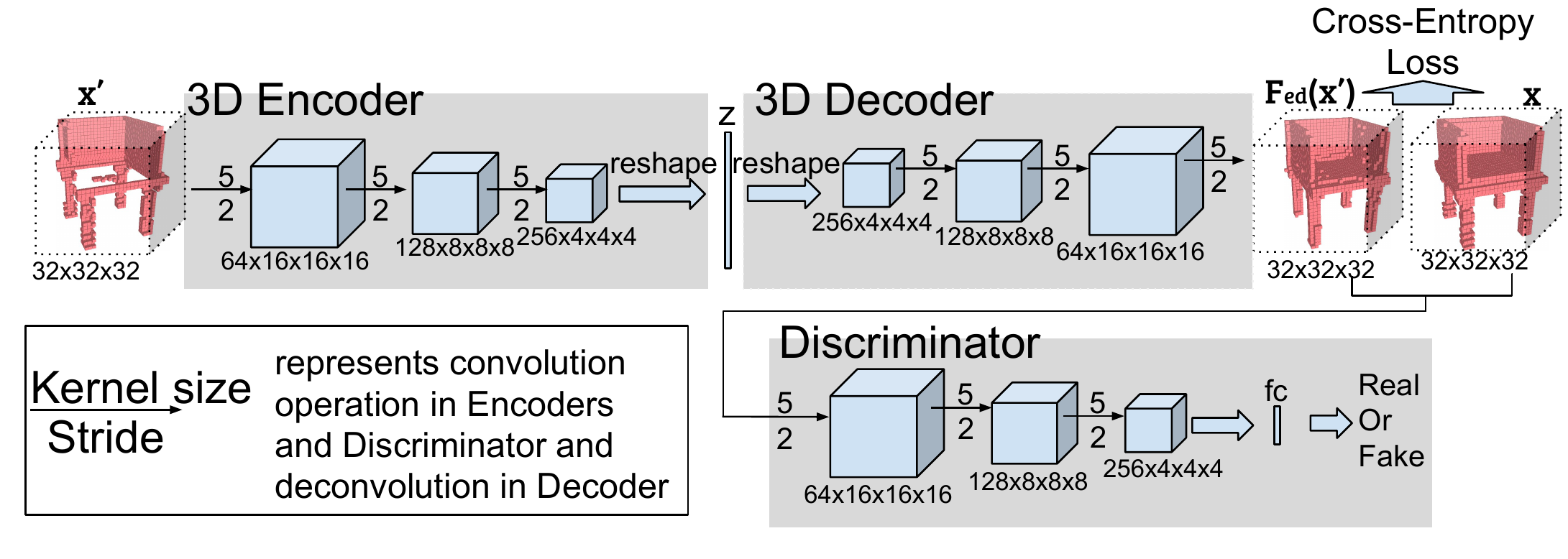}
    \caption{Network architecture of our 3D-ED-GAN.}
    \label{fig:EDGAN}
\end{figure}
\subsection{3D Encoder-Decoder Generative Adversarial Network (3D-ED-GAN)}

The Generative Adversarial Network (GAN) consists of a generator $G$ that maps a noise distribution $\mathbf{Z}$ to the data space $\mathbf{X}$, and a discriminator $D$ that classifies whether the generated sample is real or fake. $G$ and $D$ are both deep networks that are learned jointly. $D$ distinguishes real samples from synthetic data. $G$ tries to generate "real" samples to confuse $D$. Concretely, the objective of GAN is to achieve the following optimization:
  \begin{align}
\min_{G} \max_{D}  ( &\mathbb{E}_{\mathbf{x} \sim p_{data}(\mathbf{x})}[\log D(\mathbf{x})]+ \nonumber\\
&\mathbb{E}_{\mathbf{z} \sim p_{\mathbf{z}}(\mathbf{z})}[\log (1-D(G(\mathbf{z})))]), \label{eq:gan}
  \end{align}
where $p_{data}$ is data distribution and $p_{z}$ is noise distribution.

\paragraph*{Network structure} 3D-ED-GAN extends the general GAN framework by modeling the generator $G$ as a fully-convolutional Encoder-Decoder network, where the encoder maps input data into a latent vector $\mathbf{z}$. Then the decoder maps $\mathbf{z}$ to a cube. The 3D-ED-GAN consists of three components: an encoder, a decoder and a discriminator. Figure~\ref{fig:EDGAN} depicts the algorithmic architecture of 3D-ED-GAN.


The encoder takes a corrupted 3D volume $\mathbf{x}'$ of size ${d_l}^3$ (say $d_l = 32$) as input. It consists of three 3D convolutional layers with kernel size 5 and stride 2, connected via batch normalization (BN)~\cite{bn} and ReLU~\cite{prelu} layers. The last convolutional layer is reshaped into a vector $z$, which is the latent feature representation. There is no fully-connection (fc) layers. The noise vector in GAN is replaced with $z$. Therefore, the 3D-ED-GAN network conditions $z$ using the 3D encoder. We show that this latent vector carries informative features for supervised tasks in Section~\ref{sec:featurelearning}.

The decoder has the same architecture as $G$ in GAN, which maps the latent vector $z$ to a 3D voxel of size ${d_l}^3$. It has three volumetric full-convolution (also known as deconvolution) layers of kernel size 5 and strides 2 respectively, with BN and ReLU layers added in between. A $\tanh$ activation layer is added after the last layer. The Encoder-Decoder network is a fully-convolutional neural network without linear or pooling layers.

The discriminator has the same architecture as the encoder with an fc layer and a sigmoid layer at the end. 

\paragraph*{Loss function} The generator $G$ in 3D-ED-GAN is modeled by the Encoder-Decoder network. This can be viewed as a conditional GAN, in which the latent distribution is conditioned on given context data. Therefore, the loss function can been derived by reformulating the objective function in Equation~\ref{eq:gan}
  \begin{align}
L_{GAN}= &\mathbb{E}_{\mathbf{x} \sim p_{data}(\mathbf{x})}[\log D(\mathbf{x})+ \nonumber\\
&\log (1-D(F_{ed}(\mathbf{x}')))], \label{eq:3dgan}
  \end{align}
where $F_{ed}(\cdot): \mathbf{X} \rightarrow \mathbf{X}$ is the Encoder-Decoder network, and $\mathbf{x}'$ is the corrupted model of complete volume $\mathbf{x}$.

Similar to \cite{contextencoder}, we add an object reconstruction Cross-Entropy loss, $L_{recon}$, defined by
  \begin{align}
L_{recon}= &\frac{1}{N}\sum_{i=1}^{N} [x_{i} \log F_{ed}(\mathbf{x}')_{i} +\nonumber\\
&(1-x_{i}) \log (1-F_{ed}(\mathbf{x}')_{i})], \label{eq:recon}
  \end{align}
where $N={d_l}^3$, $x_{i}$ represents for the $i$th voxel of the complete volume $\mathbf{x}$ and $F_{ed}(\mathbf{x}')_{i}$ is the $i$th voxel of the generated volume. In this way, the output of the Encoder-Decoder network $F_{ed}(\mathbf{x}')$ is the probability of a voxel being filled.

The overall loss function for 3D-ED-GAN is
  \begin{align}
L_{3D-ED-GAN} = \alpha_1 L_{GAN} + \alpha_2 L_{recon},
  \end{align}
where $\alpha_1$ and $\alpha_2$ are weight parameters.

The loss function can effectively infer the structures of missing regions to produce conceivable reconstructions from the data distribution. Inpainting requires maintaining coherence of given context and producing plausible information according to the data distribution. 3D-ED-GAN has the capability of capturing the correlation between a latent space and the data distribution, thus producing appropriate plausible hypothesis.
\subsection  {Long-term Recurrent Convolutional Network (LRCN) Model}
\label{sec:LRCN}

\begin{figure}[t]
    \includegraphics[width=.5\textwidth]{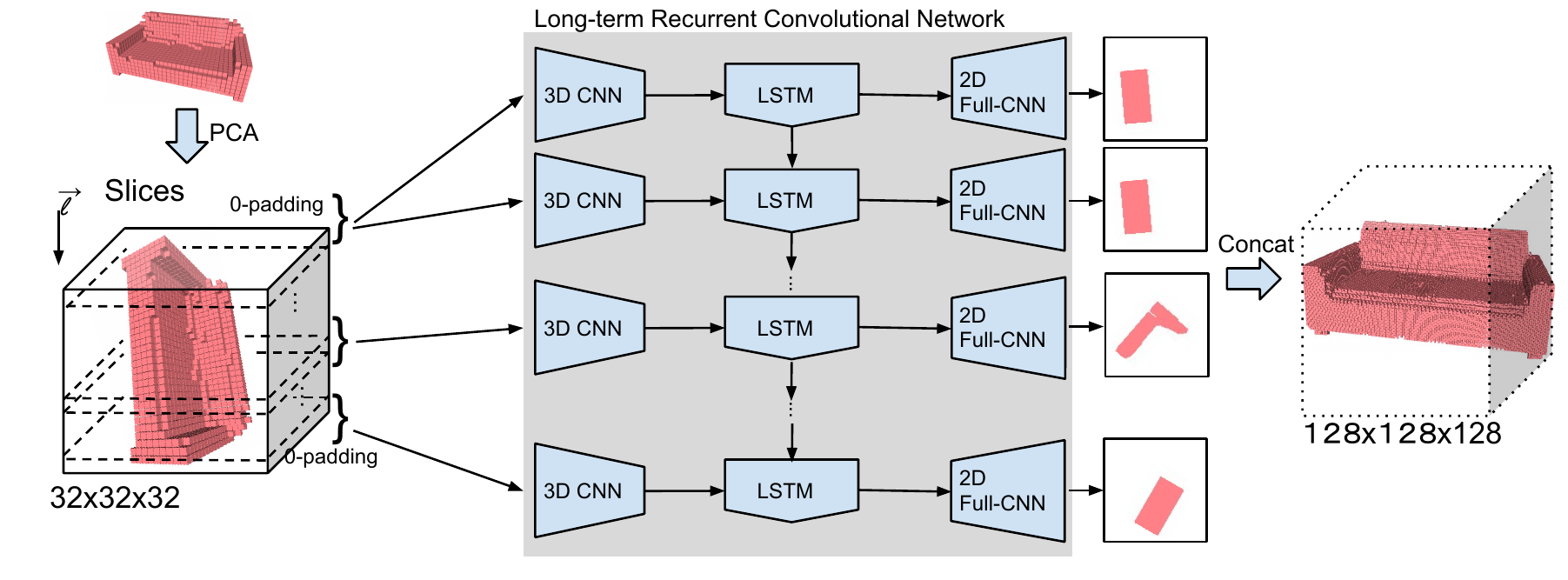}
    \caption{Framework for LRCN. The 3D input volumes are aligned by PCA and sliced along the first principle component into 2D images. LRCN processes $c$ ($c=5$) consecutive images with a 3D CNN, whose outputs are fed into LSTM. The outputs of LSTM further go through a 2D CNN and produce a sequence of high-resolution 2D images. The concatenations of these 2D images are the high-resolution 3D completion results.}
    \label{fig:LRCN}
\end{figure}

3D CNN consumes much more GPU memory than 2D CNN. Extending 3D-ED-GAN by adding 3D convolution layers to produce high resolution output is improbable due to memory limitation. We take advantage of the capability of RNN to handle long-term sequential dependencies and treat the 3D object volume as slices of 2D images. The network is required to map a volume with dimension ${d_l}^3$ to a volume with dimension ${d_h}^3$ (we have $d_l=32, d_h=128$). For a sequence-to-sequence problem with different input and output dimensions, we integrate an encoder-decoder pair to the LSTM cell inspired by the video processing work~\cite{LRCNvideo}. Our LRCN model combines an LSTM, a 3D CNN, and 2D deep fully-convolutional network. It works by passing each 2D slice with its neighboring slices through a 3D CNN to produce a fixed-length vector representation as input to LSTM. The output vector of LSTM is passed through a 2D fully-convolutional decoder network and mapped to a high-resolution image. A sequence of high-resolution 2D images formulate the output 3D object volume. Figure~\ref{fig:LRCN} depicts our LRCN architecture.

\paragraph*{Formulation of Sequential Input} In order to obtain the maximal amount of contextual data from each 3D object volume, we would like to maximize the number of nonempty slices for the volume. So given a 3D object volume of dimension ${d_l}^3$, we firstly use principle component analysis (PCA) to align the 3D object and denote the aligned volume as $\mathbf{I}$ and its first principle component as direction $\overrightarrow{l}$\footnote{In our experiment implementation, we use PCA to align the corrupted objects instead of the output of 3D-ED-GAN.}. Then $\mathbf{I}$ is treated as a sequence of $d_l\times d_l$ 2D images along $\overrightarrow{l}$, denoted as $\{I_1, I_2,..., I_{d_l}\}$. Since the output of LRCN is a sequence with length $d_h$, the input sequence length should also be $d_h$. As illustrated in Figure~\ref{fig:LRCN}, for each step, a slice with its 4 neighboring slices (so 5 slices total) is formed into a thin volume and fed into the network, say for step $t$. And slices with negative indices, or indices beyond $d_l$, are 0-padded. The input of the 3D CNN is then $\mathbf{v}_t' = \{ I_{{\frac{t}{d_h/d_l} }- 2},I_{{\frac{t}{d_h/d_l} }-1},I_{\frac{t}{d_h/d_l} },I_{{\frac{t}{d_h/d_l} }+1},I_{{\frac{t}{d_h/d_l} }+2} \}$.

\paragraph*{Network structure} As illustrated in Figure~\ref{fig:LRCN}, the 3D CNN encoder takes a $d_l\times d_l \times c$ volume as input, where $c$ represents number of slices (we have $c=5$). At step $t$, the 3D CNN transforms $c$ slices of 2D images $\mathbf{v}_t'$ into a $200D$ vector $v_t$. The 3D CNN encoder has the same structure with the 3D encoder in 3D-ED-GAN with an fc layer at the end.  
After the 3D CNN, the recurrent model LSTM takes over. We use the LSTM cell as described in \cite{lstm}: Given input $v_t$, the LSTM updates at timestep $t$ are:
  \begin{flalign}
  &i_t=\sigma(W_{vi}v_t+W_{hi}h_{t-1}+W_{ci}c_{t-1}+b_i)&\notag\\
  &f_t=\sigma(W_{vf}v_t+W_{hf}h_{t-1}+W_{cf}c_{t-1}+b_f)&\notag\\
  &c_t=f_tc_{t-1}+i_t\tanh(W_{vc}v_t+W_{hc}h_{t-1}+b_c)\\
  &o_t=\sigma(W_{vo}v_t+W_{ho}h_{t-1}+W_{co}c_t+b_o)&\notag\\
  &h_t=o_t\tanh(c_t)&\notag
  \end{flalign}
where $\sigma$ is the logistic sigmoid function, $i,f,o,c$ are respectively $input gate, forget gate, output gate, cell gate$, $W_{vi,vf,vc,vo,hi,hf,hc,ho,ci,co,cf}$ and $b_{i,f,c,o}$ are parameters. 

The output vector of LSTM $o_t$ is further going through a 2D fully-convolutional neural network to generate a $d_h \times d_h$ image. It has two fully-convolutional layers of kernel size 5 and stride 2, with BN and ReLU in between followed by a $\tanh$ layer at the end.
\paragraph*{Loss function}
We experimented with both $l_1$ and $l_2$ losses and found the $l_1$ loss obtains higher-quality results. In this way, the $l_1$ loss is adopted to train our LRCN, denoted as $L_{LRCN}$.

The overall loss to jointly train the hybrid network (combination of 3D-ED-GAN and LRCN) is
  \begin{align}
  \label{eq:overallloss}
L= \alpha_3 L_{3D-ED-GAN}  + \alpha_4 L_{LRCN},
  \end{align}
  where $\alpha_3$ and $\alpha_4$ are weight parameters.

Although the LRCN contains a 3D CNN encoder, the thin input slices makes the network sufficiently small compared to a regular volumetric CNN. By taking advantage of RNN's ability to manipulate sequential data and long-range dependencies, our memory efficient network is able to produce high-resolution completion result.
\subsection {Training the hybrid network}
\label{sec:training}


Training our 3D-ED-GAN and LRCN both jointly and from scratch is a challenging task. Therefore, we propose a three-phase training procedure. 

In the first stage, 3D-ED-GAN is trained independently with corrupted 3D input and complete output references in low resolution. Since the discriminator learns much faster than the generator, we first train the Encoder-Decoder network independently without discriminator (with only reconstruction loss).  The learning rate is fixed to $10^{-5}$, and $20$ epochs are trained. Then we jointly train the discriminator and the Encoder-Decoder as in \cite{dcgan} for $100$ epochs.  We set the learning rate of the Encoder-Decoder to $10^{-4}$, and $D$ to $10^{-6}$. Then $\alpha_1$ and $\alpha_2$ are set to 0.001 and 0.999 respectively. For each batch, we only update the discriminator if its accuracy in the last batch is not higher than 80\% as in \cite{3dgan}. ADAM~\cite{adam} optimization is employed with $\beta=0.5$ and a batch size of $4$.

In the second stage, LRCN is trained independently with perfect 3D input in low resolution and high-resolution output references for $100$ epochs. We use a learning rate of $10^{-4}$ and a batch size of $4$. In this stage, LRCN works as an upsampling network capable of predicting fine-grained details from trained data distributions.

In the final training phase, we jointly finetune the hybrid network on the pre-trained networks in the first and second stages with loss defined as in Equation~\ref{eq:overallloss}. The learning rate of the discriminator is $10^{-7}$ and the learning rate of the remaining network is set to be $10^{-6}$ with batch size $1$. Then $\alpha_3$ and $\alpha_4$ are both set to 0.5. We observe that most of the parameter updates happen in LRCN. The input of LRCN in this stage is imperfect and the output reference is still complete high-resolution model, which indicates that LRCN works as a denoising network while maintaining its power of upsampling and preserving details.

For convenience, we use the aforementioned PCA method to align all models before training instead of aligning the predictions of 3D-ED-GAN.

\section{Experiments}
\label{sec:experiment}

Our network architecture is implemented using the deep learning library Tensorflow~\cite{tensorflow}. We extensively test and evaluate our method using various datasets canonical to 3D inpainting and feature learning.

We split each category in the ShapeNet dataset~\cite{shapenet} to mutually-excluded 80 training points and 20 testing points. Our network is trained on the training points as stated in Section~\ref{sec:training}. We train separate networks for seven major categories (chairs, sofas, tables, boats, airplanes, lamps, dressers, and cars) without fine-tuning on any existing models. 3D meshes are voxelized into $32^3$ grids for low-resolution input and $128^3$ grids for high-resolution output reference. The input 3D volumes are synthetically corrupted to simulate the imperfections of a real-world 3D scanner.

The following experiments are conducted with the trained model: We firstly evaluate the inpainting performance of 3D-ED-GAN on both real-world 3D range scans data and the ShapeNet 20-point testing set with various injected noise. Ablation experiments are conducted to assess the capability of producing high-resolution completion results from the combination of 3D-ED-GAN and LRCN. We also compare with the state-of-the-art method. Then, we evaluate the capability of 3D-ED-GAN as a feature learning framework. Please refer to the supplementary material for more results and comparisons.

\subsection{3D Objects Inpainting}
Our hybrid network has $26.3M$ parameters and requires $7.43$GB GPU memory. If we add two more full-convolution layers in the decoder and two more convolution layers in the discriminator of 3D-ED-GAN to produce high-resolution, the network has $116.4M$ parameters and won't fit into the GPU memory. For comparison between low-resolution and high-resolution results, we simply upsample the prediction of 3D-ED-GAN and do numerical comparisons. 
\subsubsection{Real-World Scans}
\begin{figure}[tb]
\centering

    \setlength{\tabcolsep}{0.1em}
\newcolumntype{C}{>{\centering\arraybackslash}p{3.53em}}
\renewcommand{\arraystretch}{0} 
    \begin{tabular}{CCC|CCC}
\includegraphics[width=0.06\textwidth]{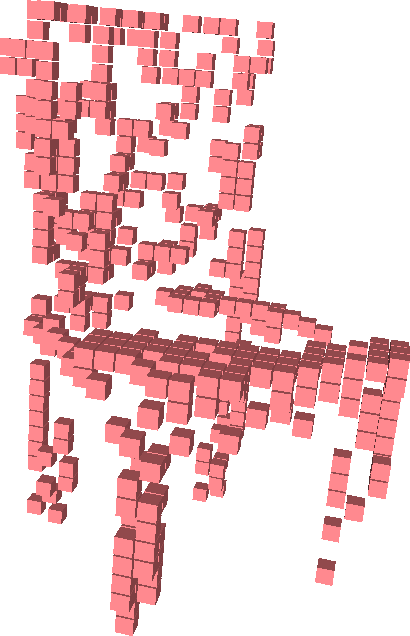}
&\includegraphics[width=0.054\textwidth]{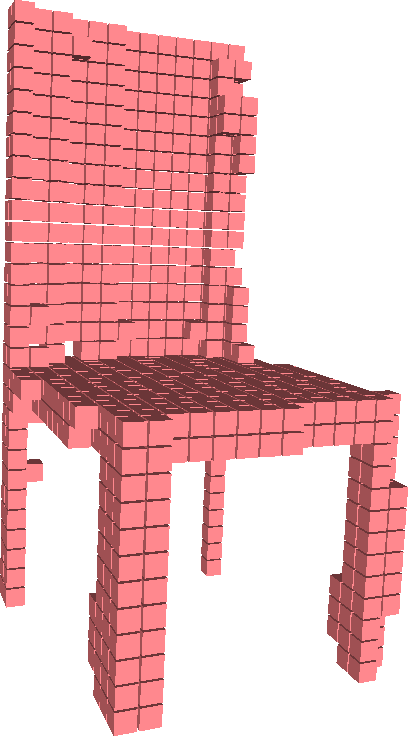}
&\includegraphics[width=0.052\textwidth]{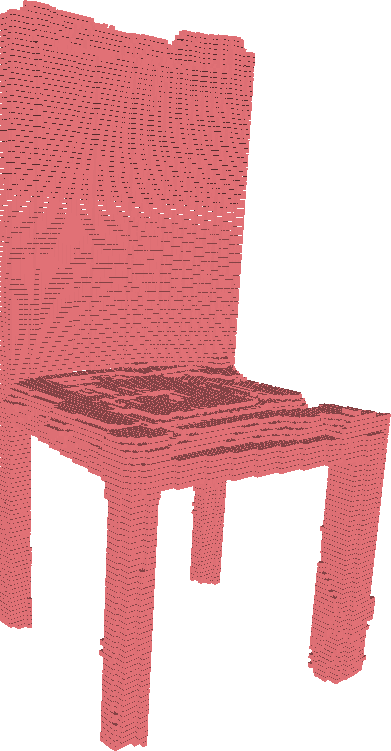}
&\includegraphics[width=0.07\textwidth]{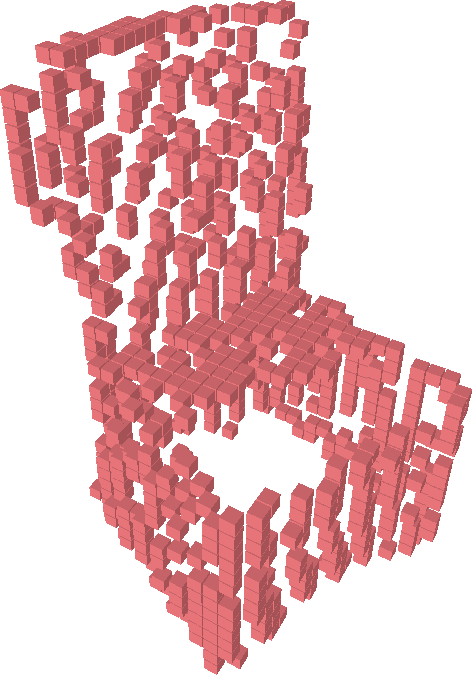}
&\includegraphics[width=0.07\textwidth]{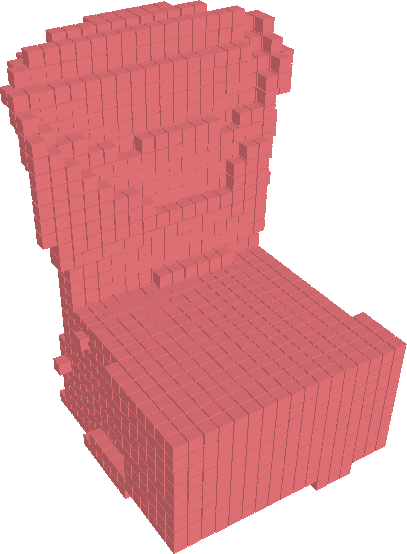}
&\includegraphics[width=0.07\textwidth]{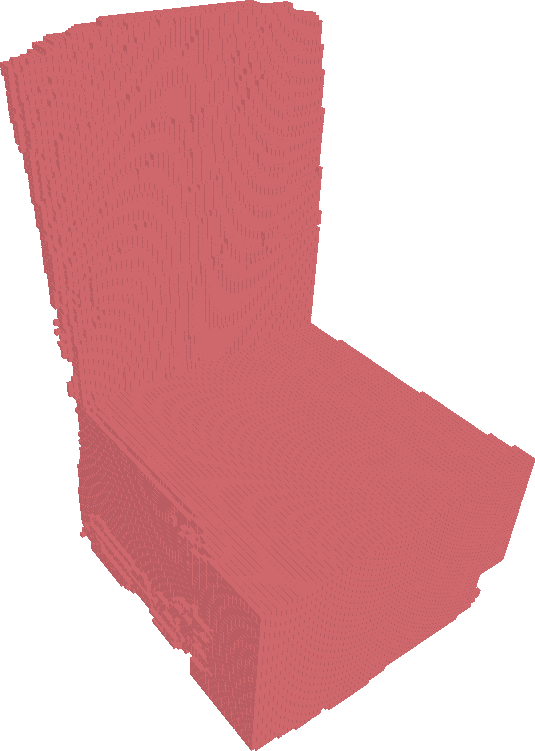}
\\
\includegraphics[width=0.073\textwidth]{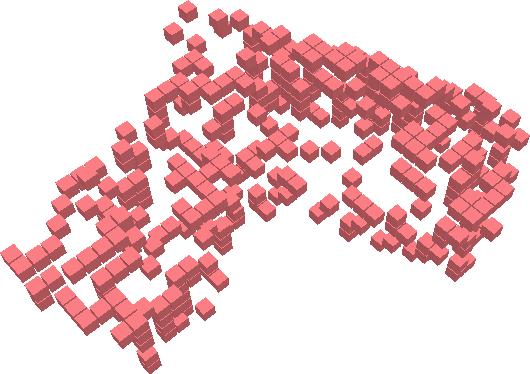}
&\includegraphics[width=0.073\textwidth]{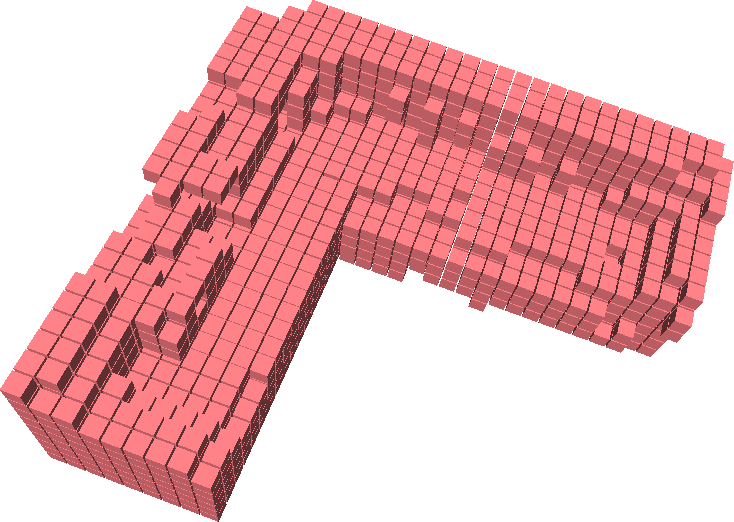}
&\includegraphics[width=0.073\textwidth]{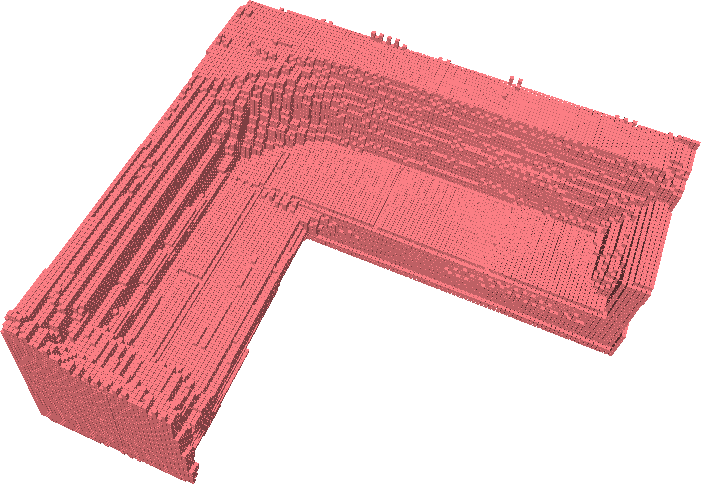}
&\includegraphics[width=0.073\textwidth]{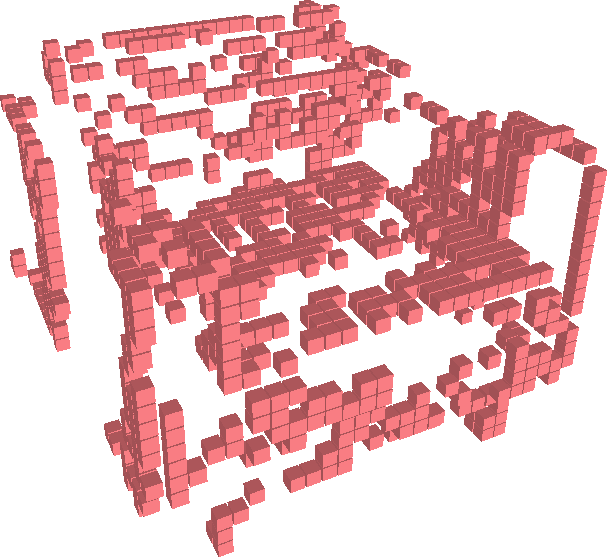}
&\includegraphics[width=0.073\textwidth]{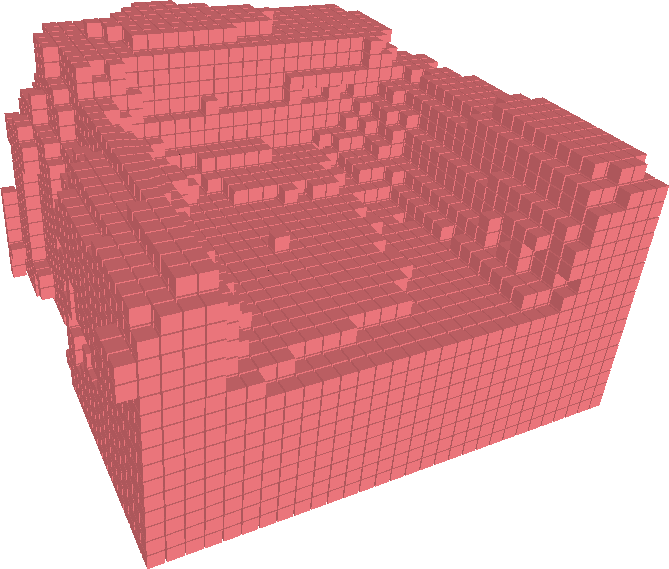}
&\includegraphics[width=0.073\textwidth]{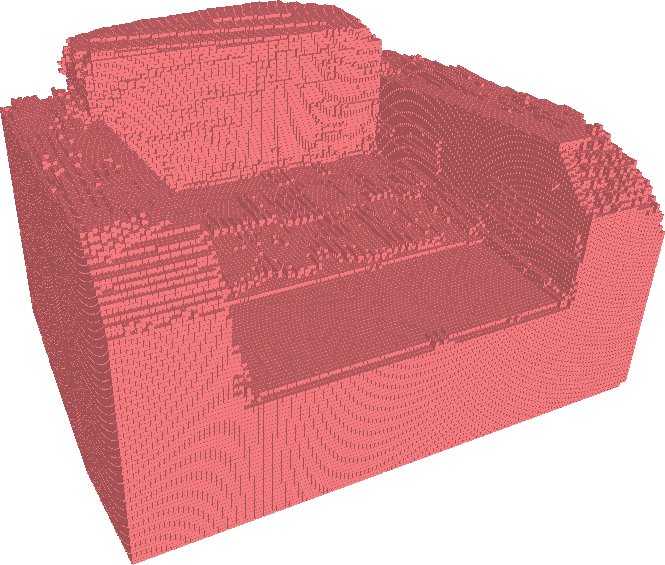}
\\
\includegraphics[width=0.073\textwidth]{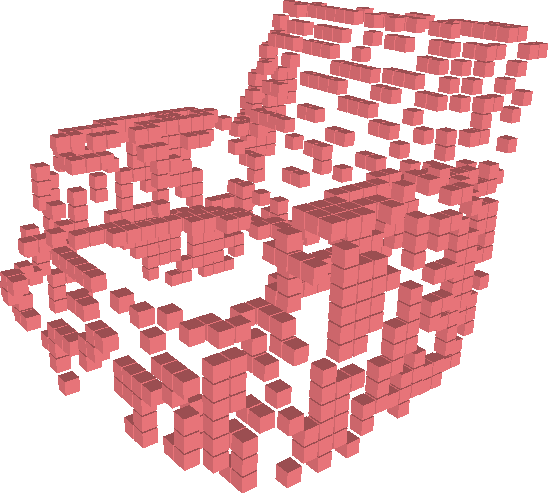}
&\includegraphics[width=0.073\textwidth]{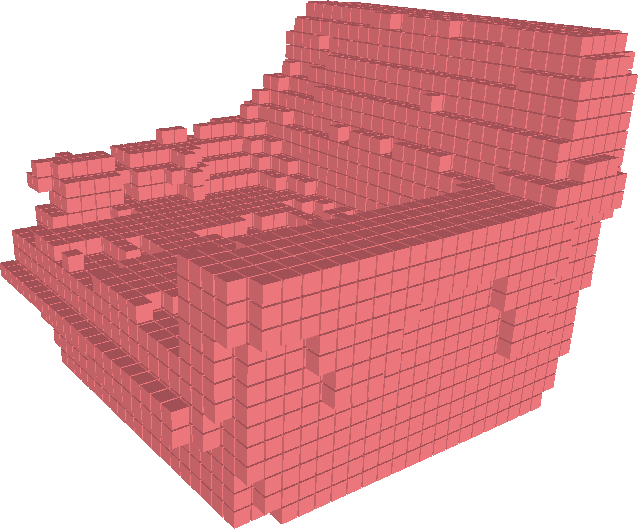}
&\includegraphics[width=0.073\textwidth]{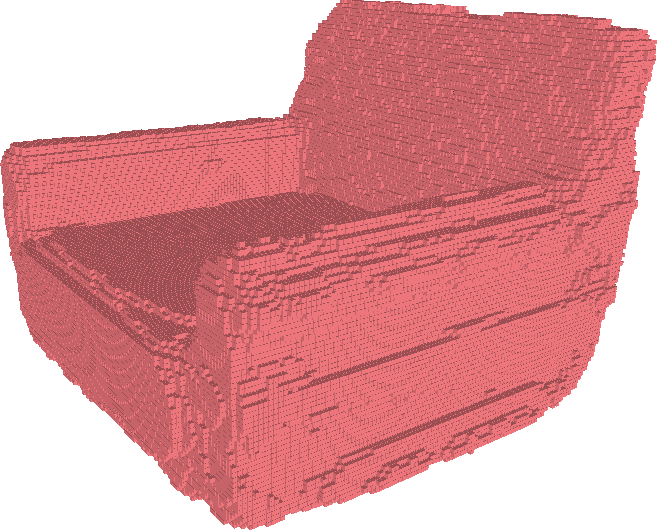}
&\includegraphics[width=0.073\textwidth]{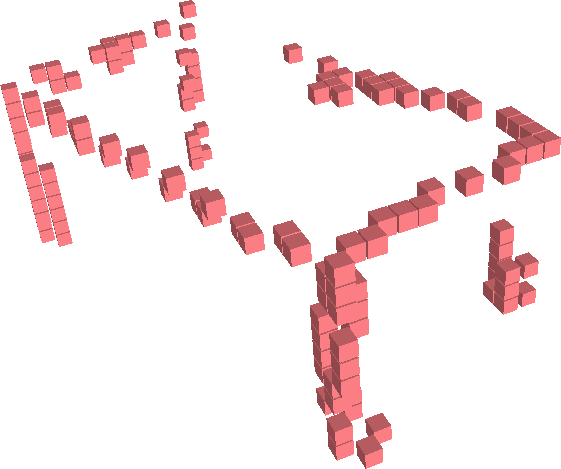}
&\includegraphics[width=0.073\textwidth]{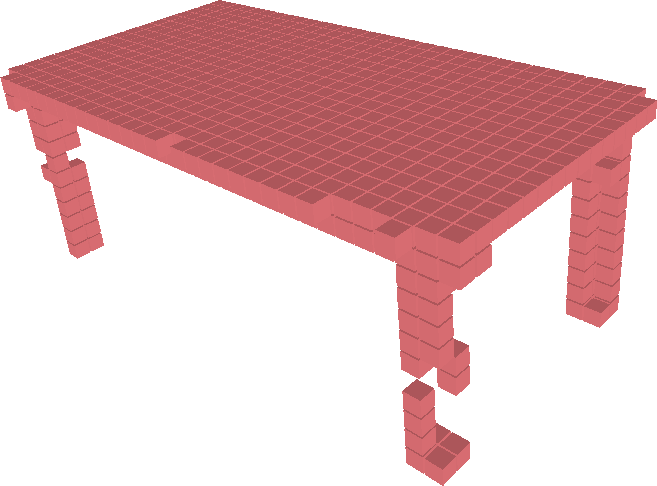}
&\includegraphics[width=0.073\textwidth]{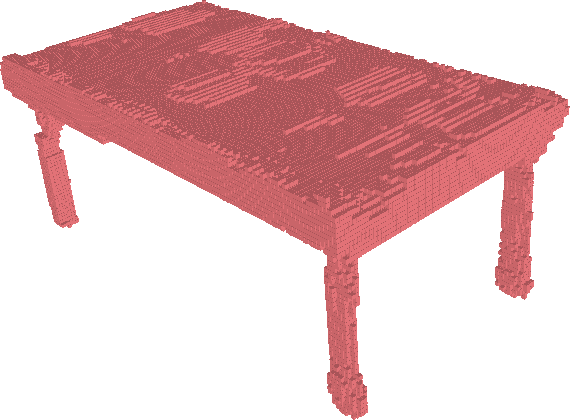}
\\
\includegraphics[width=0.073\textwidth]{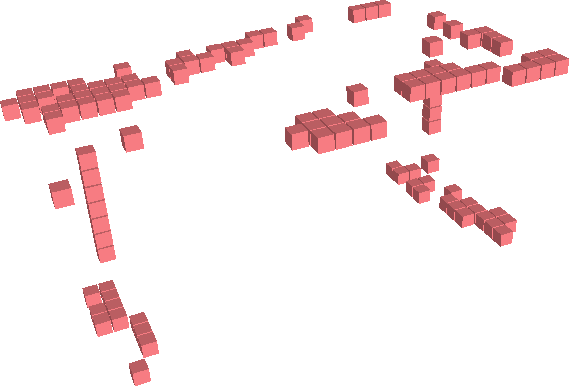}

&\includegraphics[width=0.073\textwidth]{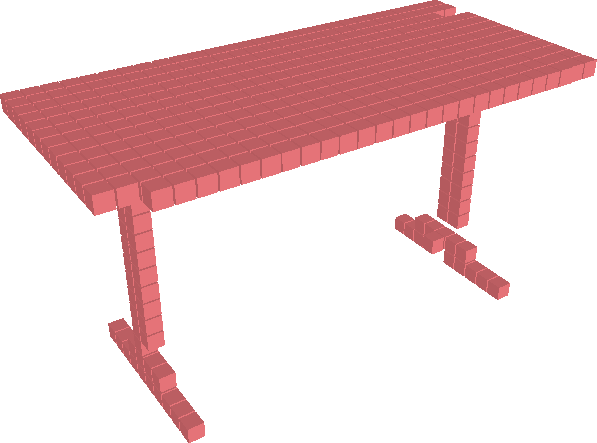}

&\includegraphics[width=0.073\textwidth]{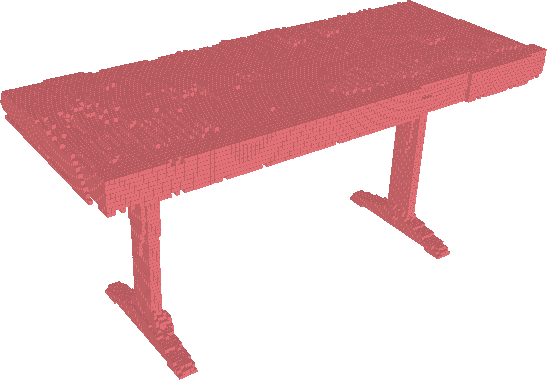}

&\includegraphics[width=0.073\textwidth]{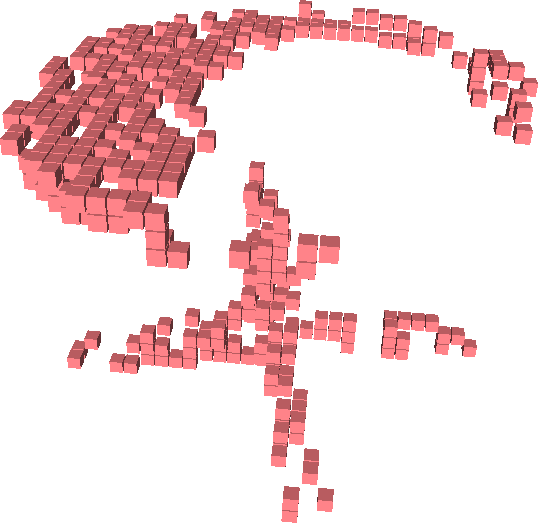}

&\includegraphics[width=0.073\textwidth]{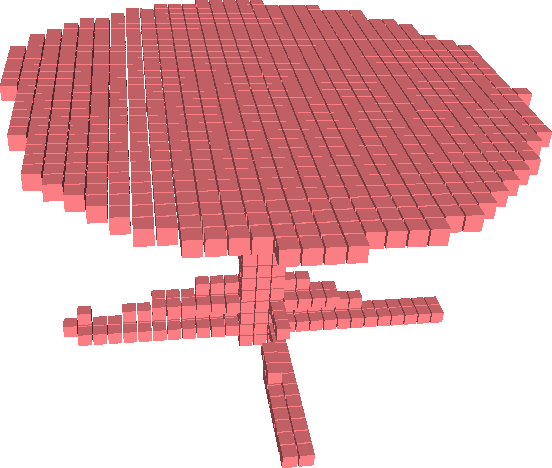}

&\includegraphics[width=0.073\textwidth]{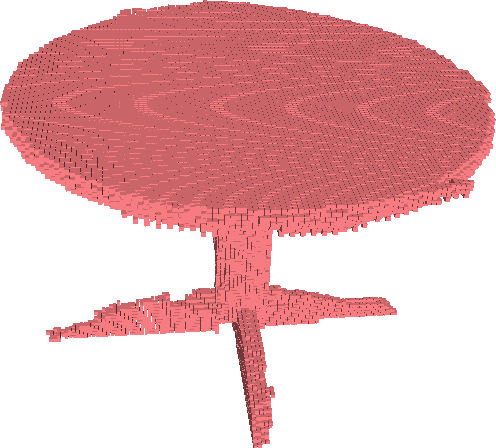}

\\

\scriptsize Input & \scriptsize 3D-ED-GAN & \scriptsize Hybrid&

\scriptsize Input & \scriptsize 3D-ED-GAN & \scriptsize Hybrid
    \end{tabular}
    \caption{3D completion results on real-world scans. Inputs are the voxelized scans. 3D-ED-GAN represents the low-resolution completion result without going through LRCN. Hybrid represents the high-resolution completion result of the combination of 3D-ED-GAN and LRCN.}
\label{fig:realdata}
\end{figure}

We test 3D-ED-GAN and LRCN on both real-world and synthetic data. The real-world scans are from the work of \cite{qimultiview}. They reconstructed 3D mesh from RGB-D data and we voxelized these 3D meshes into $32^3$ grids for test. Our network is trained on ShapeNet dataset as in Section~\ref{sec:training}. Before testing, all shapes are aligned using PCA as stated in Section~\ref{sec:LRCN}. Figure~\ref{fig:realdata} shows shape completion examples on real-world scans for both low-resolution and high-resolution outputs. We use 3D-ED-GAN to represent the low-resolution output of 3D-ED-GAN and Hybrid to denote the high-resolution output of the combination of 3D-ED-GAN and LRCN. As we can see, our network is able to produce plausible completion results even with large missing area. The 3D-ED-GAN itself can result conceivable outputs while LRCN further improves fine-grained details.

\subsubsection{Random Noise}
\label{sec:randomnoise}
We then assess our model with the splitted testing data of ShapeNet. This is applicable in cases where capturing the geometry of objects with 3D scanners results in holes and incomplete shapes. Since it is hard to obtain ground truth for real-world objects, we rely on the ShapeNet dataset where complete object geometry of diversified categories is available, and we test on data with simulated noises. 

Because it is hard to predict the exact noise from 3D scanning, we test different noise characteristics and show the robustness of our trained model. We do the following ablation experiments:
\begin{enumerate}
\item 3D-ED-GAN: 3D-ED-GAN is trained the first training stage of Section \ref{sec:training}.
\item LRCN: After the LRCN is pre-trained as the second training stage in Section~\ref{sec:training}, we directly feed the partially scanned 3D volume into the LRCN as input and train LRCN independently for $100$ epochs with a learning rate of $10^{-5}$ and a batchsize 4. We test the shape completion ability of this single network.
\item Hybrid: Our overall network, i.e. the combination of 3D-ED-GAN and LRCN, is trained with the aforementioned procedure.
\end{enumerate} 

To have a better understanding of the effectiveness of our generative adversarial model, we also compare qualitatively and quantitatively with VConv-DAE~\cite{vconvdae}. They adopted a full convolutional volumetric autoencoder network architecture to estimate voxel occupancy grids from noisy data. The major difference between 3D-ED-GAN and VConv-DAE is introduction of GAN. In our implementation of VConv-DAE, we simply remove the discriminator from 3D-ED-GAN and compare the two networks with the same parameters.

\begin{figure}[tb]
\centering

    \setlength{\tabcolsep}{0.1em}
\newcolumntype{C}{>{\centering\arraybackslash}p{3.53em}}
\renewcommand{\arraystretch}{0} 
    \begin{tabular}{CCCCCC}
\includegraphics[width=0.073\textwidth]{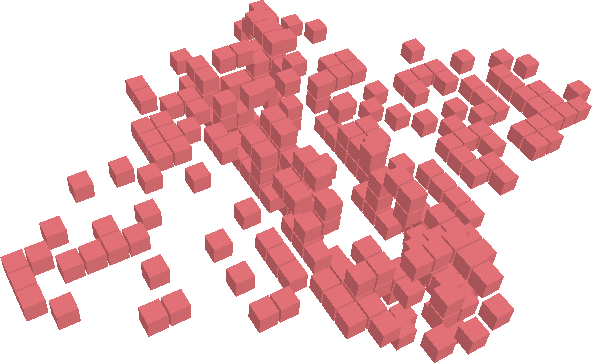}
&\includegraphics[width=0.073\textwidth]{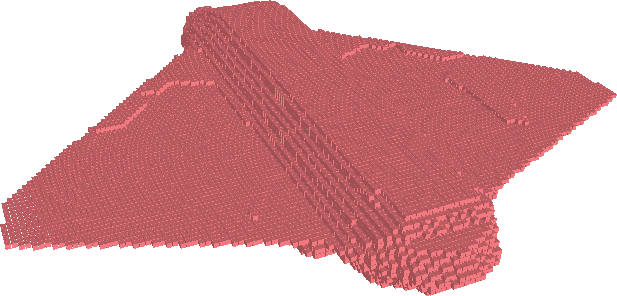}
&\includegraphics[width=0.073\textwidth]{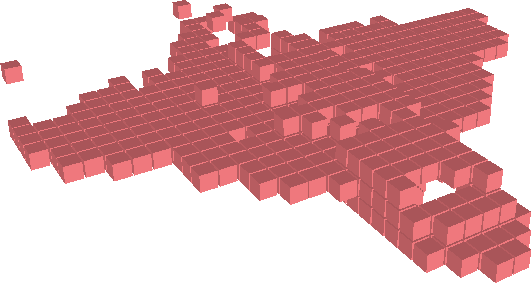}
&\includegraphics[width=0.073\textwidth]{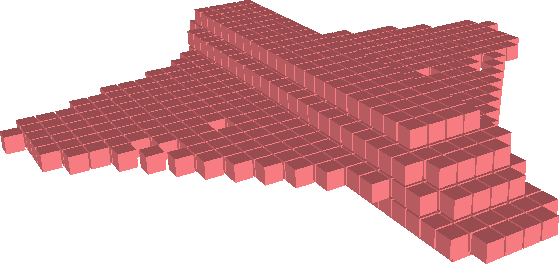}
&\includegraphics[width=0.073\textwidth]{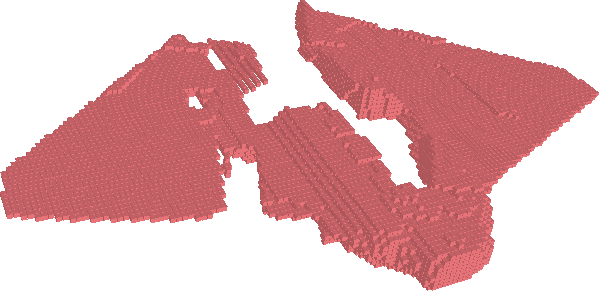}
&\includegraphics[width=0.073\textwidth]{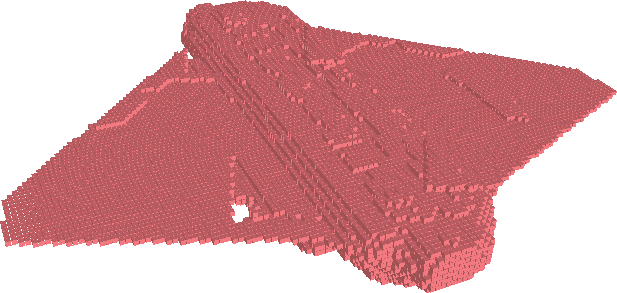}
\\
\includegraphics[width=0.073\textwidth]{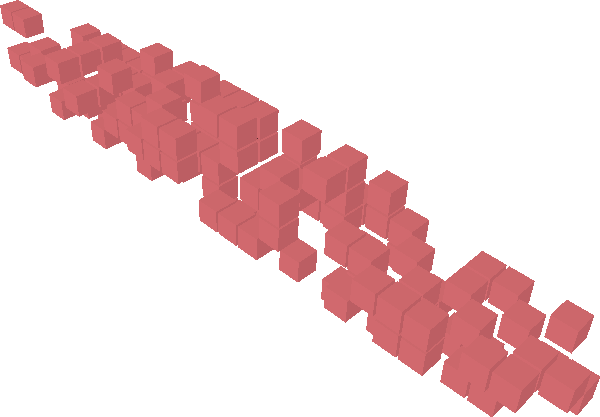}
&\includegraphics[width=0.073\textwidth]{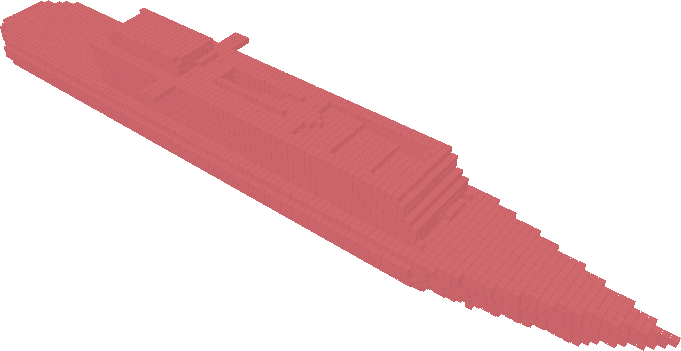}
&\includegraphics[width=0.073\textwidth]{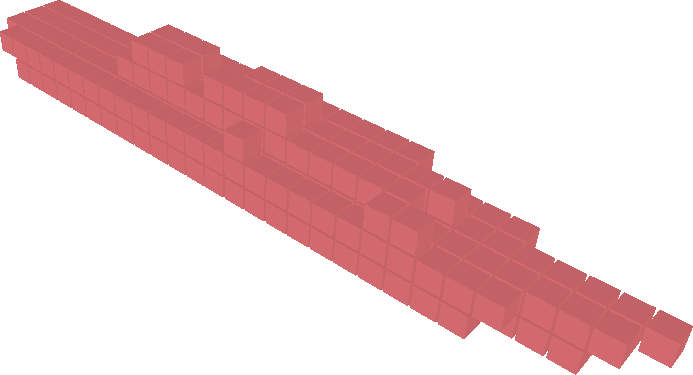}
&\includegraphics[width=0.073\textwidth]{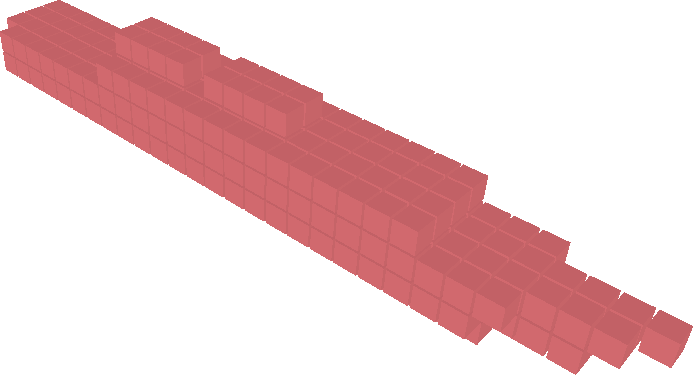}
&\includegraphics[width=0.073\textwidth]{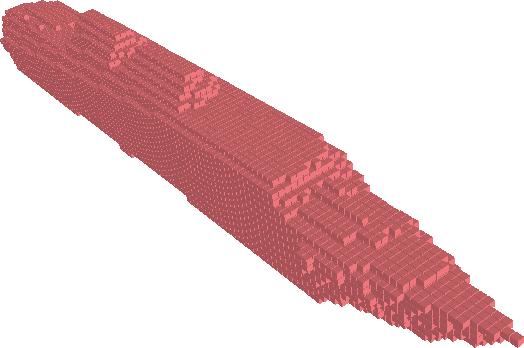}
&\includegraphics[width=0.073\textwidth]{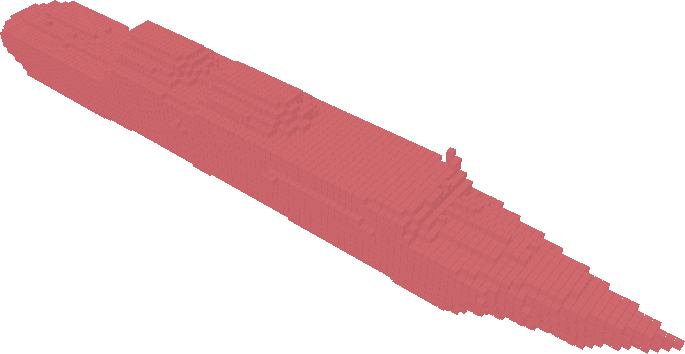}
\\
\includegraphics[width=0.065\textwidth]{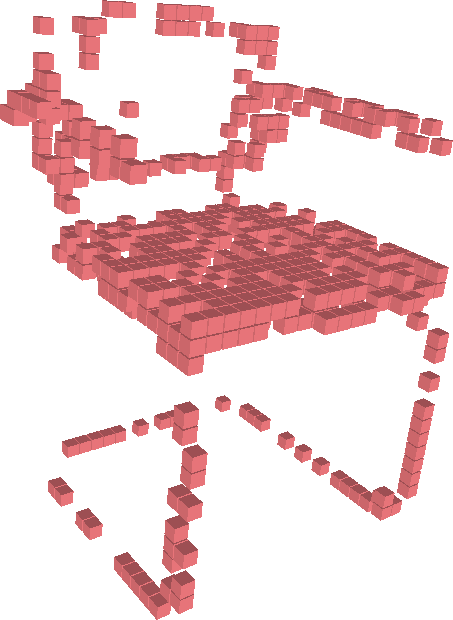}

&\includegraphics[width=0.065\textwidth]{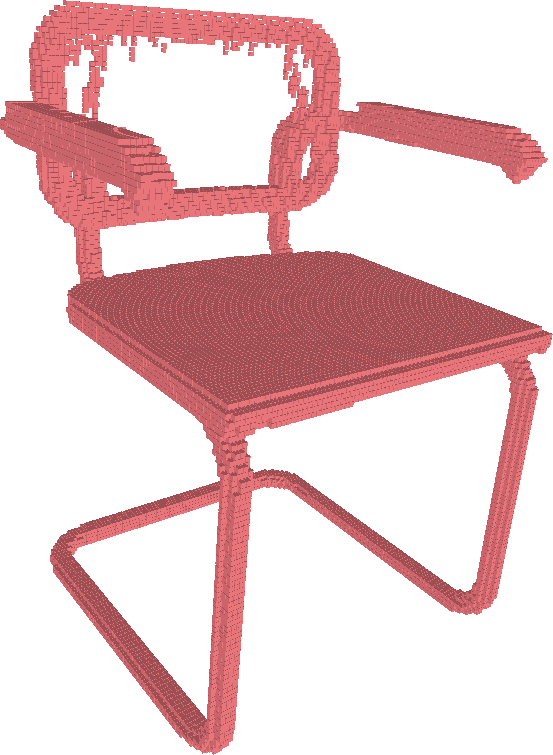}

&\includegraphics[width=0.06\textwidth]{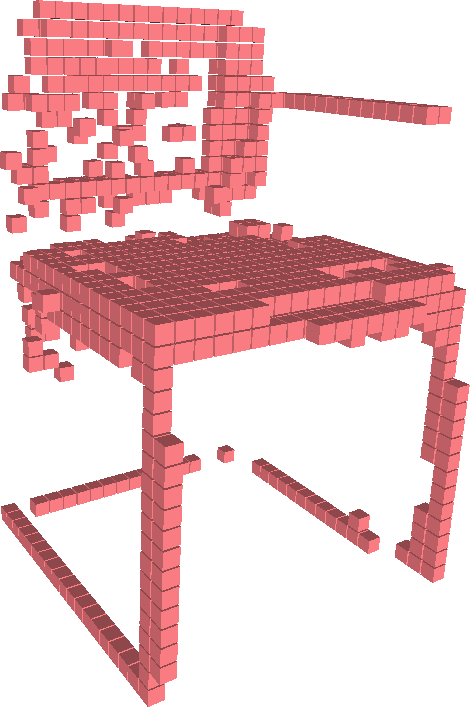}

&\includegraphics[width=0.065\textwidth]{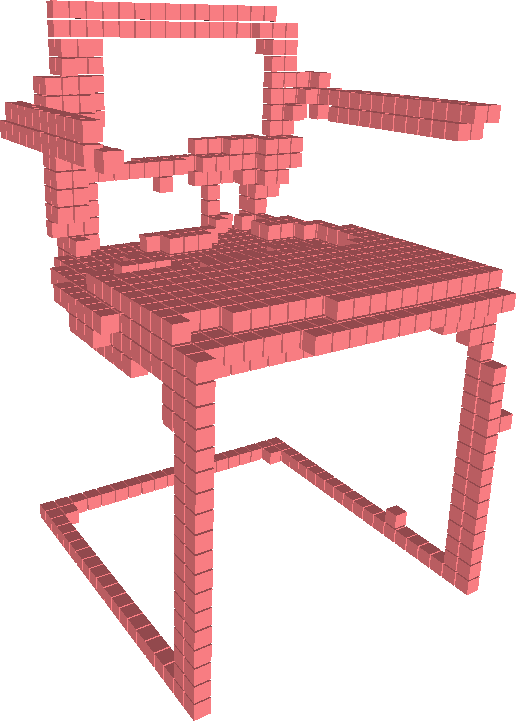}

&\includegraphics[width=0.065\textwidth]{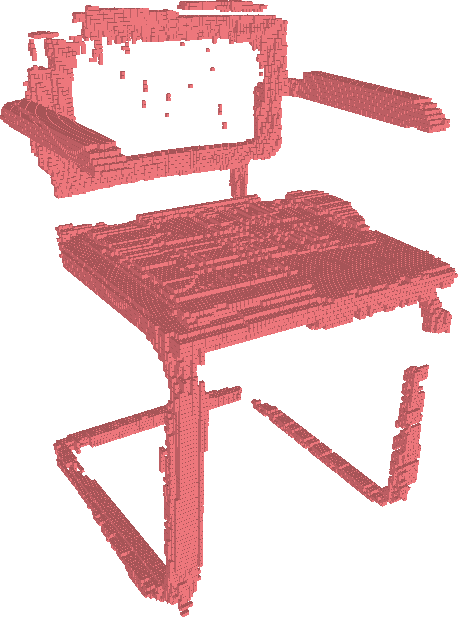}

&\includegraphics[width=0.065\textwidth]{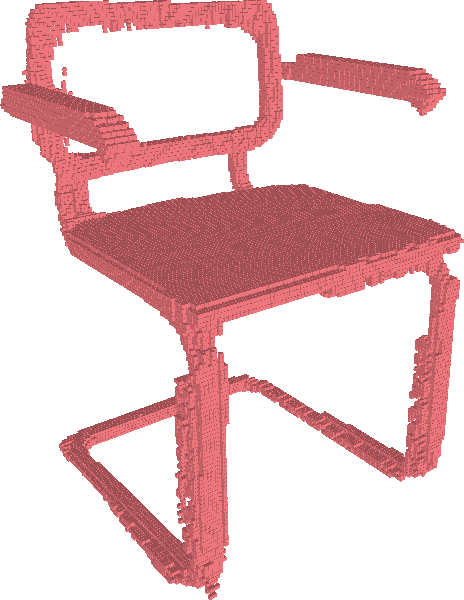}

\\

\scriptsize Input &

\scriptsize  Ground Truth & \scriptsize VConv-DAE & \scriptsize 3D-ED-GAN & \scriptsize LCRN & \scriptsize Hybrid
    \end{tabular}
    \caption{3D inpainting results with $50\%$ injected noise on ShapeNet test dataset. For this noise type, detailed information is missing while the global structure is preserved.}
\label{fig:randomcompletion}
\end{figure}

We first evaluate our model on test data with random noise. As stated above, we adopted simulated scanning noise in our training procedure. With random noise, volumes have to be recovered from limited given information, where the testing set and the training set have different patterns. Figure~\ref{fig:randomcompletion} shows the results of different methods for shape completion with $50\%$ noise injected.

\begin{figure}[t]
    \centering \includegraphics[width=.2\textwidth]{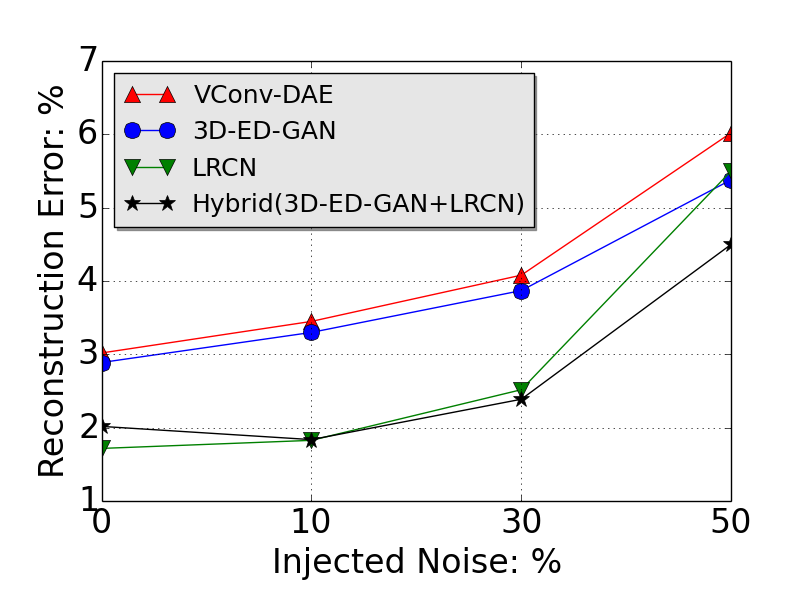}
    \caption{We vary the amount of random noise injected to test data and quantitatively compare the reconstruction error.}
    \label{fig:randomnoise}
\end{figure}

We also vary the amount of noise injected to the data. For numerical comparison, the number $n$ of generated voxels (at $128^3$ resolution) which differ from ground truth (object volume before corruption) is counted for each sample. The reconstruction error is $n$ divided by total number of grids $128^3$. For 3D-ED-GAN and VConv-DAE, their predictions are computed by upsampling the low resolution output. We use mean error for different object categories as our evaluation metric. The results are reported in Figure~\ref{fig:randomnoise}.
It can be seen from Figure~\ref{fig:randomcompletion} that different methods produce similar results. Even though $50\%$ noise is injected, the corrupted input still maintains the global semantic structure of the original 3D shape. In this way, this experiment measures the denoising ability of these models. As illustrated in Figure~\ref{fig:randomnoise}, these models introduce noise when $0\%$ noise injected. LRCN performs better than the other three when the noise percentage is low. When the input gets more corrupted, 3D-ED-GAN tends to perform better than others.

\subsubsection{Simulated 3D scanner}

\begin{figure*}[tb]
\centering

    \setlength{\tabcolsep}{0.1em}
\newcolumntype{C}{>{\centering\arraybackslash}m{3.53em}}
\newcolumntype{V}{>{\centering\arraybackslash}m{2.53em}}
\newcolumntype{M}{>{\centering\arraybackslash}m{4em}}
\renewcommand{\arraystretch}{0} 
    \begin{tabular}{VVVVVV|VVVVVV|VVVVVV}

\includegraphics[width=0.048\textwidth]{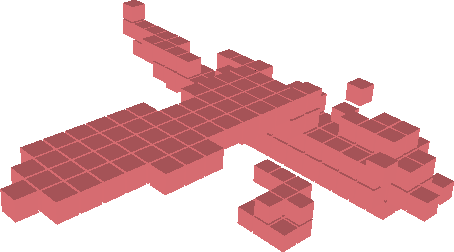}
&\includegraphics[width=0.048\textwidth]{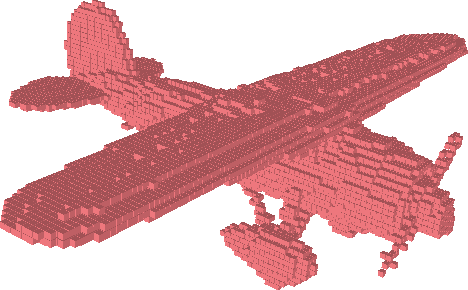}
&\includegraphics[width=0.048\textwidth]{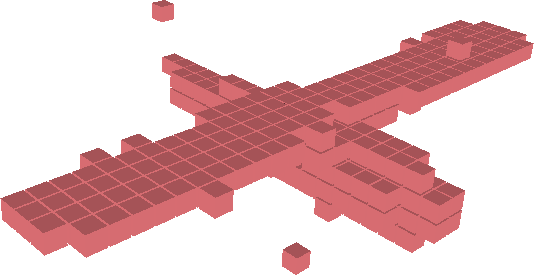}
&\includegraphics[width=0.048\textwidth]{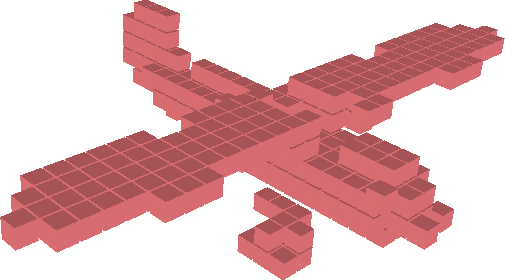}
&\includegraphics[width=0.048\textwidth]{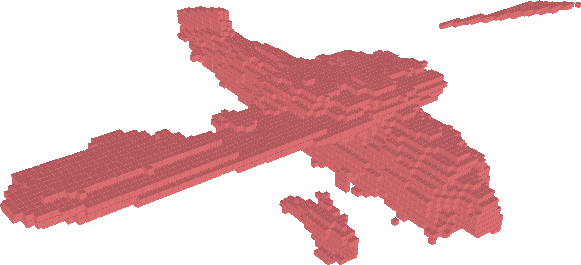}
&\includegraphics[width=0.048\textwidth]{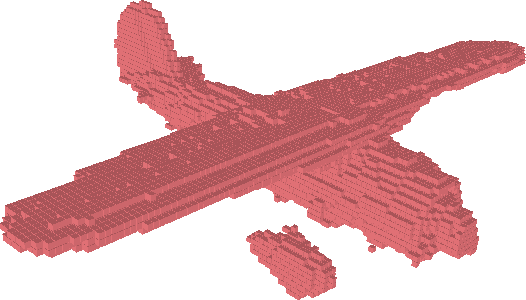}
&\includegraphics[width=0.048\textwidth]{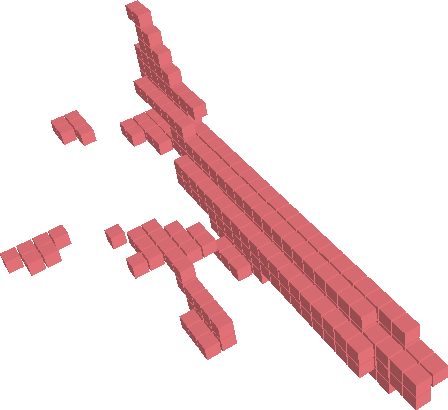}
&\includegraphics[width=0.048\textwidth]{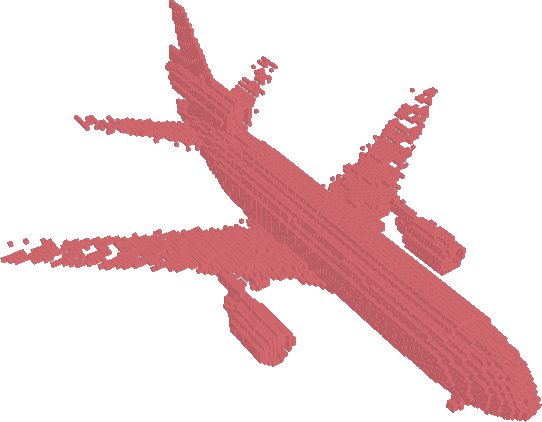}
&\includegraphics[width=0.048\textwidth]{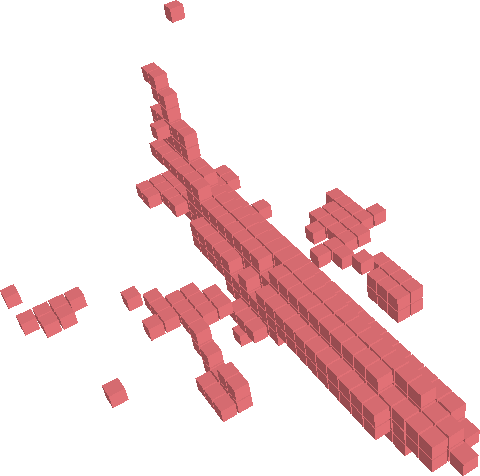}
&\includegraphics[width=0.048\textwidth]{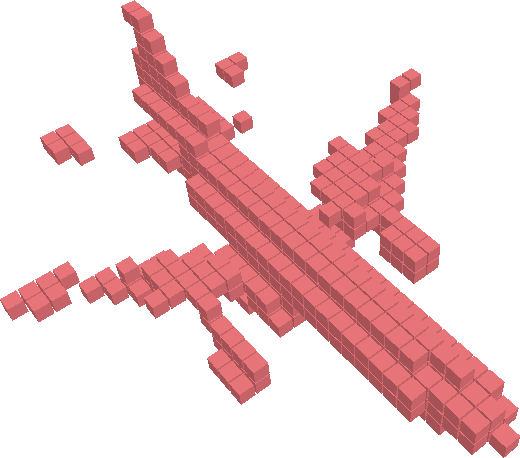}
&\includegraphics[width=0.048\textwidth]{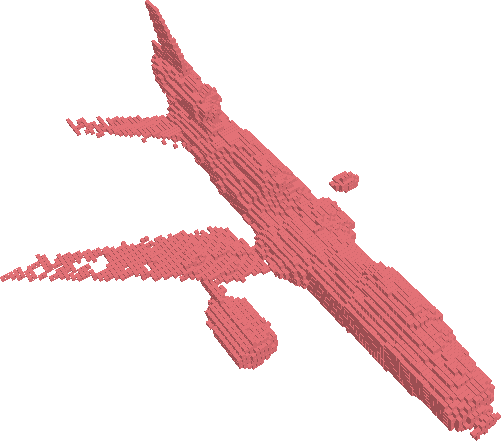}
&\includegraphics[width=0.048\textwidth]{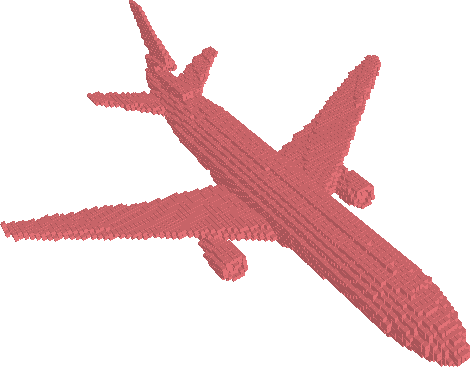}
&\includegraphics[width=0.048\textwidth]{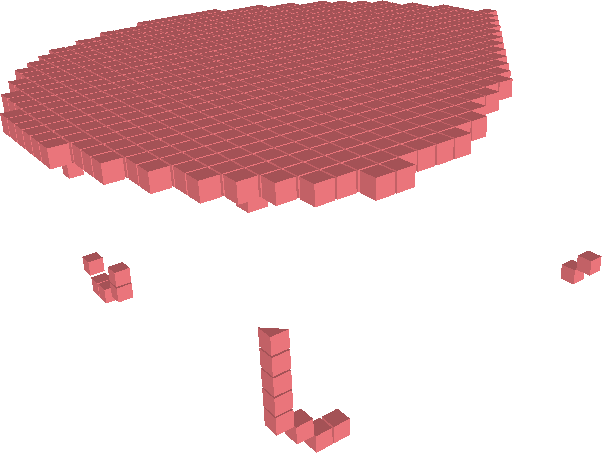}
&\includegraphics[width=0.048\textwidth]{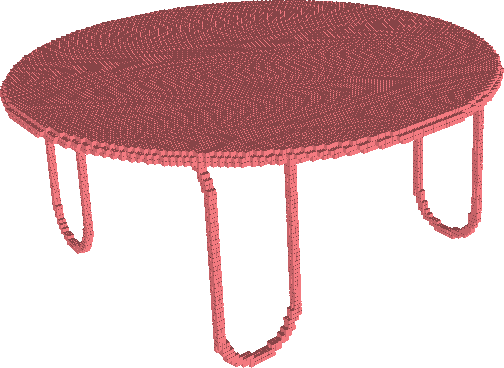}
&\raisebox{.85\height}{\includegraphics[width=0.048\textwidth]{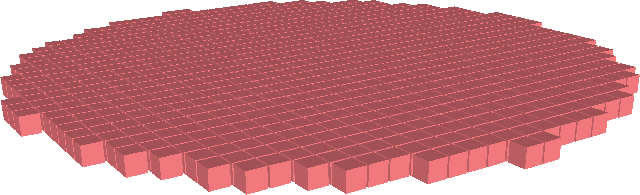}}
&\includegraphics[width=0.048\textwidth]{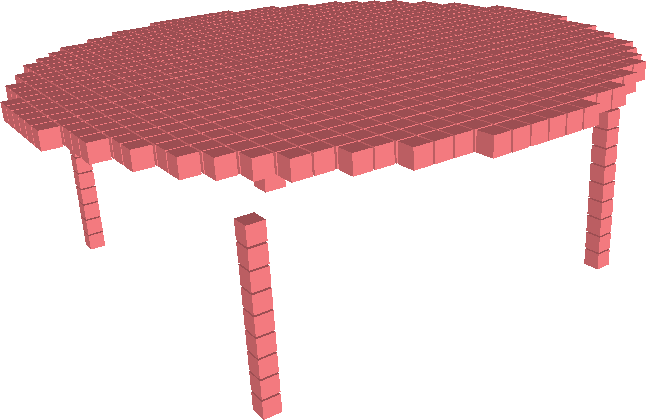}
&\includegraphics[width=0.048\textwidth]{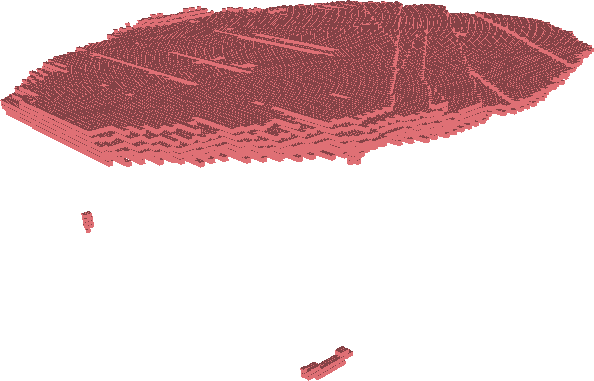}
&\includegraphics[width=0.048\textwidth]{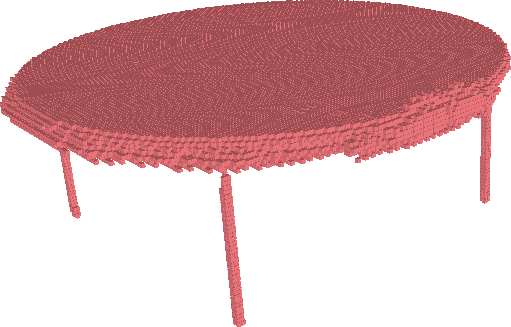}
\\
\includegraphics[width=0.048\textwidth]{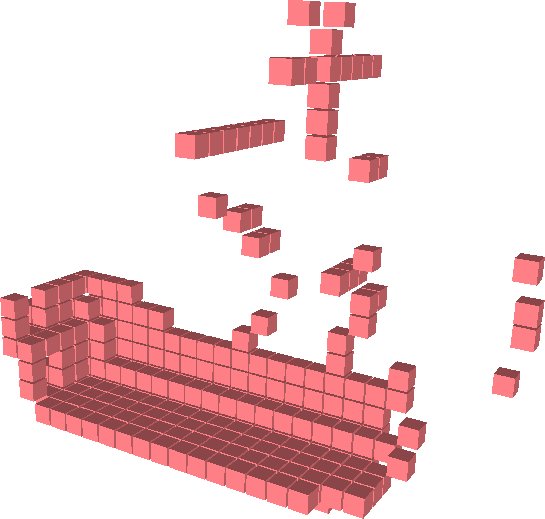}
&\includegraphics[width=0.048\textwidth]{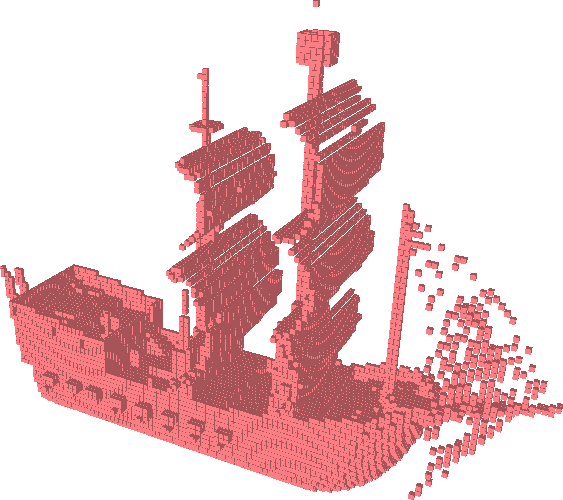}
&\includegraphics[width=0.048\textwidth]{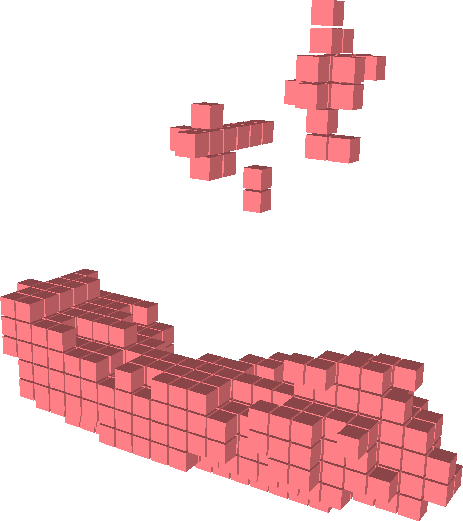}
&\includegraphics[width=0.048\textwidth]{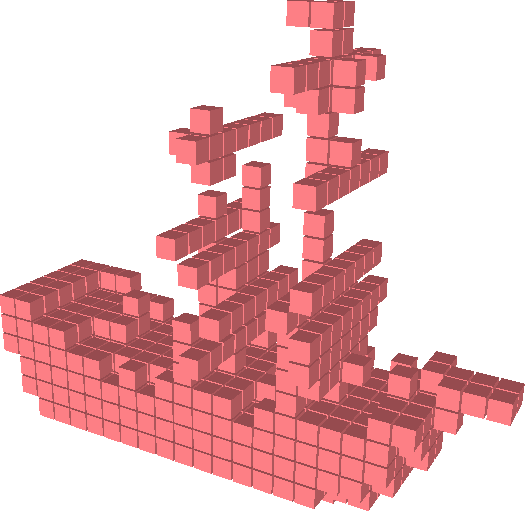}
&\includegraphics[width=0.048\textwidth]{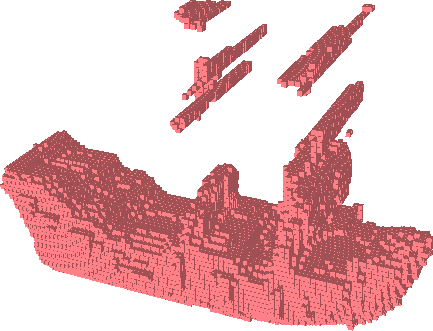}
&\includegraphics[width=0.048\textwidth]{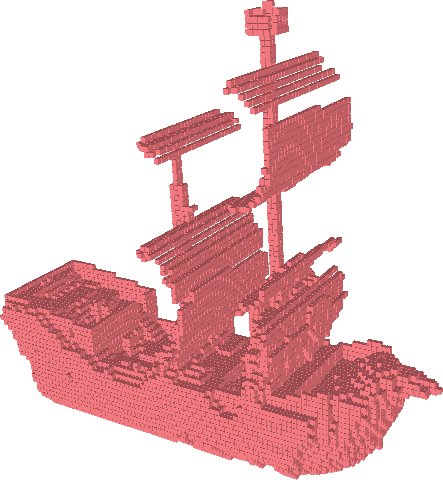}
&\includegraphics[width=0.048\textwidth]{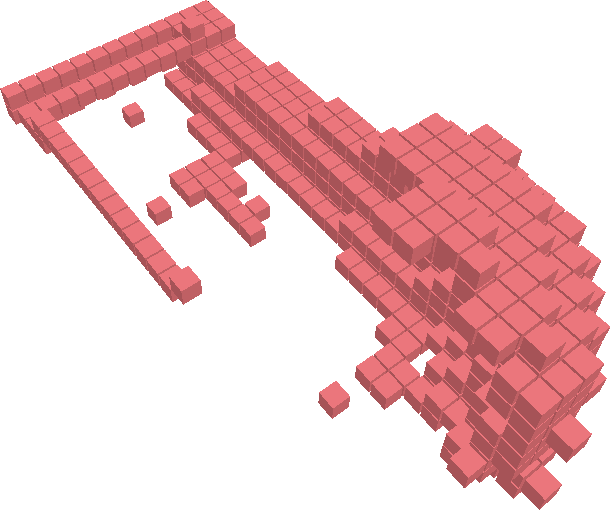}
&\includegraphics[width=0.048\textwidth]{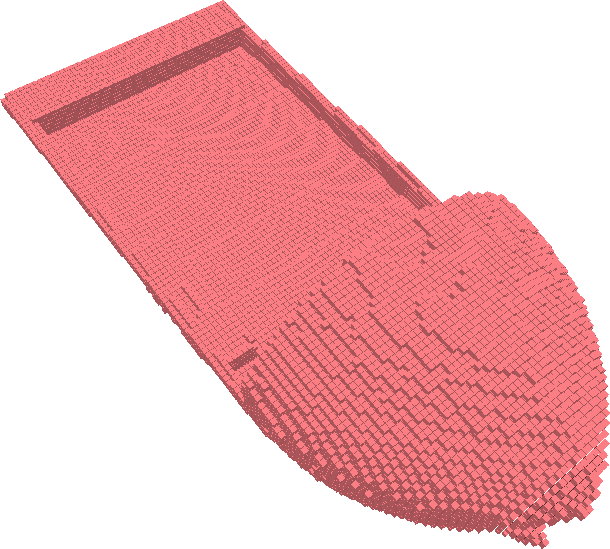}
&\includegraphics[width=0.048\textwidth]{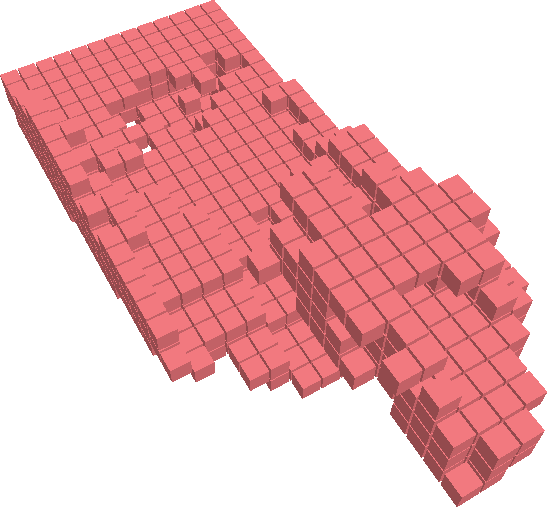}
&\includegraphics[width=0.048\textwidth]{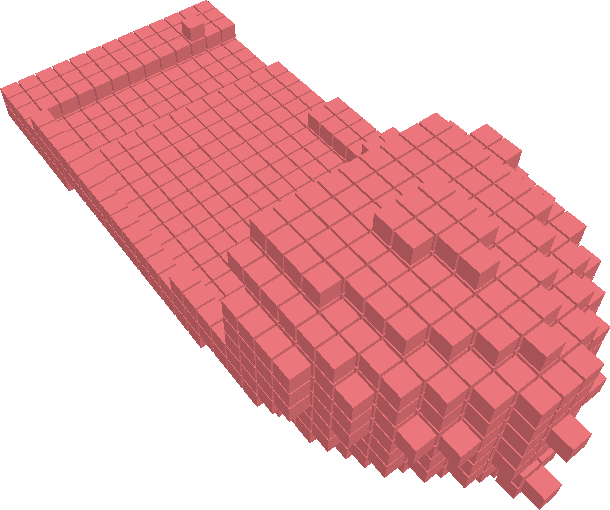}
&\includegraphics[width=0.048\textwidth]{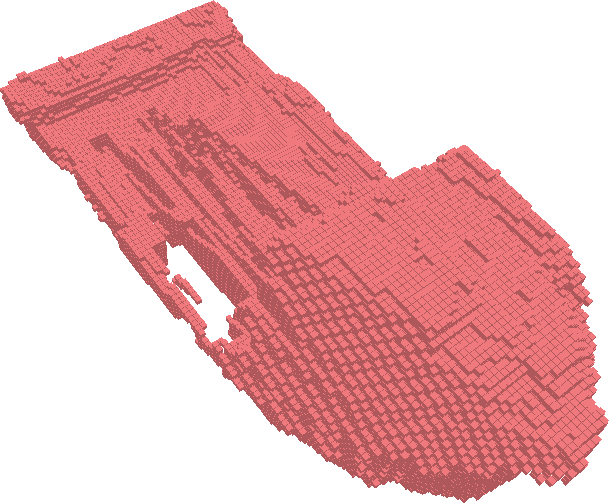}
&\includegraphics[width=0.048\textwidth]{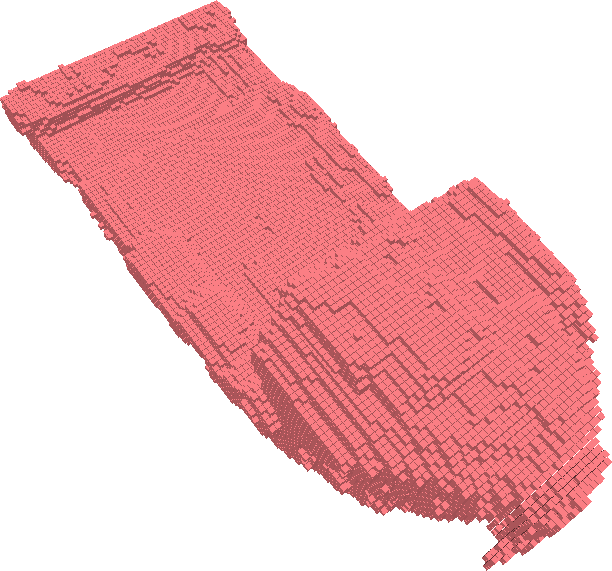}
&\includegraphics[width=0.048\textwidth]{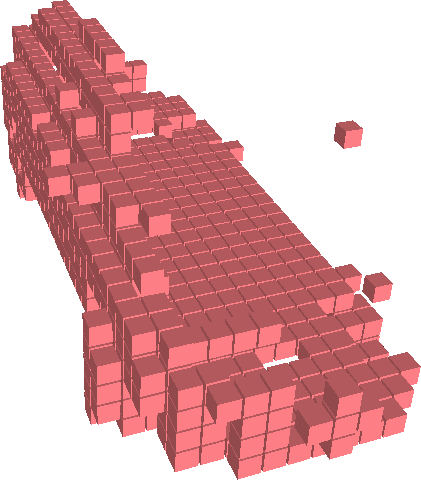}
&\includegraphics[width=0.048\textwidth]{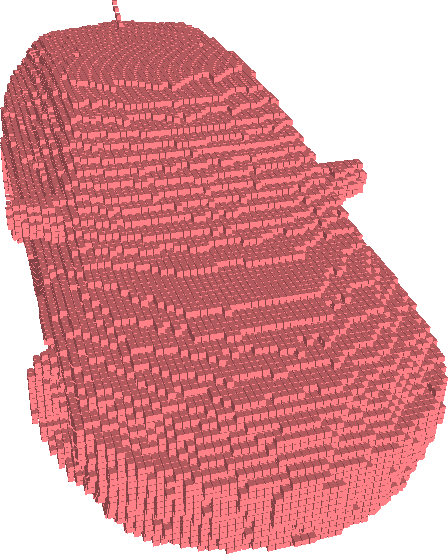}
&\includegraphics[width=0.048\textwidth]{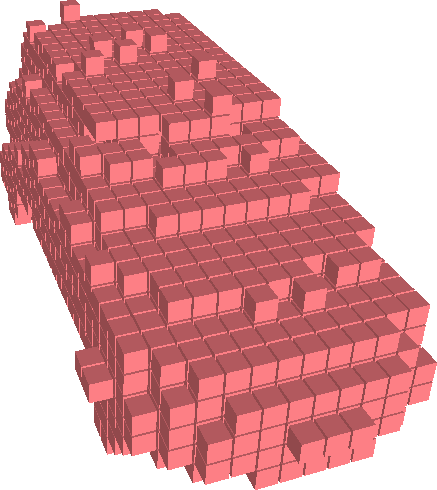}
&\includegraphics[width=0.048\textwidth]{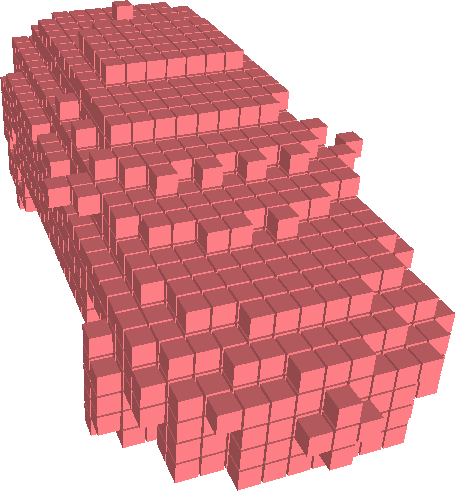}
&\includegraphics[width=0.048\textwidth]{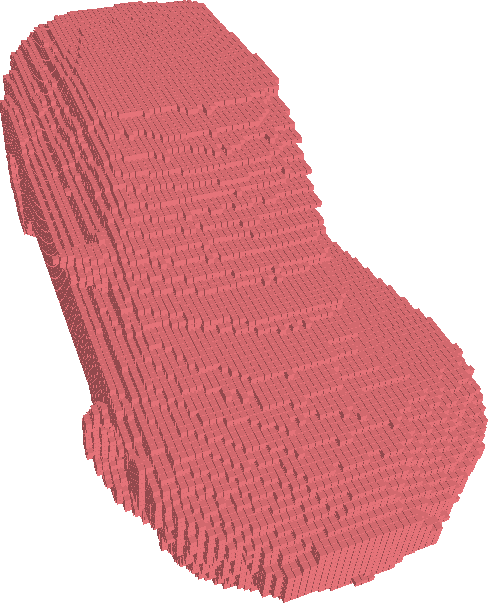}
&\includegraphics[width=0.048\textwidth]{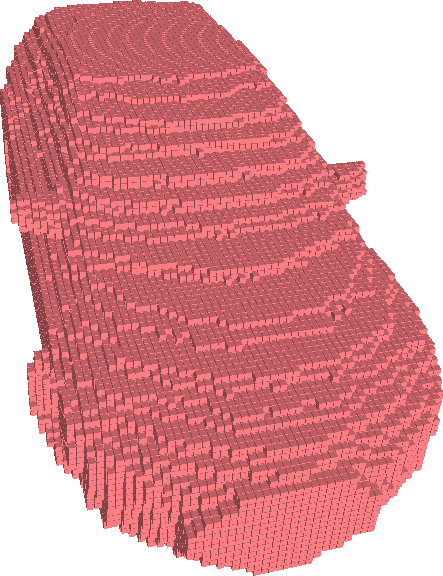}
\\
\includegraphics[width=0.048\textwidth]{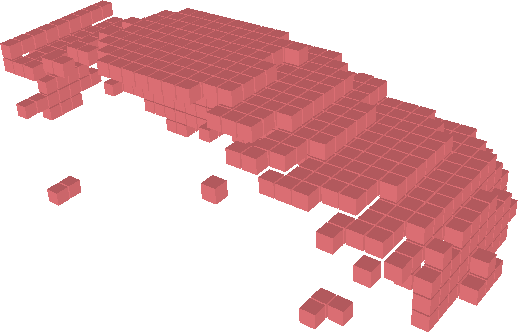}
&\includegraphics[width=0.048\textwidth]{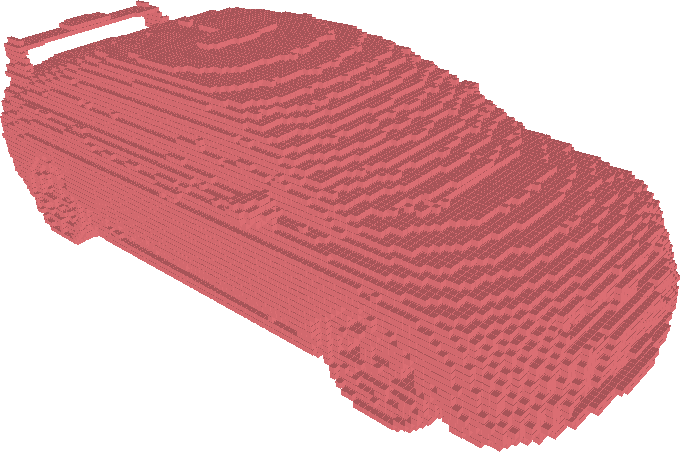}
&\includegraphics[width=0.048\textwidth]{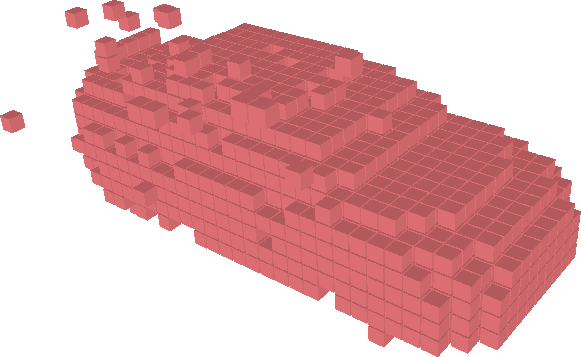}
&\includegraphics[width=0.048\textwidth]{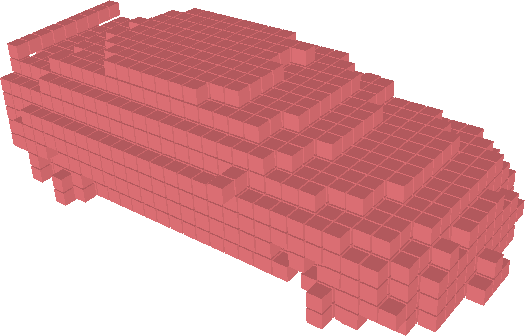}
&\includegraphics[width=0.048\textwidth]{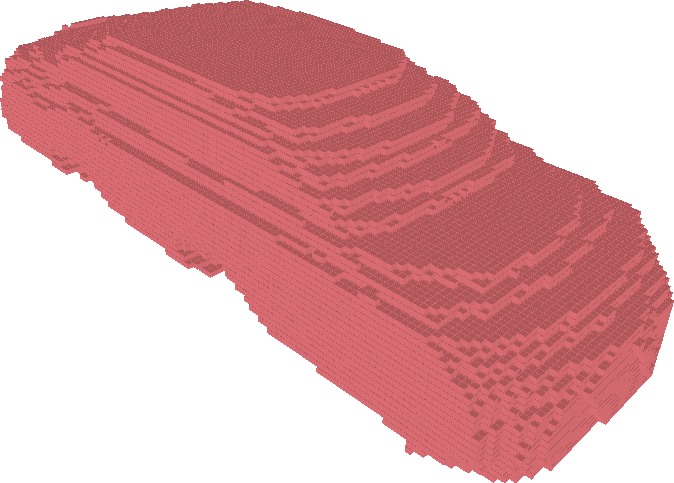}
&\includegraphics[width=0.048\textwidth]{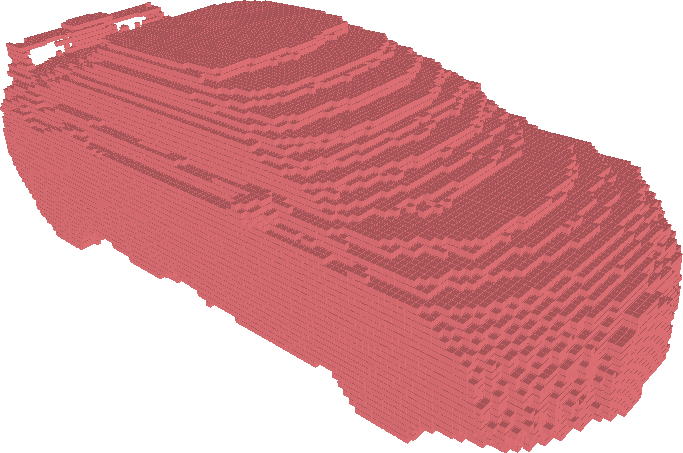}
&\includegraphics[width=0.048\textwidth]{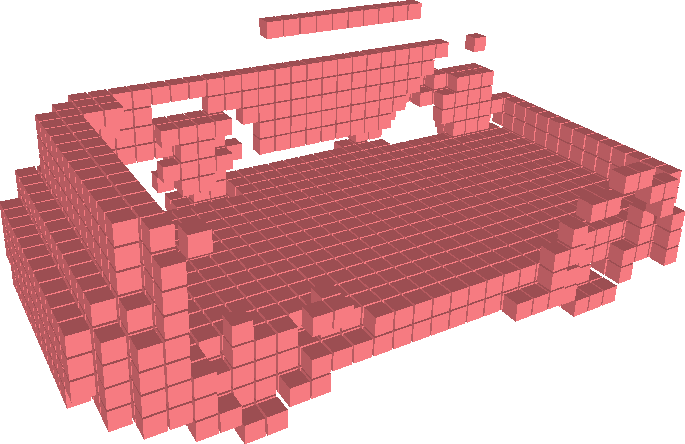}
&\includegraphics[width=0.048\textwidth]{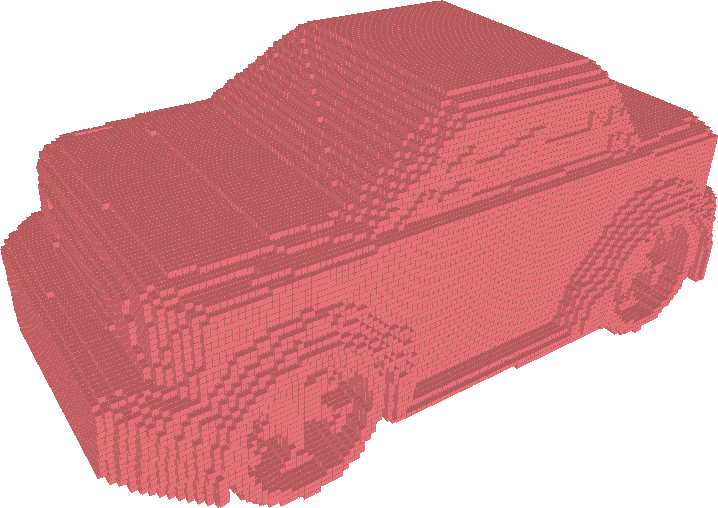}
&\includegraphics[width=0.048\textwidth]{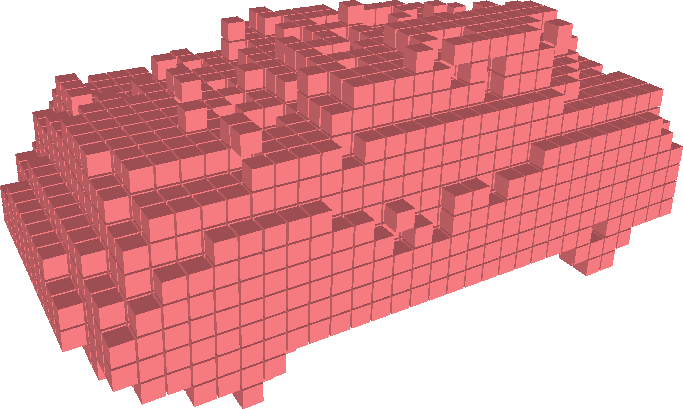}
&\includegraphics[width=0.048\textwidth]{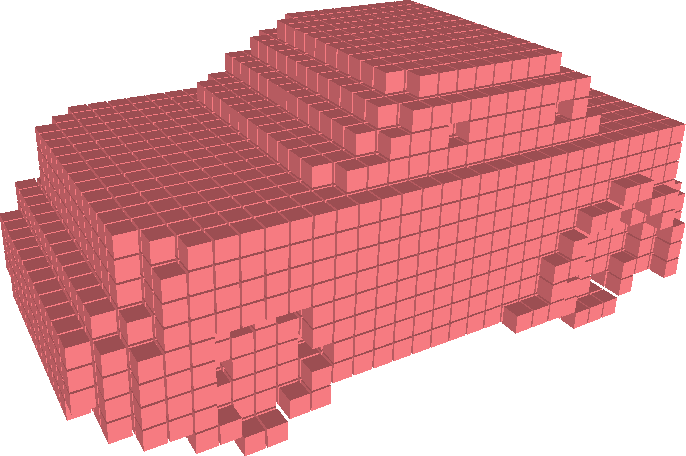}
&\includegraphics[width=0.048\textwidth]{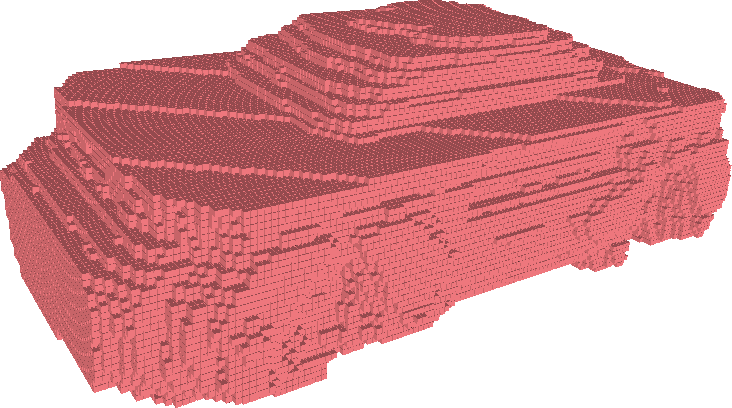}
&\includegraphics[width=0.048\textwidth]{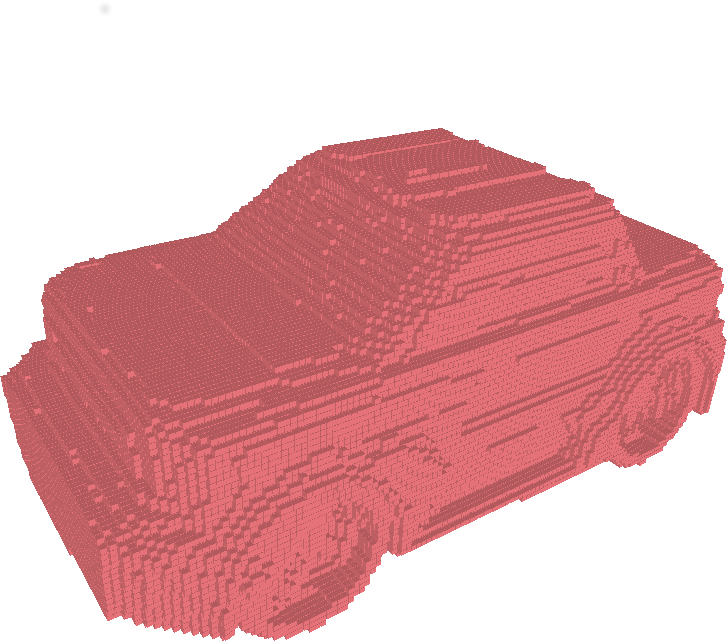}
&\includegraphics[width=0.048\textwidth]{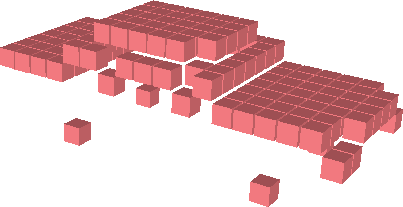}
&\includegraphics[width=0.048\textwidth]{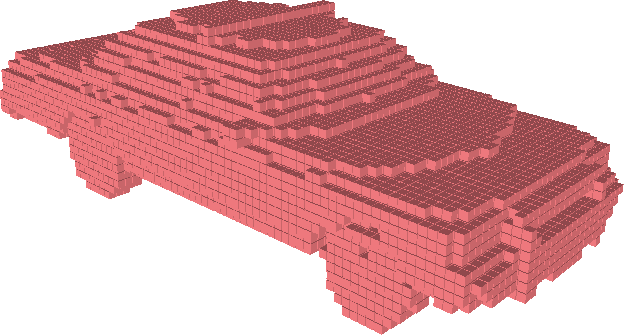}
&\includegraphics[width=0.048\textwidth]{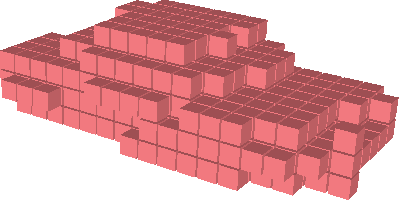}
&\includegraphics[width=0.048\textwidth]{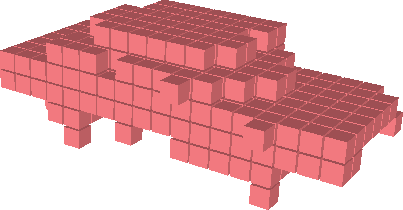}
&\includegraphics[width=0.048\textwidth]{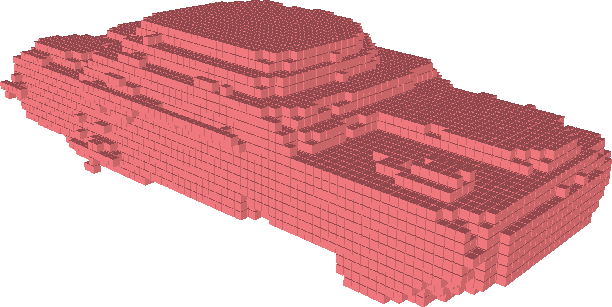}
&\includegraphics[width=0.048\textwidth]{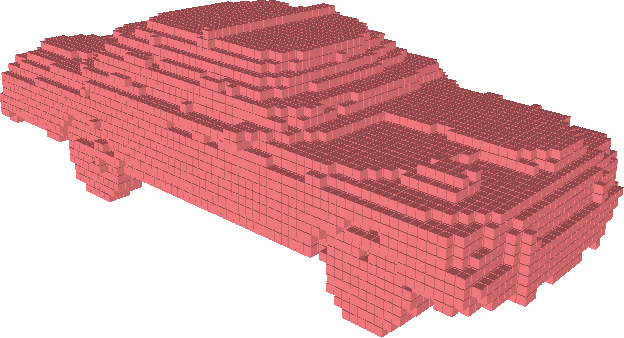}
\\
\includegraphics[width=0.048\textwidth]{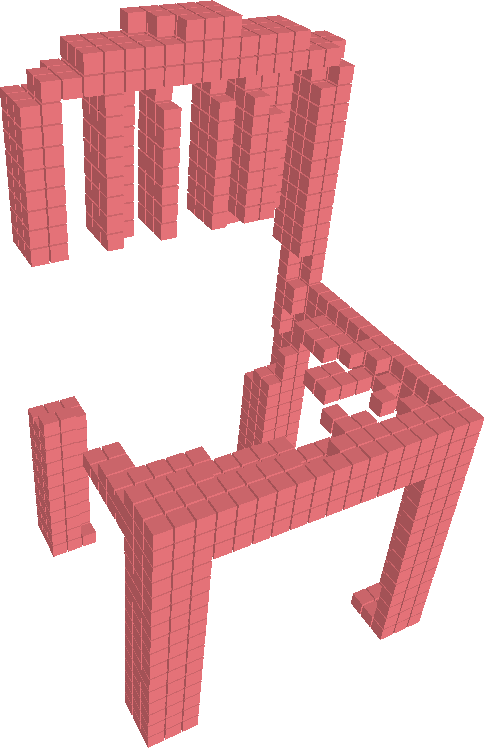}
&\includegraphics[width=0.048\textwidth]{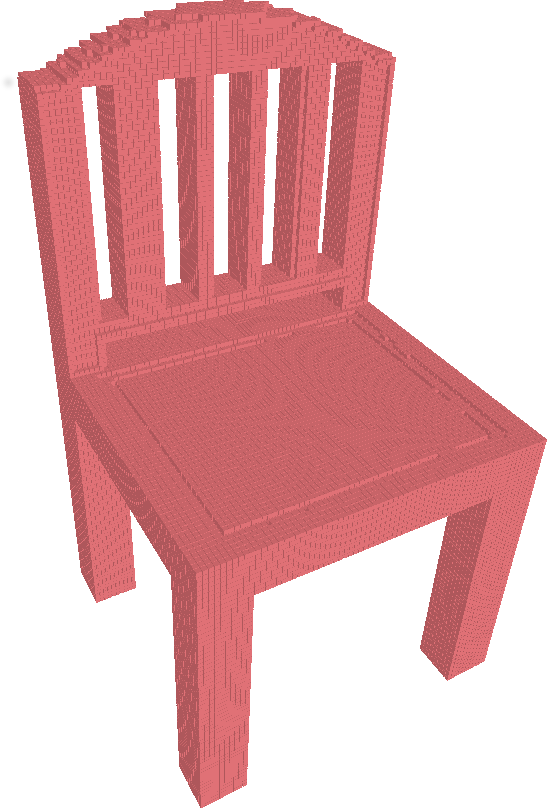}
&\includegraphics[width=0.048\textwidth]{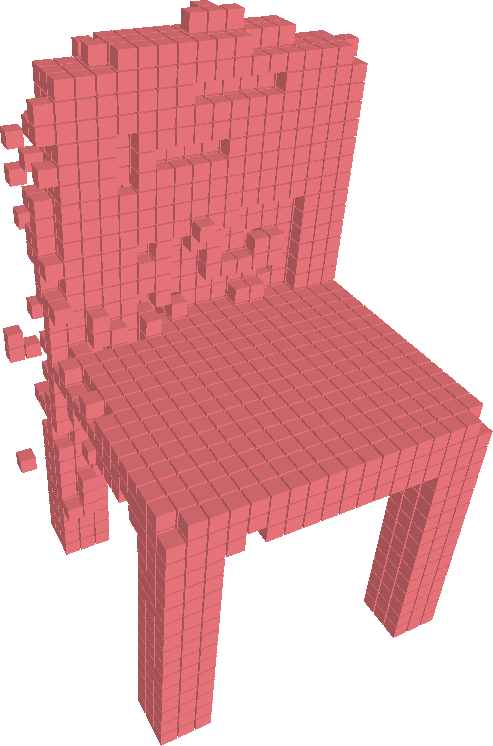}
&\includegraphics[width=0.048\textwidth]{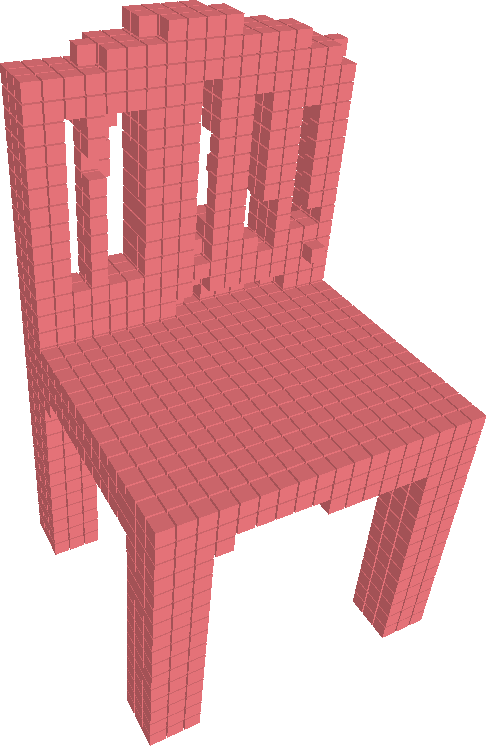}
&\includegraphics[width=0.048\textwidth]{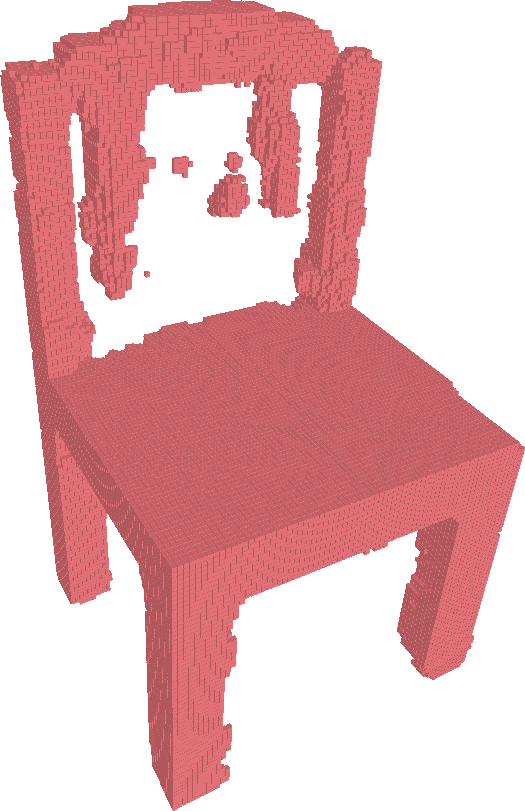}
&\includegraphics[width=0.048\textwidth]{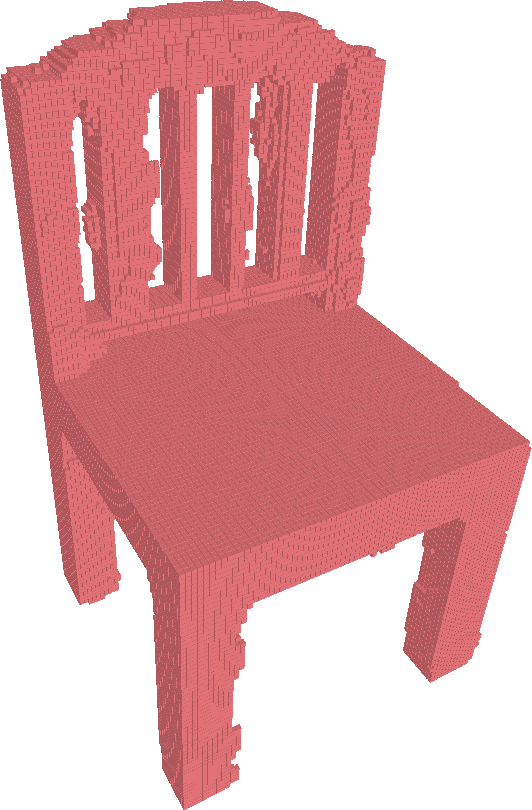}
&\includegraphics[width=0.048\textwidth]{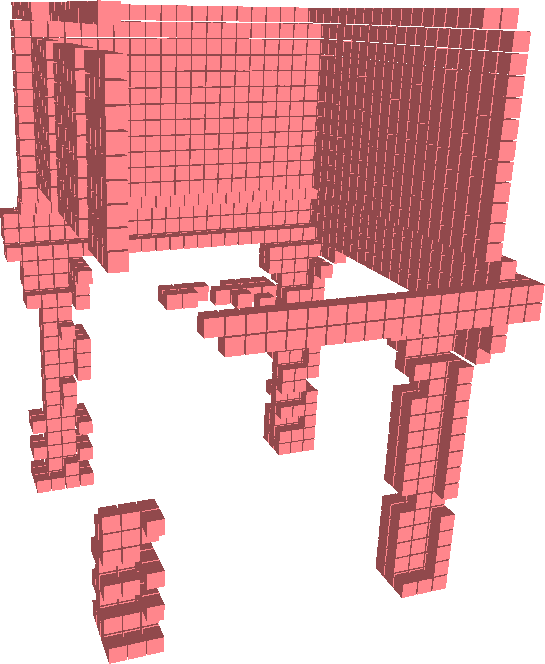}
&\includegraphics[width=0.048\textwidth]{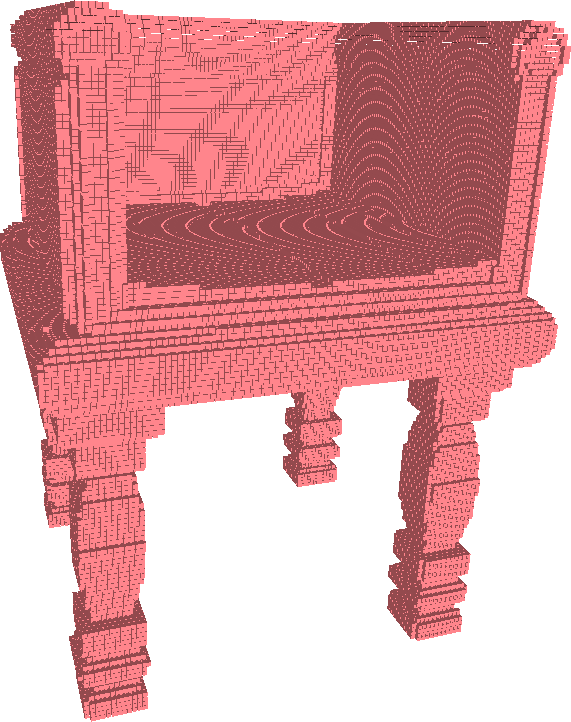}
&\includegraphics[width=0.048\textwidth]{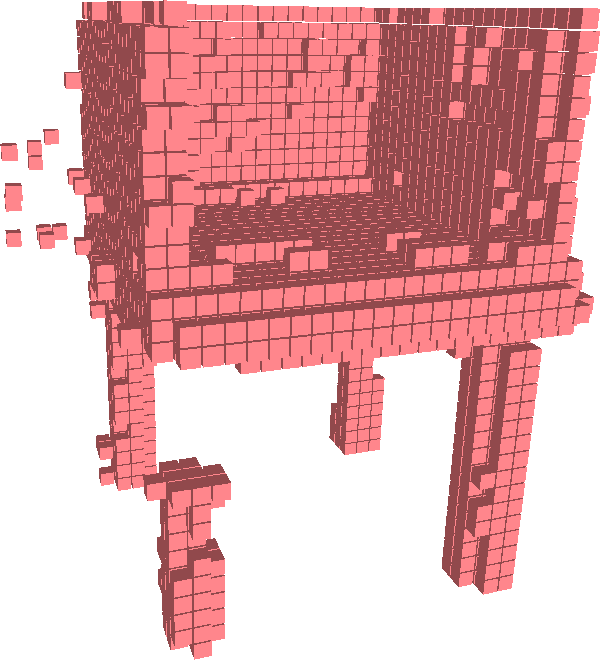}
&\includegraphics[width=0.048\textwidth]{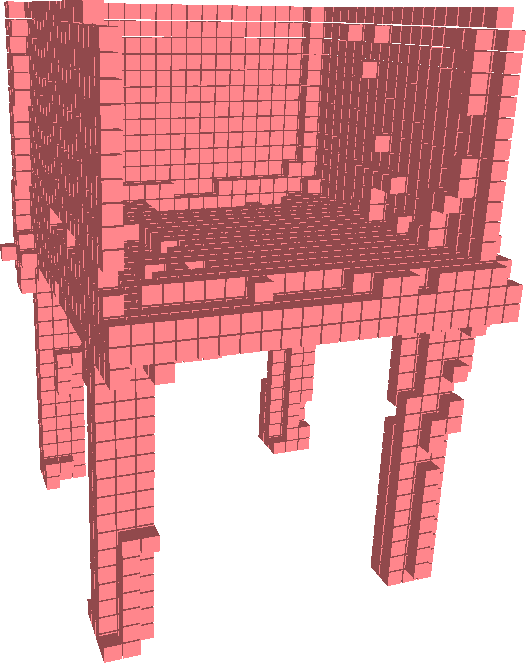}
&\includegraphics[width=0.048\textwidth]{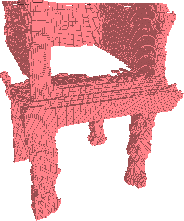}
&\includegraphics[width=0.048\textwidth]{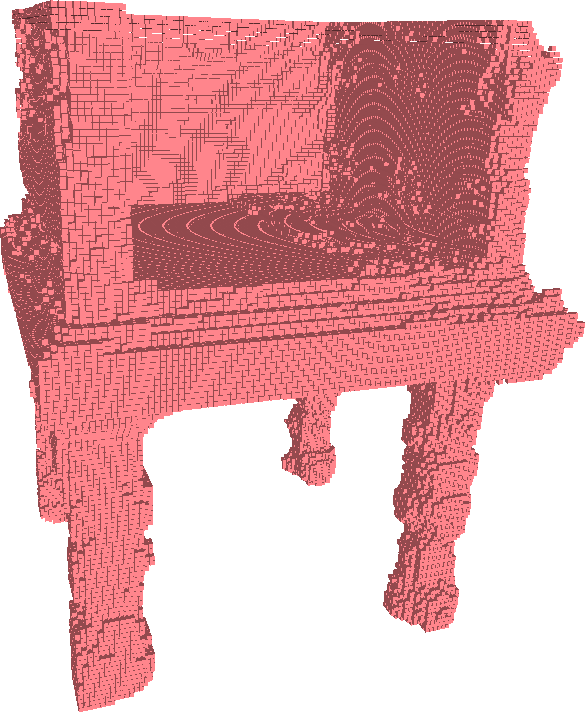}
&\includegraphics[width=0.045\textwidth]{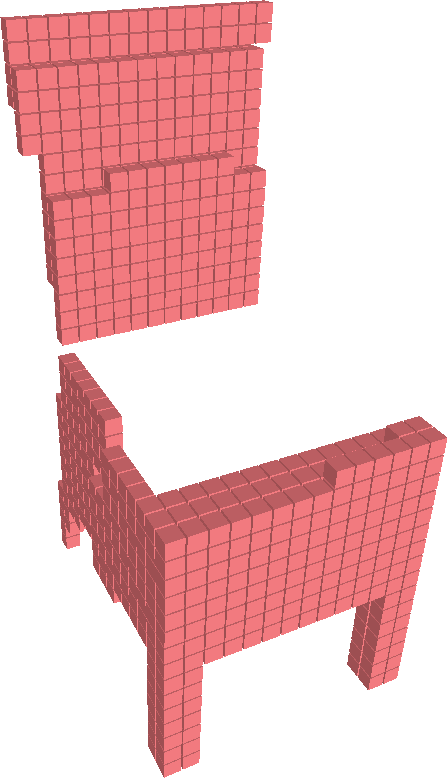}
&\includegraphics[width=0.045\textwidth]{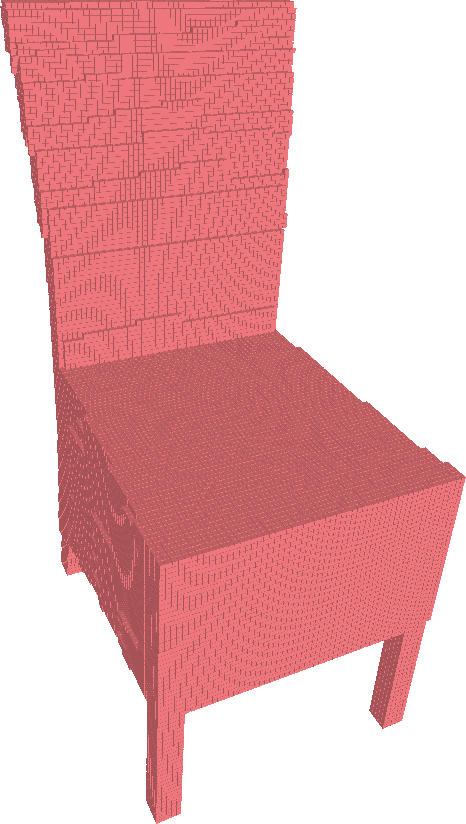}
&\includegraphics[width=0.045\textwidth]{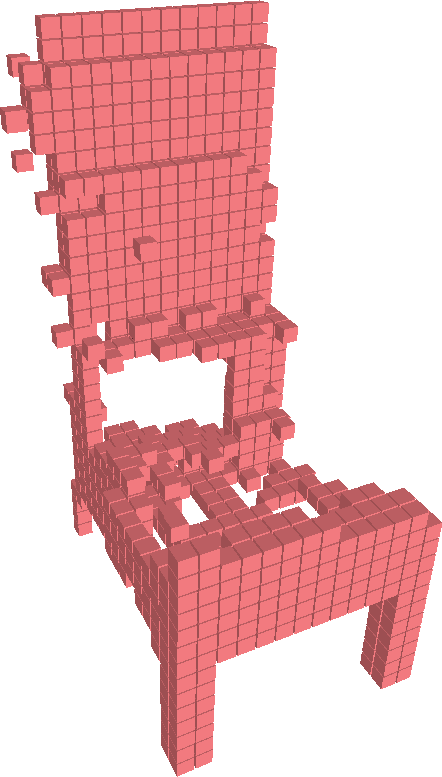}
&\includegraphics[width=0.045\textwidth]{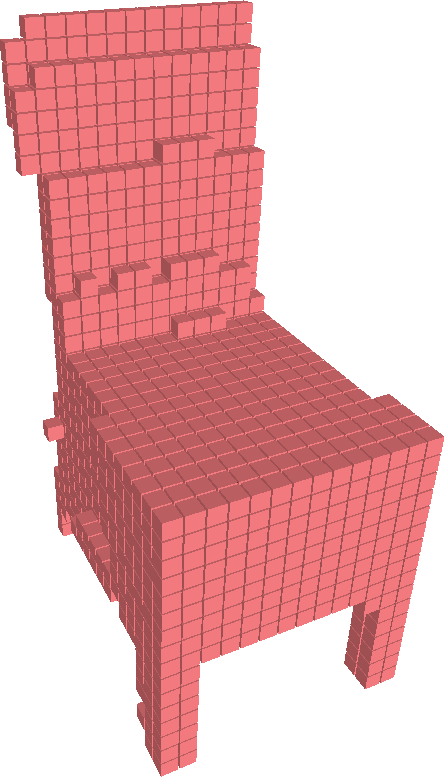}
&\includegraphics[width=0.045\textwidth]{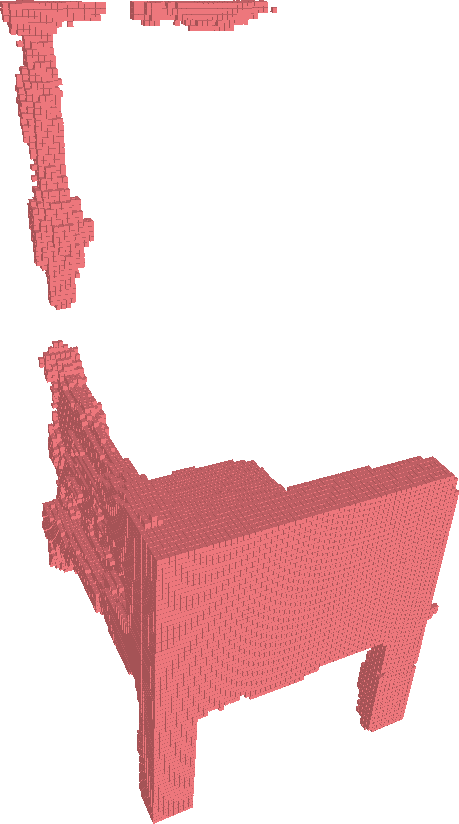}
&\includegraphics[width=0.045\textwidth]{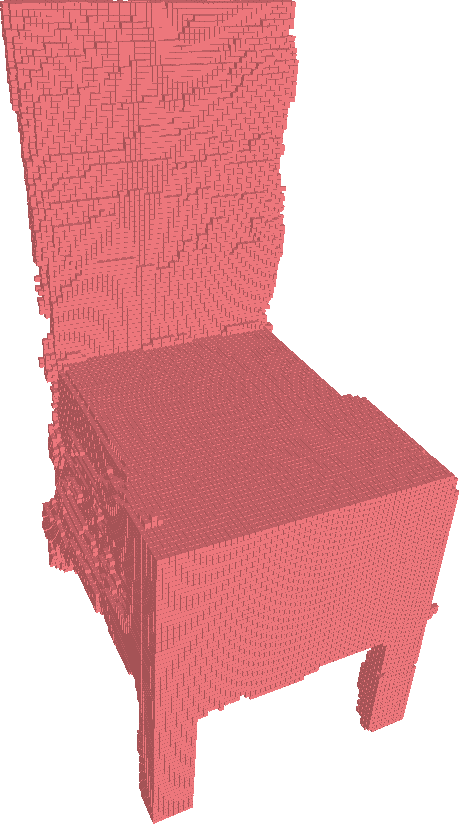}
\\
\includegraphics[width=0.045\textwidth]{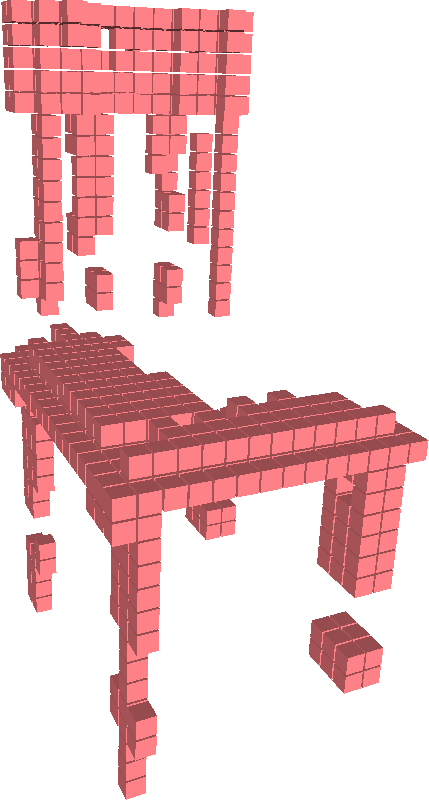}
&\includegraphics[width=0.045\textwidth]{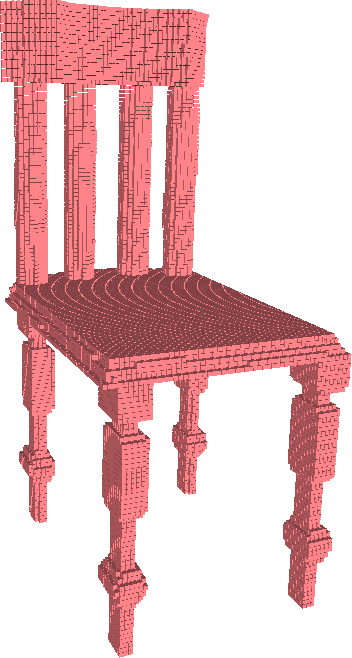}
&\includegraphics[width=0.045\textwidth]{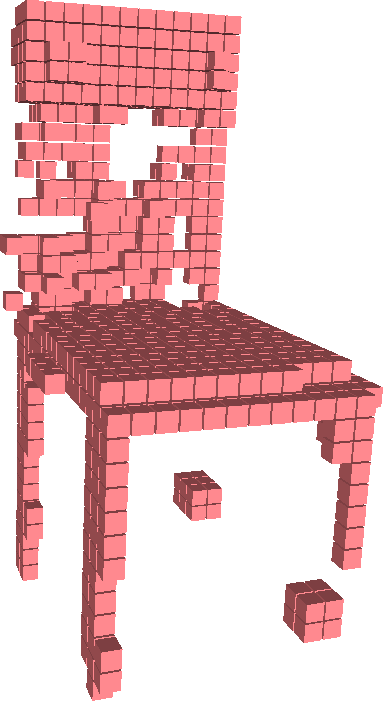}
&\includegraphics[width=0.045\textwidth]{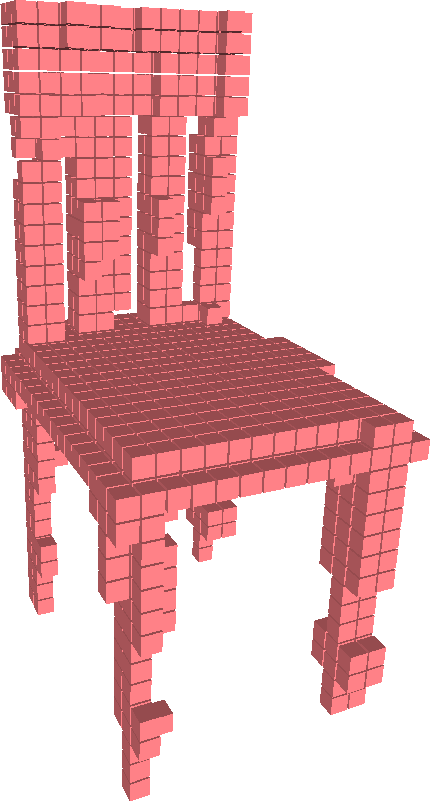}
&\includegraphics[width=0.045\textwidth]{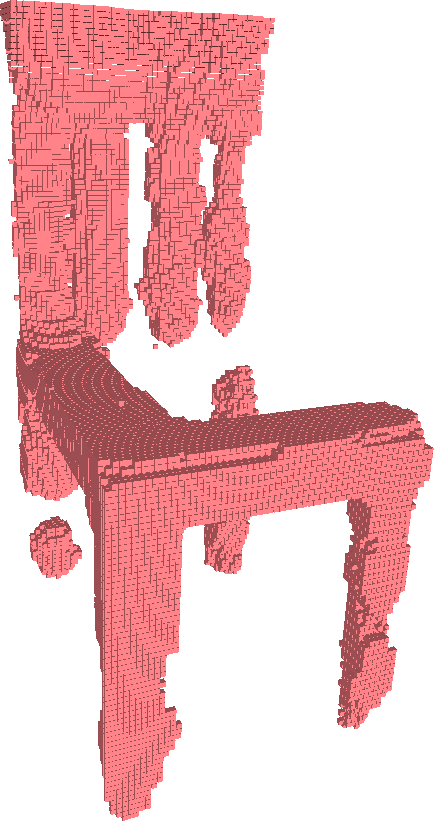}
&\includegraphics[width=0.045\textwidth]{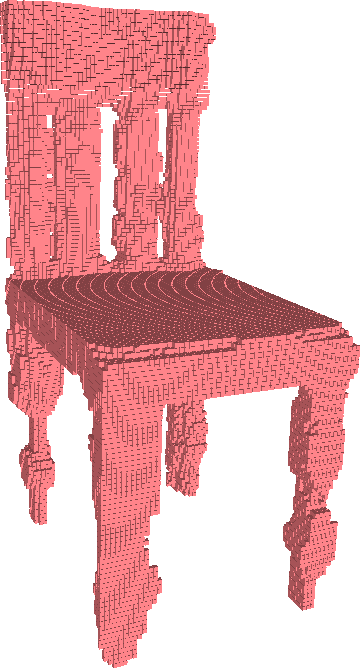}
&\includegraphics[width=0.048\textwidth]{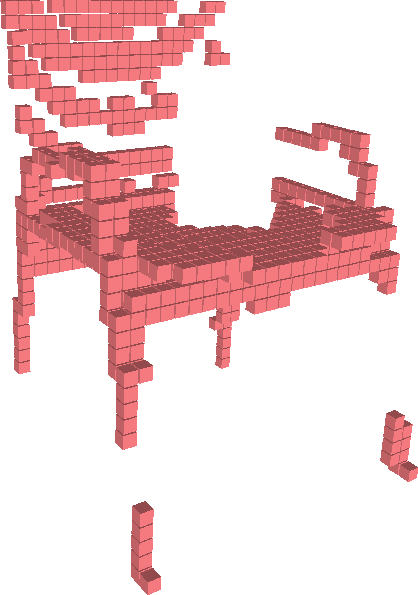}
&\includegraphics[width=0.048\textwidth]{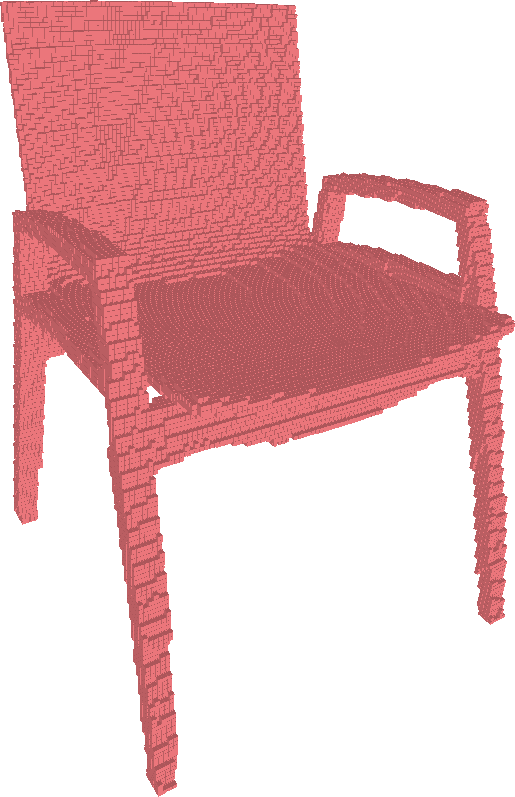}
&\includegraphics[width=0.048\textwidth]{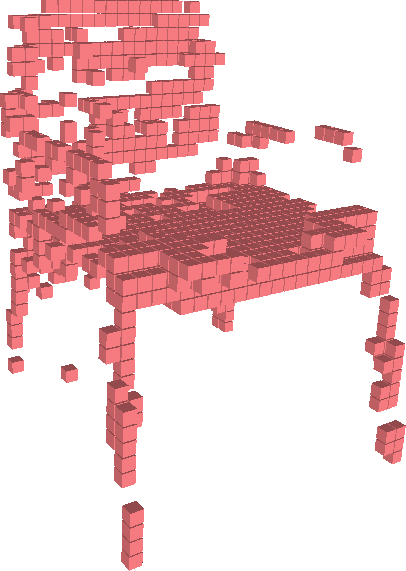}
&\includegraphics[width=0.048\textwidth]{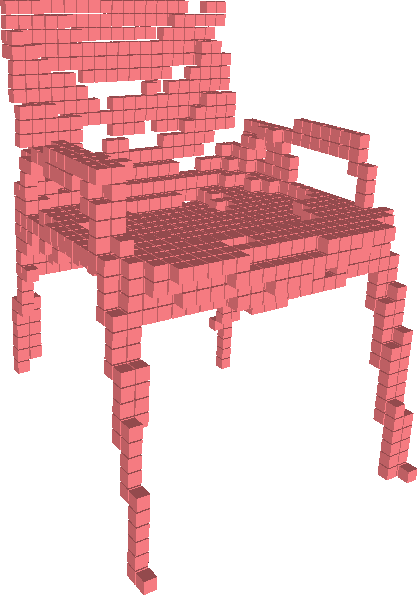}
&\includegraphics[width=0.048\textwidth]{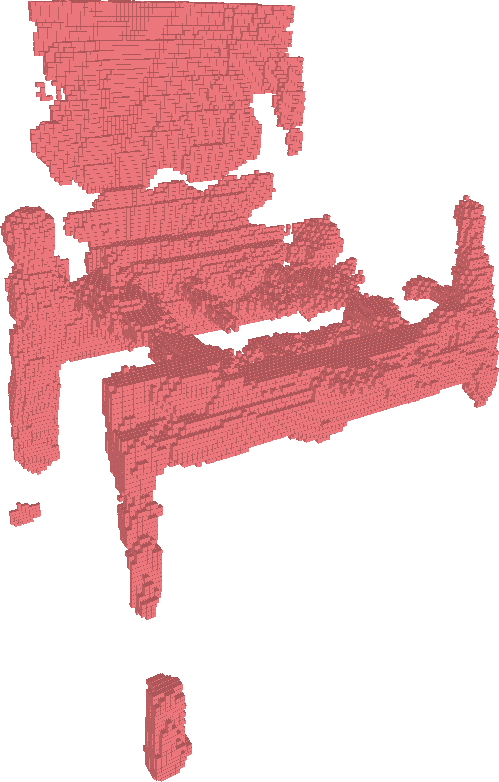}
&\includegraphics[width=0.048\textwidth]{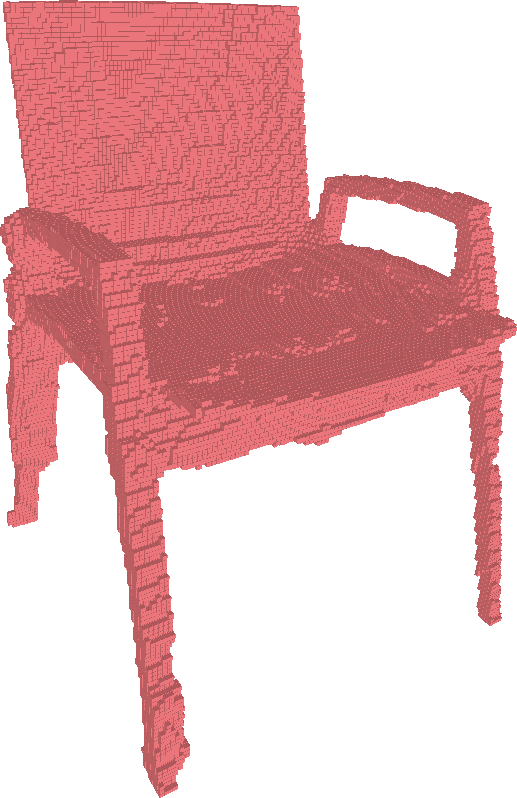}
&\includegraphics[width=0.048\textwidth]{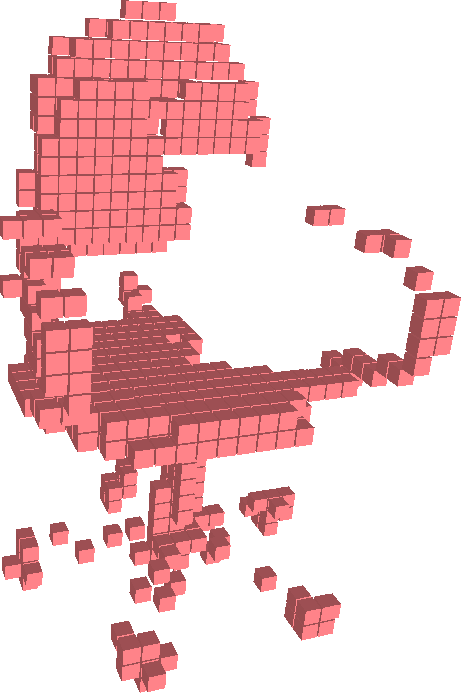}
&\includegraphics[width=0.048\textwidth]{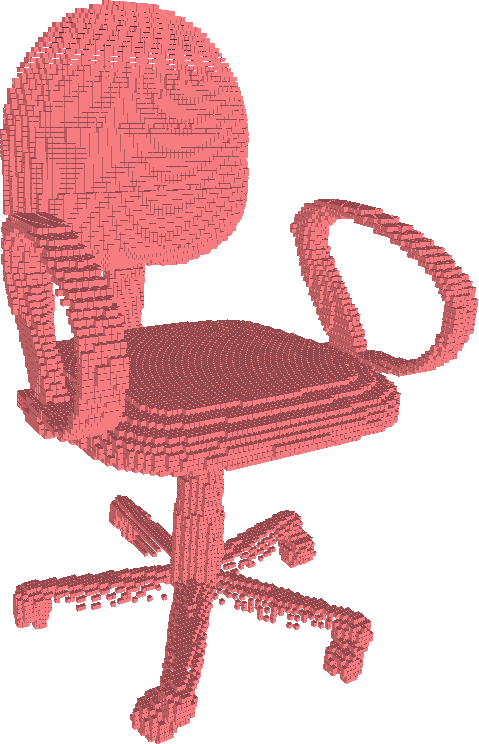}
&\includegraphics[width=0.048\textwidth]{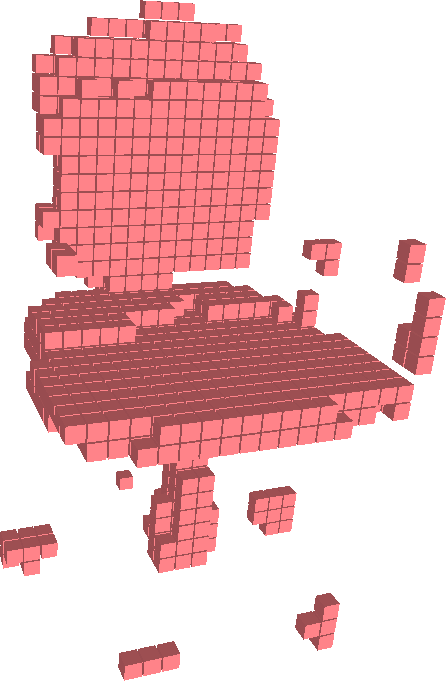}
&\includegraphics[width=0.048\textwidth]{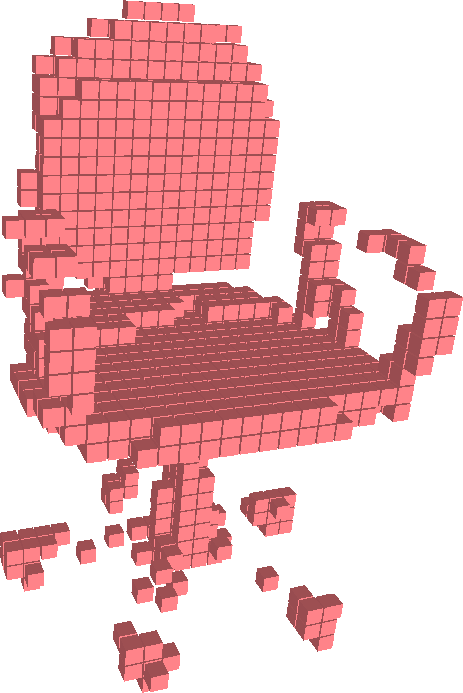}
&\includegraphics[width=0.048\textwidth]{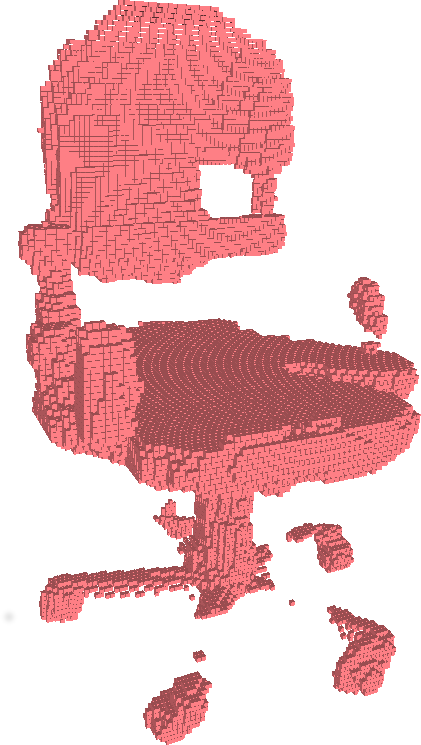}
&\includegraphics[width=0.048\textwidth]{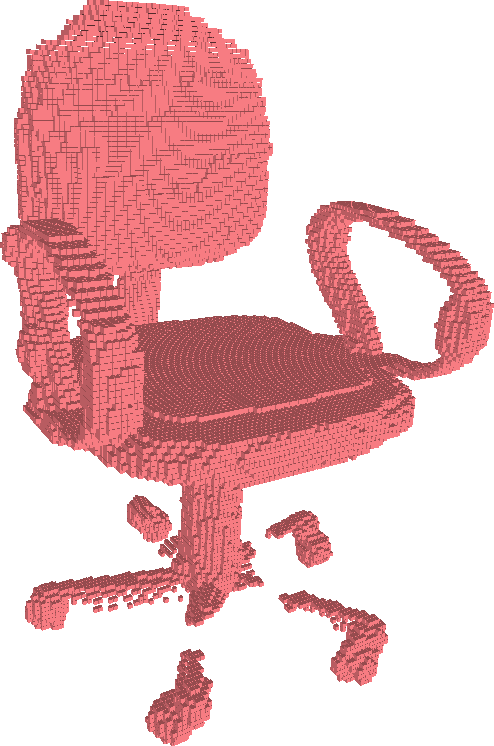}
\\
\includegraphics[width=0.028\textwidth]{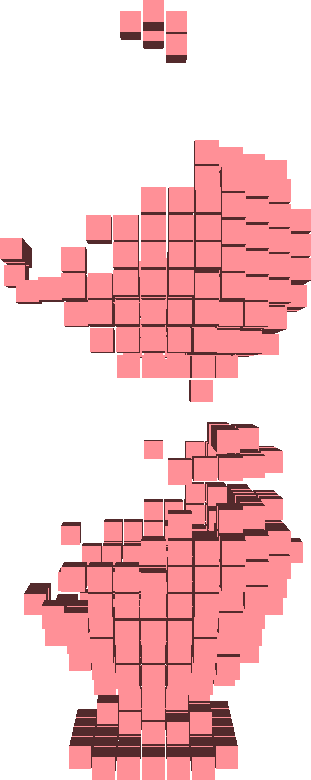}
&\includegraphics[width=0.028\textwidth]{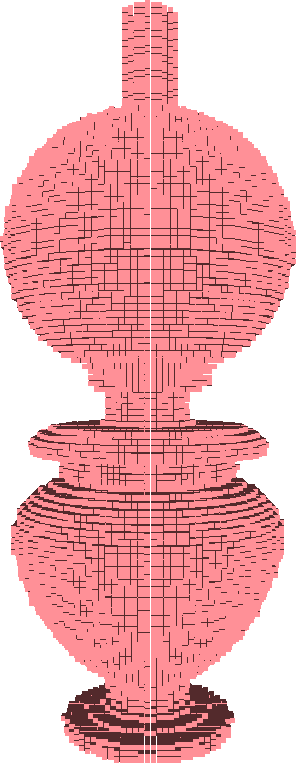}
&\includegraphics[width=0.028\textwidth]{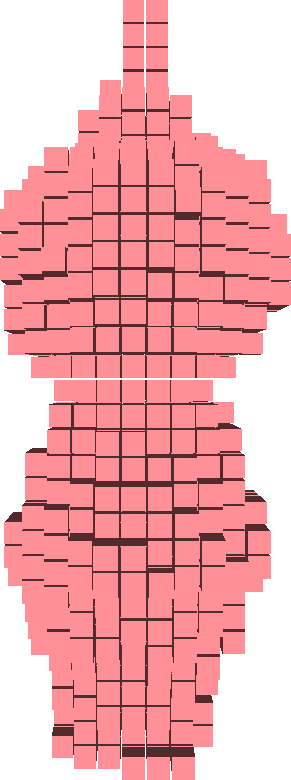}
&\includegraphics[width=0.028\textwidth]{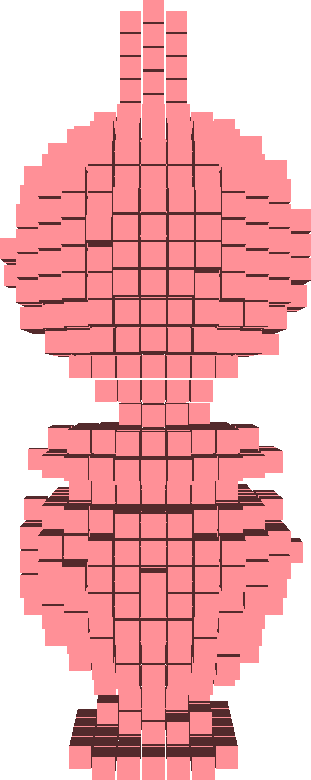}
&\includegraphics[width=0.028\textwidth]{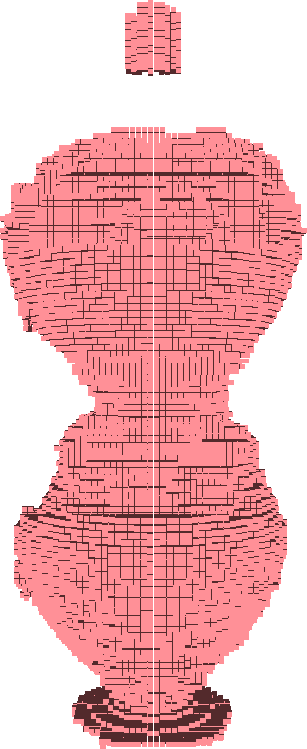}
&\includegraphics[width=0.028\textwidth]{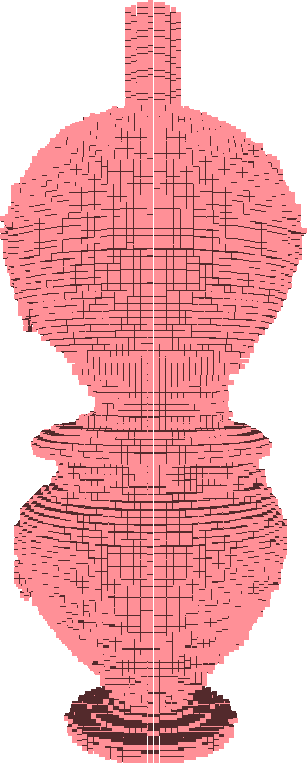}
&\includegraphics[width=0.048\textwidth]{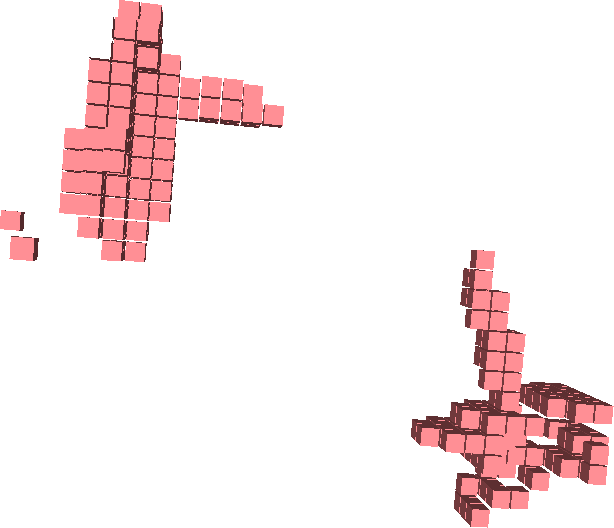}
&\includegraphics[width=0.048\textwidth]{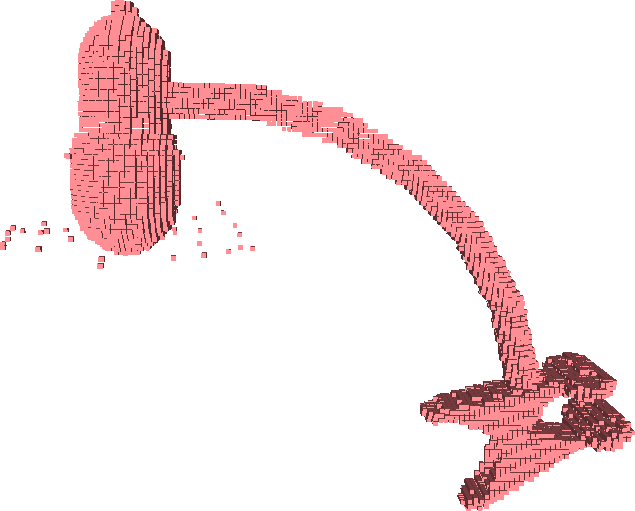}
&\includegraphics[width=0.048\textwidth]{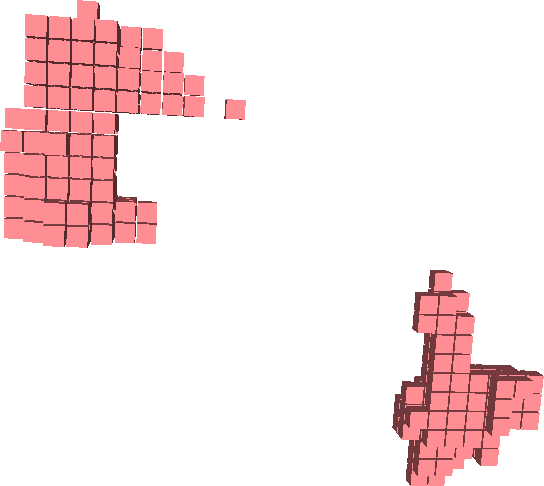}
&\includegraphics[width=0.048\textwidth]{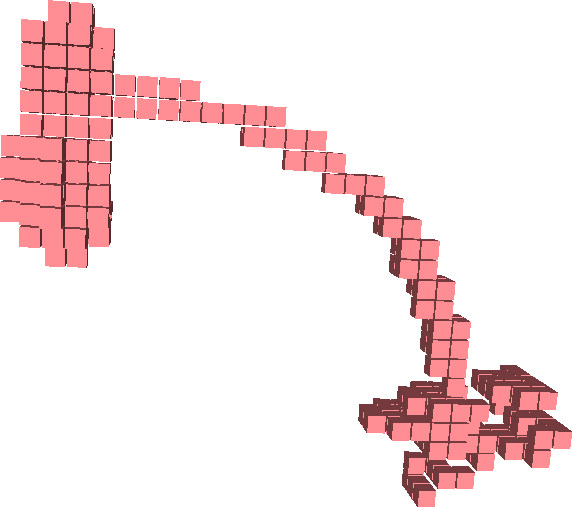}
&\includegraphics[width=0.048\textwidth]{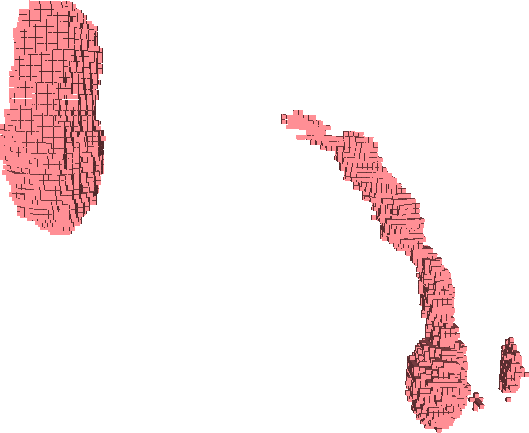}
&\includegraphics[width=0.048\textwidth]{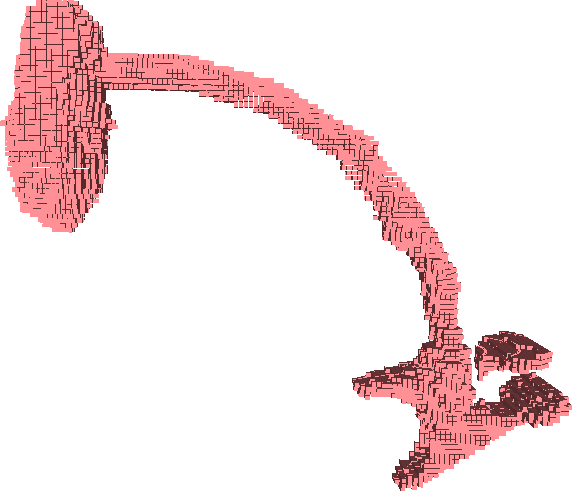}
&\includegraphics[width=0.048\textwidth]{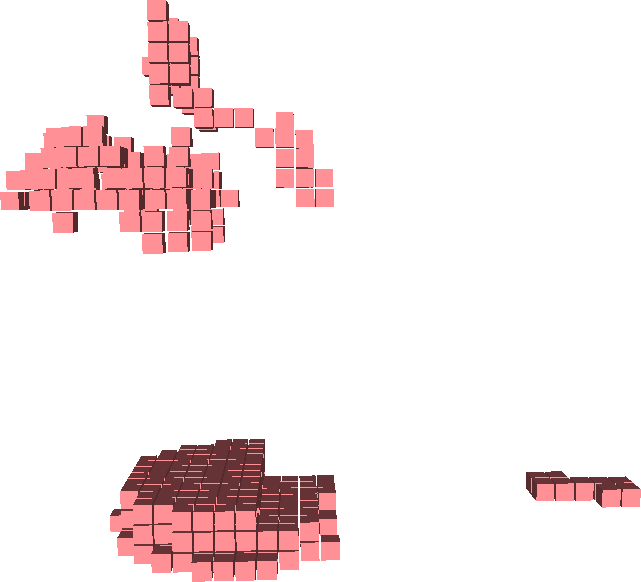}
&\includegraphics[width=0.048\textwidth]{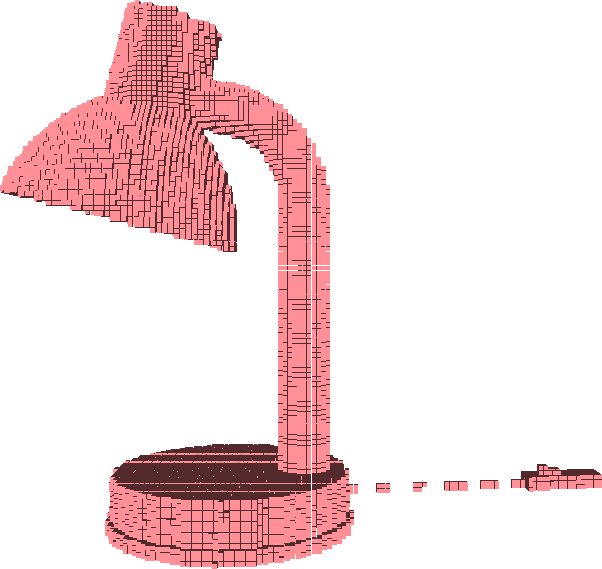}
&\includegraphics[width=0.031\textwidth]{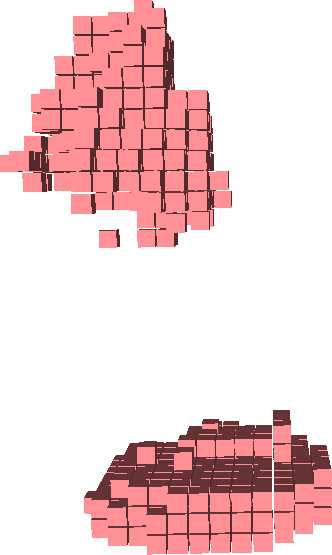}
&\includegraphics[width=0.048\textwidth]{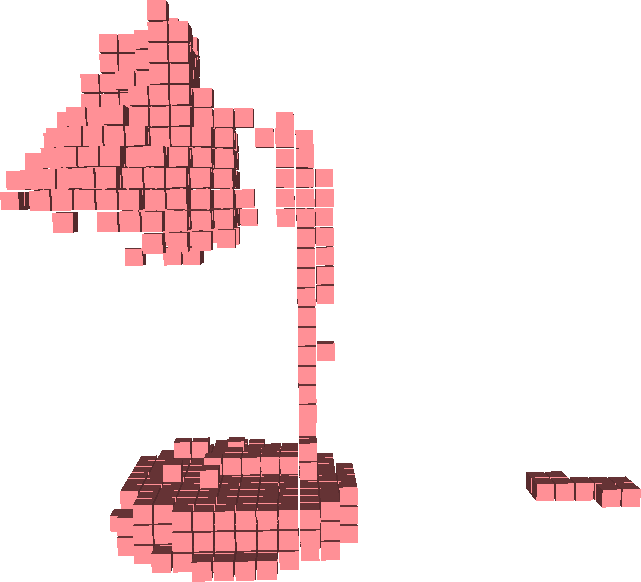}
&\includegraphics[width=0.031\textwidth]{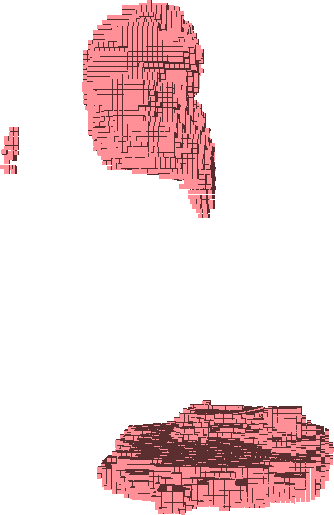}
&\includegraphics[width=0.048\textwidth]{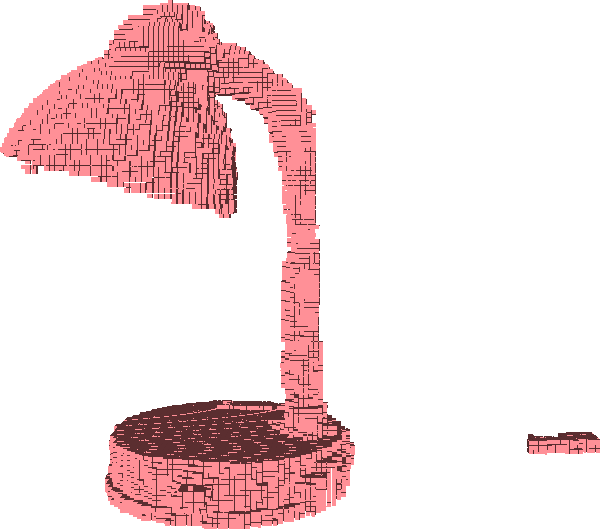}
\\
\includegraphics[width=0.048\textwidth]{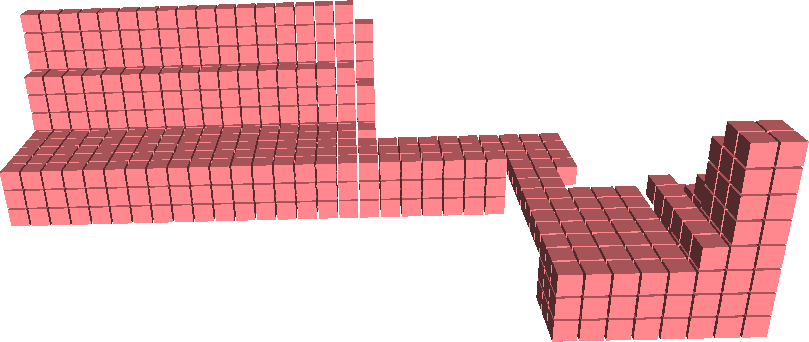}
&\includegraphics[width=0.048\textwidth]{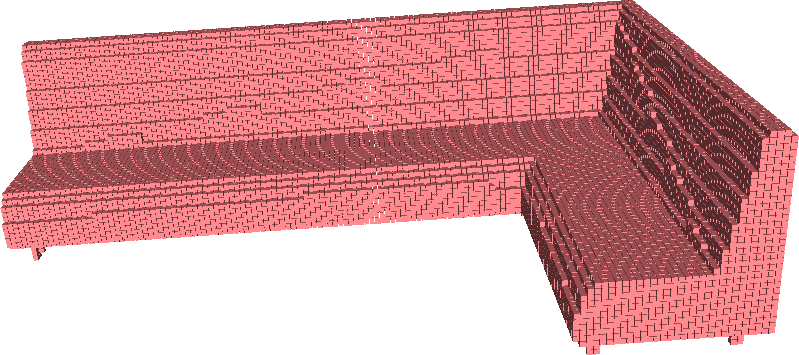}
&\includegraphics[width=0.048\textwidth]{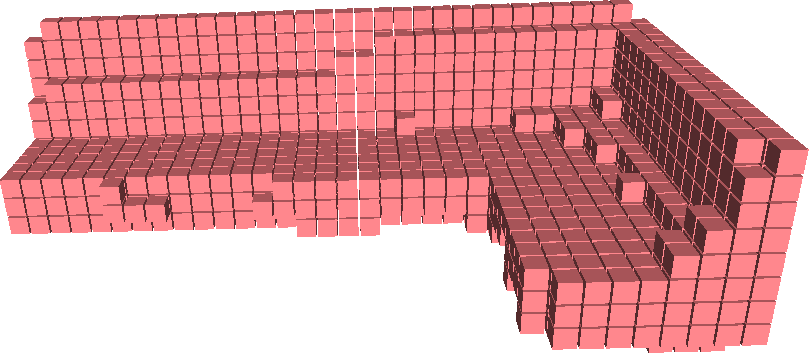}
&\includegraphics[width=0.048\textwidth]{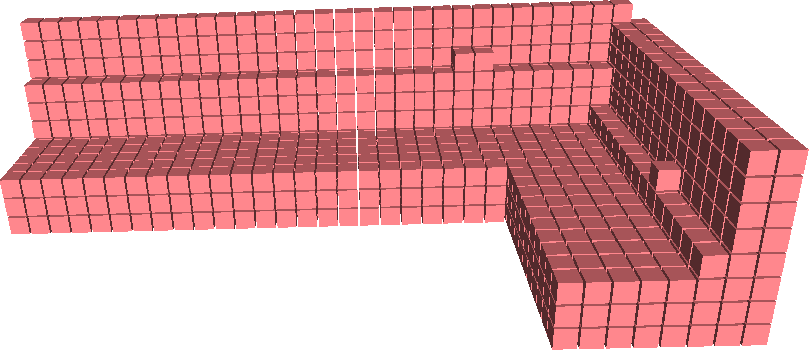}
&\includegraphics[width=0.048\textwidth]{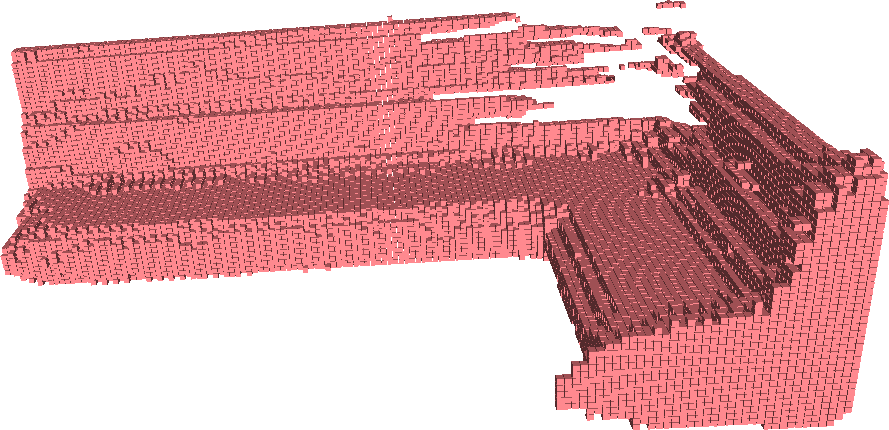}
&\includegraphics[width=0.048\textwidth]{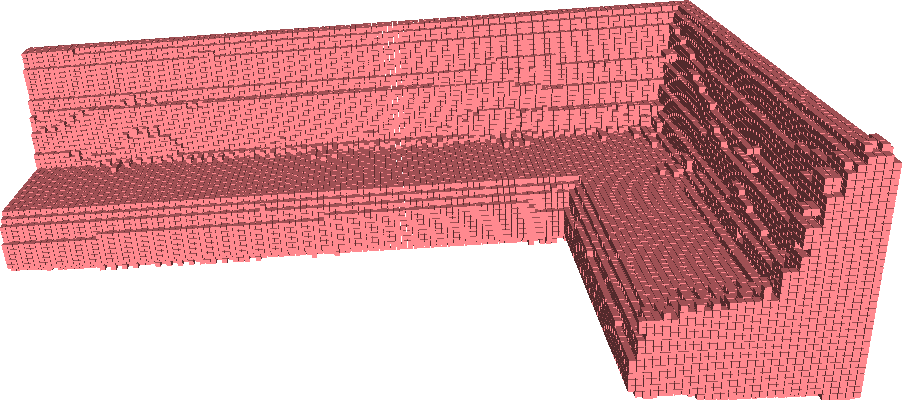}
&\includegraphics[width=0.048\textwidth]{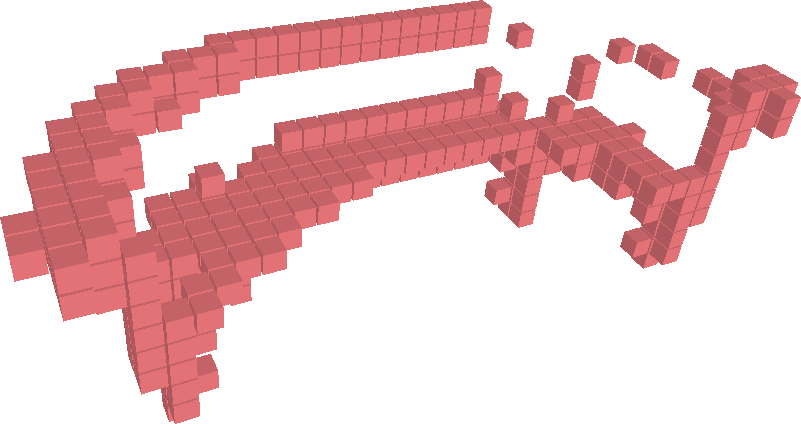}
&\includegraphics[width=0.048\textwidth]{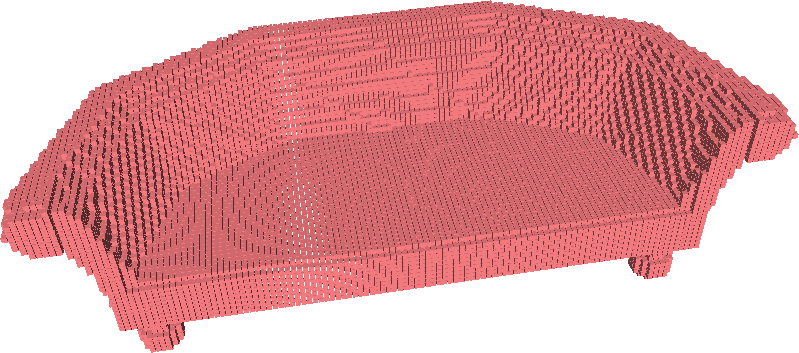}
&\includegraphics[width=0.048\textwidth]{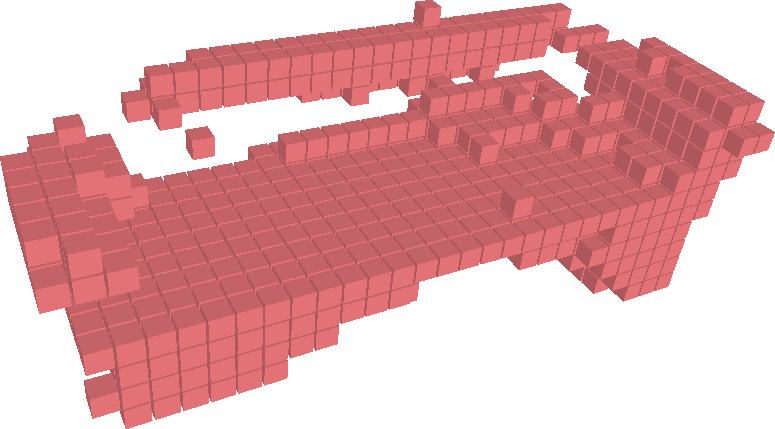}
&\includegraphics[width=0.048\textwidth]{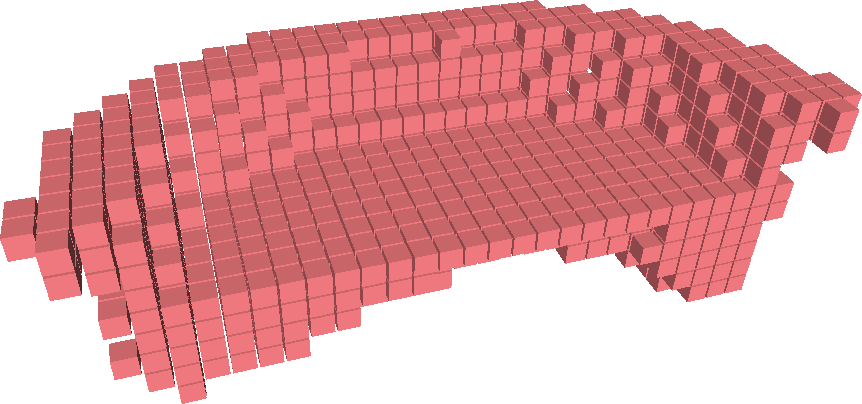}
&\includegraphics[width=0.048\textwidth]{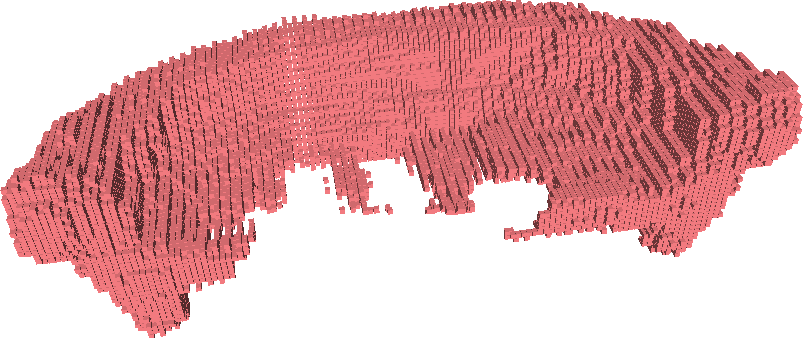}
&\includegraphics[width=0.048\textwidth]{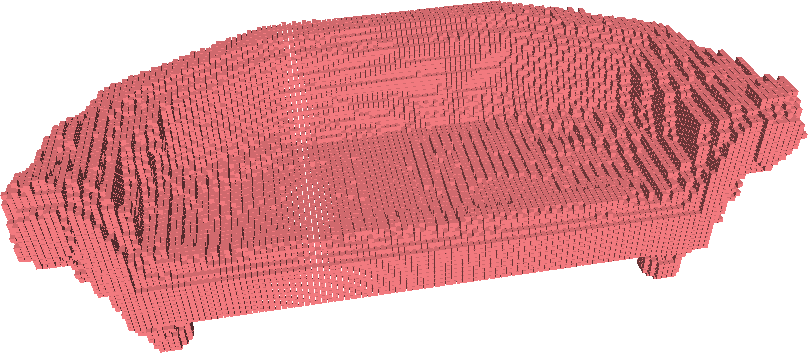}
&\includegraphics[width=0.048\textwidth]{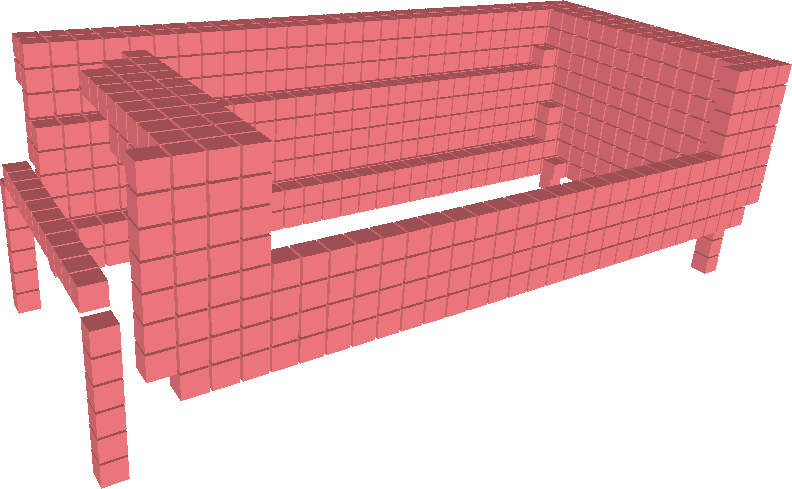}
&\includegraphics[width=0.048\textwidth]{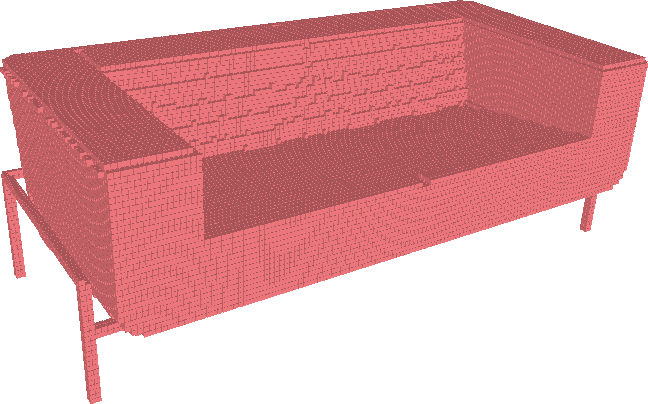}
&\includegraphics[width=0.048\textwidth]{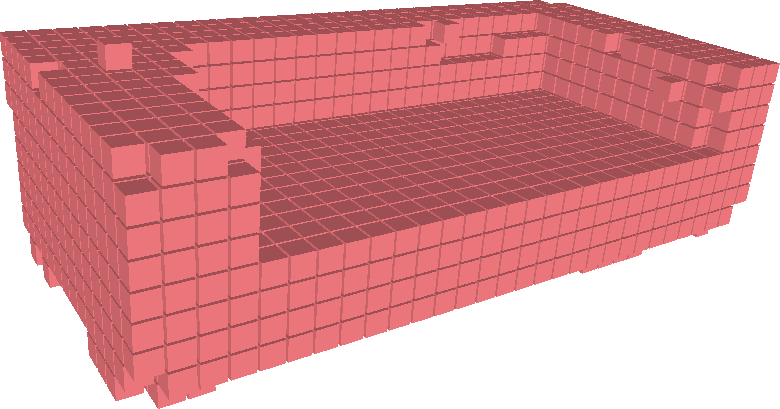}
&\includegraphics[width=0.048\textwidth]{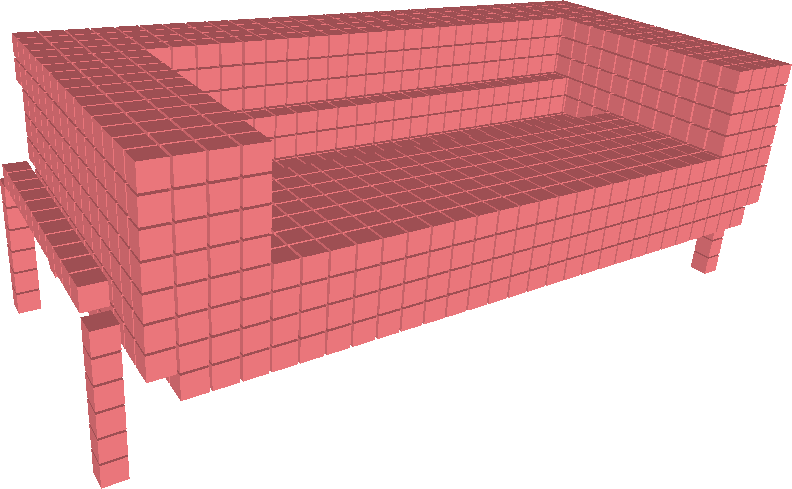}
&\includegraphics[width=0.048\textwidth]{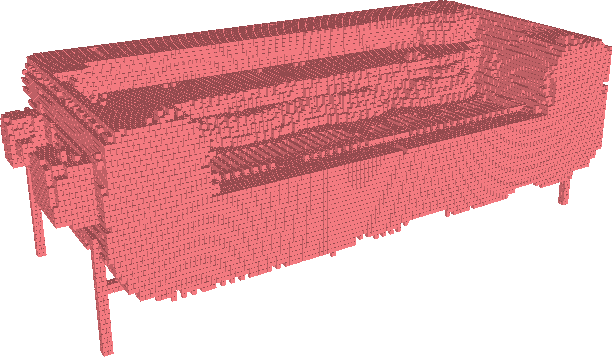}
&\includegraphics[width=0.048\textwidth]{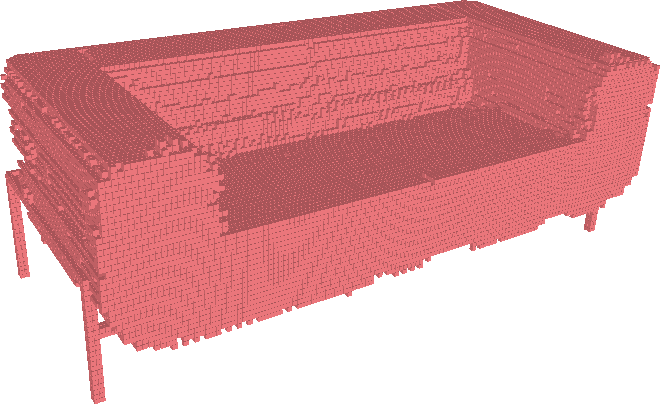}
\\
\includegraphics[width=0.048\textwidth]{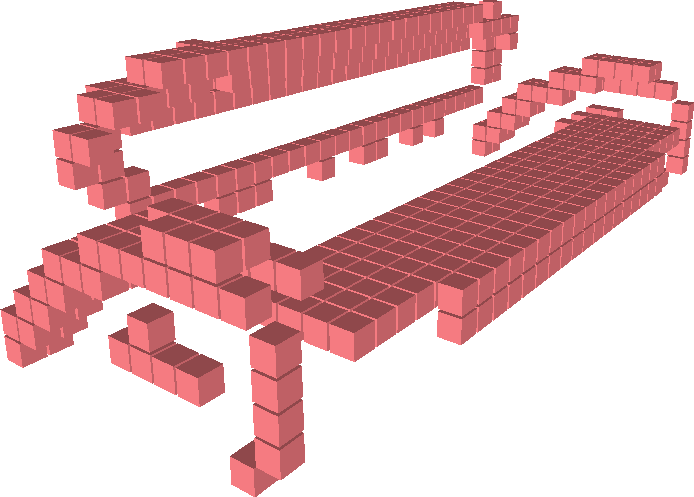}
&\includegraphics[width=0.048\textwidth]{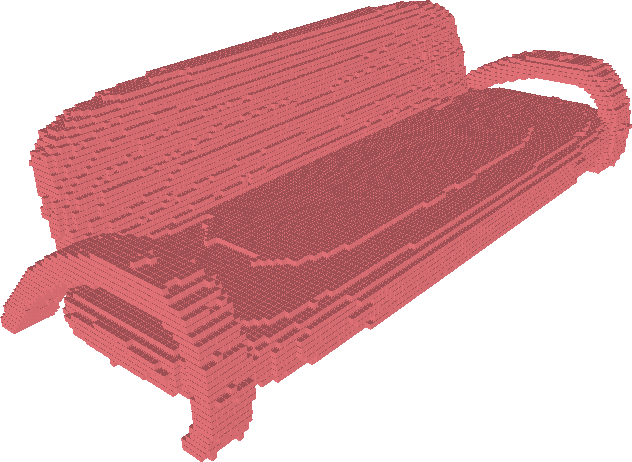}
&\includegraphics[width=0.048\textwidth]{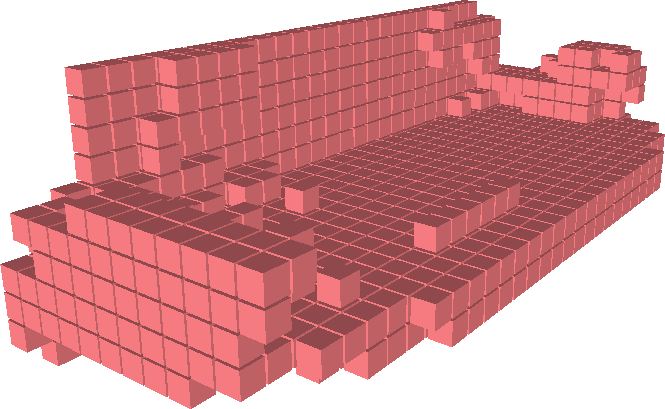}
&\includegraphics[width=0.048\textwidth]{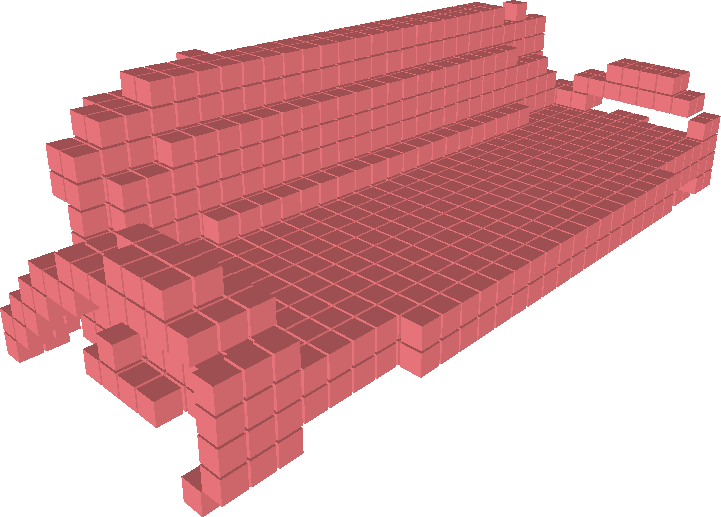}
&\includegraphics[width=0.048\textwidth]{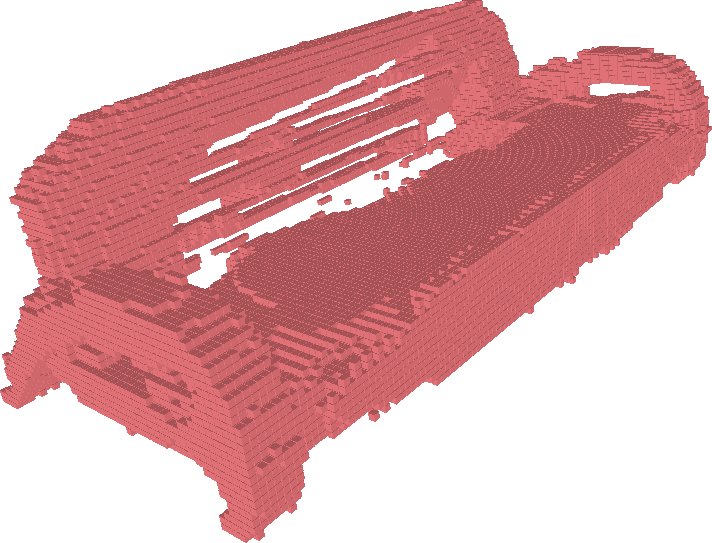}
&\includegraphics[width=0.048\textwidth]{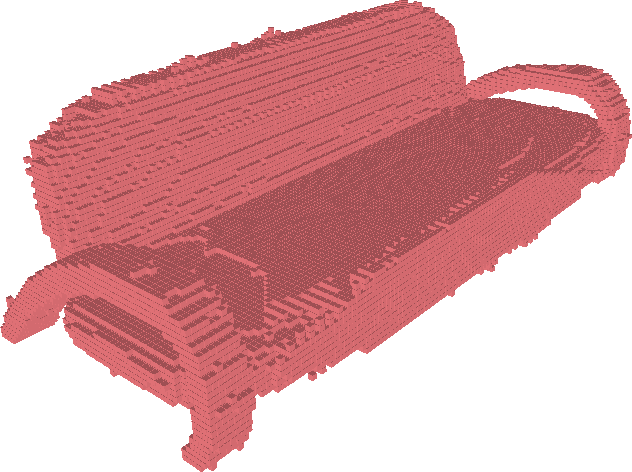}
&\includegraphics[width=0.048\textwidth]{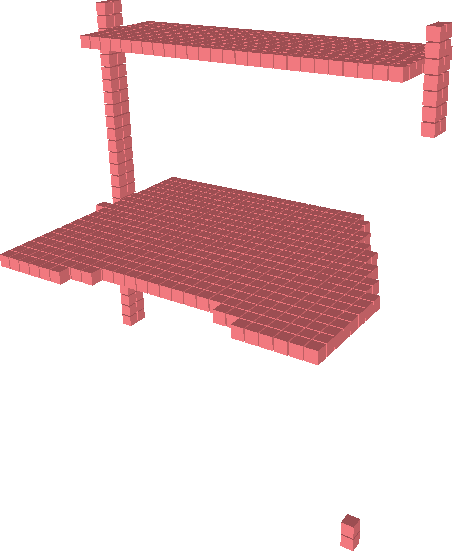}
&\includegraphics[width=0.048\textwidth]{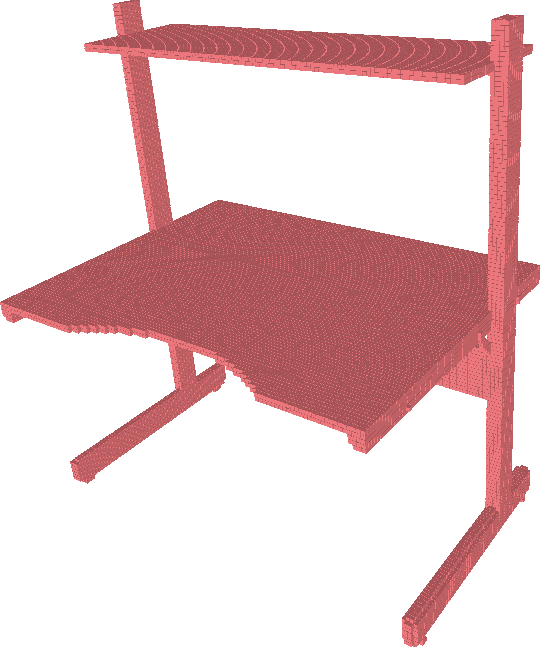}
&\raisebox{.4\height}{\includegraphics[width=0.048\textwidth]{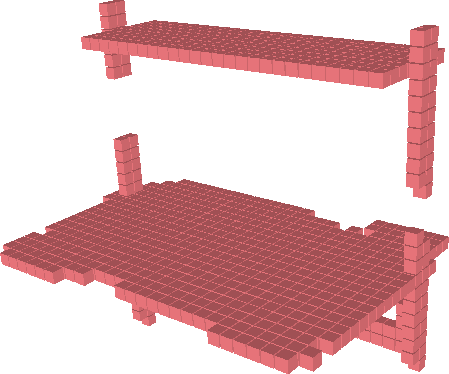}}
&\includegraphics[width=0.048\textwidth]{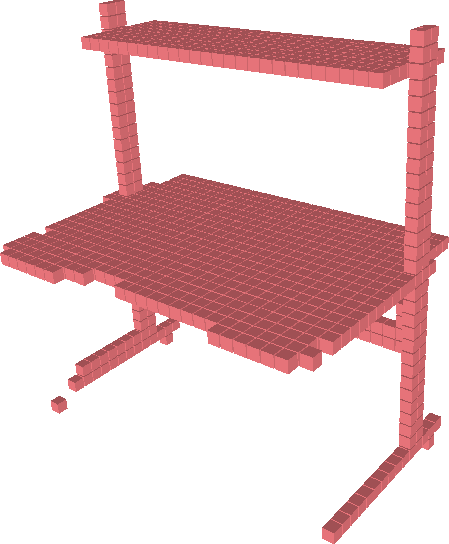}
&\raisebox{.2\height}{\includegraphics[width=0.048\textwidth]{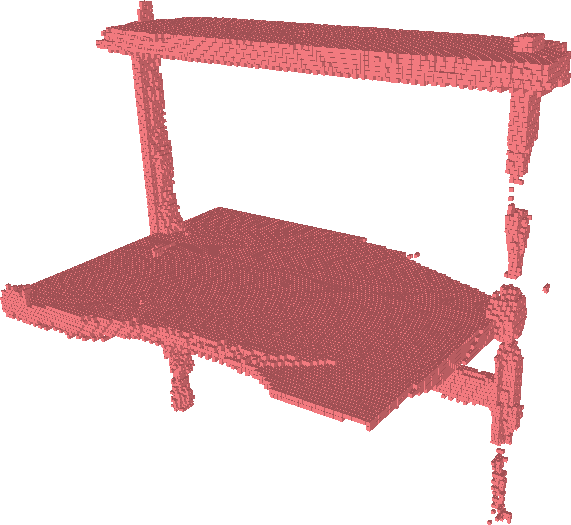}}
&\includegraphics[width=0.048\textwidth]{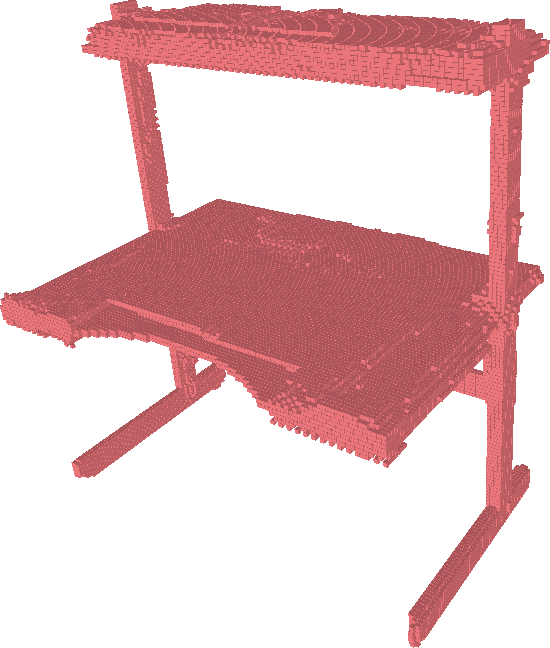}
&\raisebox{.25\height}{\includegraphics[width=0.03\textwidth]{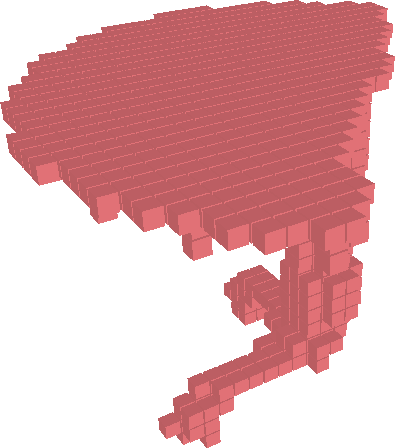}}
&\includegraphics[width=0.048\textwidth]{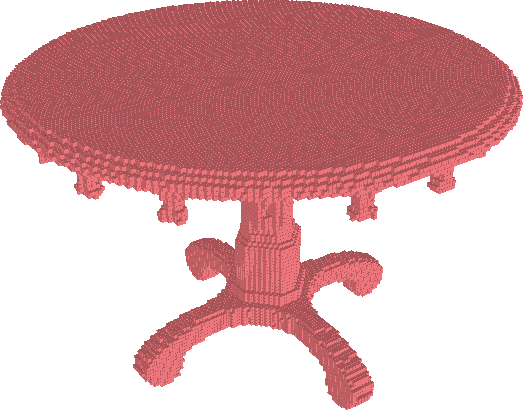}
&\raisebox{.35\height}{\includegraphics[width=0.048\textwidth]{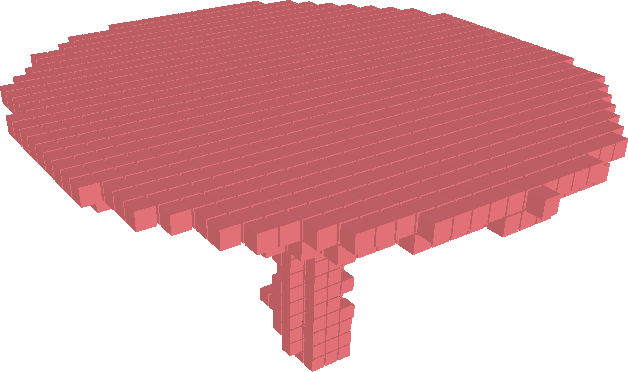}}
&\includegraphics[width=0.048\textwidth]{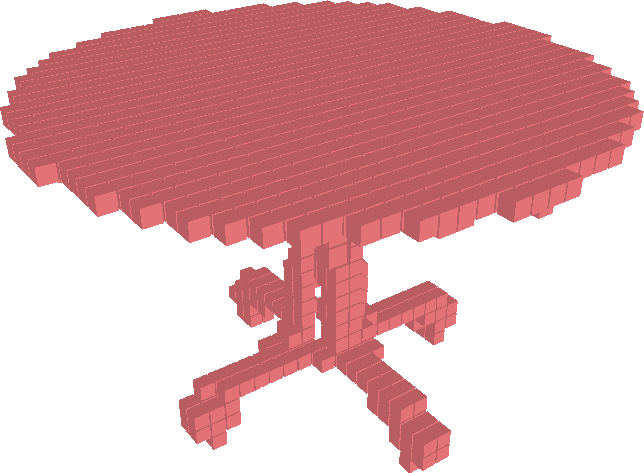}
&\includegraphics[width=0.048\textwidth]{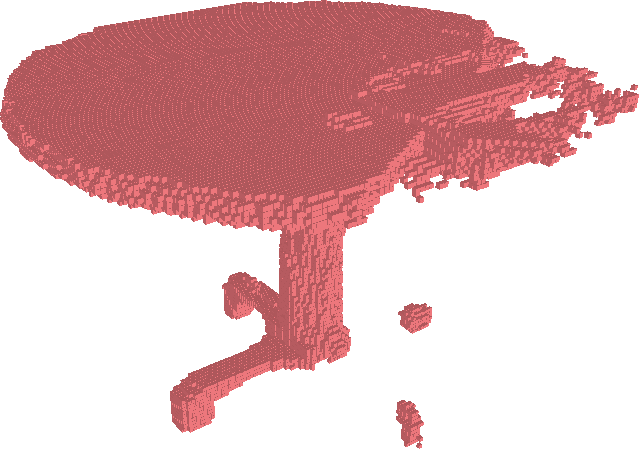}
&\includegraphics[width=0.048\textwidth]{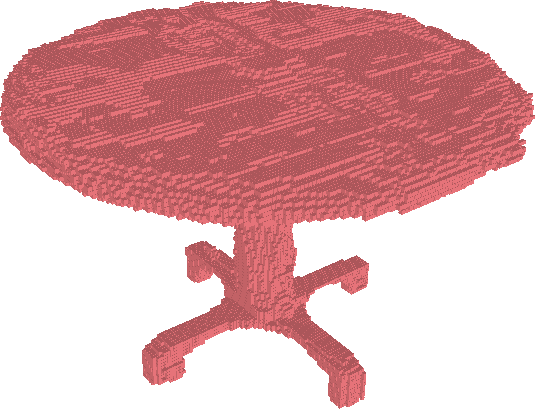}
\\

\scriptsize Input &\scriptsize  Ground Truth & \scriptsize VConv-DAE & \scriptsize 3D-ED-GAN & \scriptsize LCRN & \scriptsize Hybrid&
\scriptsize Input &\scriptsize  Ground Truth & \scriptsize VConv-DAE & \scriptsize 3D-ED-GAN & \scriptsize LCRN & \scriptsize Hybrid&
\scriptsize Input &\scriptsize  Ground Truth & \scriptsize VConv-DAE & \scriptsize 3D-ED-GAN & \scriptsize LCRN & \scriptsize Hybrid

\end{tabular}
\caption{Shape completion examples on ShapeNet testing points with simulated 3D scanner noise.}
\label{fig:completion}
\end{figure*}

We then evaluate our network on completing shapes for simulated scanned objects. 3D scanners such as Kinect can only capture object geometry from a single view at one time. In this experiment, we simulate these 3D scanners by scanning objects in the ShapeNet dataset from a single view and evaluate the reconstruction performance of our method from these scanned incomplete data. This is a challenging task since the recovered region must contain semantically correct content. Completion results can be found in  Figure~\ref{fig:completion}. Quantitative comparison results are shown in Table~\ref{table:completionhigh}. 

As illustrated in Figure~\ref{fig:completion} and Table~\ref{table:completionhigh}, our model performs better than 3D-ED-GAN, VConv-DAE and LRCN. For VConv-DAE, small or thin components of objects, such as the pole of a lamp tend to be filtered out even though these parts exist in the input volume. With the help of the generative adversarial model, our model is able to produce reasonable predictions for the large missing areas that are consistent with the data distribution. The superior performance of 3D-ED-GAN over VConv-DAE demonstrates our model benefits from the generative adversarial structure. Moreover, by comparing the results of 3D-ED-GAN and the hybrid network, we can see the capability of LRCN to recover local geometry. LRCN alone has difficulty capturing global context structure of 3D shapes. By combining 3D-ED-GAN and LRCN, our hybrid network is able to predict global structure as well as local fine-grained details.

Overall, our hybrid network performs best by leveraging 3D-ED-GAN's ability to produce plausible predictions and LRCN's power to recover local geometry.

\begin{table}
\begin{center}
\setlength{\tabcolsep}{0.1em}
\begin{tabular}{lc}
\Xhline{3\arrayrulewidth}
Methods&Reconstruction Error\\
\hline
VConv-DAE~\cite{vconvdae} & 7.48\% \\
3D-ED-GAN & 6.55\% \\
LRCN & 7.08\%\\
Hybrid&\textbf{4.74\%}\\
\Xhline{3\arrayrulewidth}
\end{tabular}
\end{center}
\caption{Quantitative shape completion results on ShapeNet with simulated 3D scanner noise.}
\label{table:completionhigh}
\end{table}

\subsection{Feature Learning}
\label{sec:featurelearning}
\subsubsection{3D object classification}
We now evaluate the transferability of unsupervised learned features obtained from inpainting to object classification. We use the popular benchmark ModelNet10 and ModelNet40, which are both subsets of the ModelNet dataset~\cite{modelnet}. Both ModelNet10 and ModelNet40 are split into mutually exclusive training and testing sets. We conduct three experiments.
\begin{enumerate}
\item Our-FT: We train 3D-ED-GAN as the first training stage stated in Section~\ref{sec:training} on all samples of ShapeNet dataset as pre-training and treat the encoder component (with a softmax layer added on top of $z$ as a loss layer) as our classifier. We fine-tune this CNN classifier on ModelNet10 and ModelNet40. 
\item RandomInit: We directly train the classifier mentioned in Our-FT with random initialization on ModelNet10 and ModelNet40.
\item Our-SVM: We generate $z$ (of dimension $16384$) with the trained 3D-ED-GAN in Section~\ref{sec:training} for samples on ModelNet10 and ModelNet40 and train a linear SVM classifier with $z$ as the feature vector.
\end{enumerate}
We also compare our algorithm with the state-of-the-art methods~\cite{3dgan,vconvdae,tl,generativediscriminative,sumultiview,qimultiview,shapenet}. VRN~\cite{generativediscriminative}, MVCNN~\cite{sumultiview}, MVCNN-Multi~\cite{qimultiview} are designed for object classification. 3DGAN~\cite{3dgan}, TL-network~\cite{tl}, and VConv-DAE-US~\cite{vconvdae} learned a feature representation for 3D objects, and trained a linear SVM as classifier for this task. VConv-DAE~\cite{vconvdae} and VRN~\cite{generativediscriminative} adopted a VAE architecture with pre-training. We report the testing accuracy in Table~\ref{table:classification}.

\begin{table}
\begin{center}
\setlength{\tabcolsep}{0.1em}
\begin{tabular}{llcc}
\Xhline{3\arrayrulewidth}
Methods&ModelNet40&ModelNet10\\
\hline
RandomInit&86.1\%&90.5\%\\
Ours-FT&\textbf{87.3\%}&\textbf{92.6\%}\\
\hline
3DGAN~\cite{3dgan}&83.3\%&\textbf{91.0\%}\\
TL-network~\cite{tl}&74.4\%&-\\
VConv-DAE-US~\cite{vconvdae}&75.5\%&80.5\%\\
Ours-SVM&\textbf{84.3\%}&89.2\%\\
\hline
3DShapeNet~\cite{shapenet}&77.0\%&83.5\%\\
VConv-DAE~\cite{vconvdae}&79.8\%&84.1\%\\
VRN~\cite{generativediscriminative}&91.3\%&\textbf{93.6\%}\\
MVCNN~\cite{sumultiview}&90.1\%&-\\
MVCNN-Multi\cite{qimultiview}&\textbf{91.4\%}&-\\
\Xhline{3\arrayrulewidth}
\end{tabular}
\end{center}
\caption{Classification Results on ModelNet Dataset.}
\label{table:classification}
\end{table}

Although our framework is not designed for object recognition, our results with 3D-ED-GAN pre-training is competitive with existing methods including models designed for recognition~\cite{generativediscriminative,sumultiview}. By comparing RandomInit and Ours-FT, we can see unsupervised 3D-ED-GAN pre-training is able to guide the CNN classifier to capture the rough geometric structure of 3D objects. The superior performance of Our-SVM training over other vector representation methods~\cite{tl,3dgan,vconvdae} demonstrate the effectiveness of our method as a feature learning architecture.


\subsubsection{Shape Arithmetic}

\begin{figure}[!tb]
\centering

    \setlength{\tabcolsep}{0.1em}
\newcolumntype{C}{>{\centering\arraybackslash}p{3.53em}}
\renewcommand{\arraystretch}{0} 
    \begin{tabular}[ht]{cCCCCC}
\includegraphics[width=0.035\textwidth]{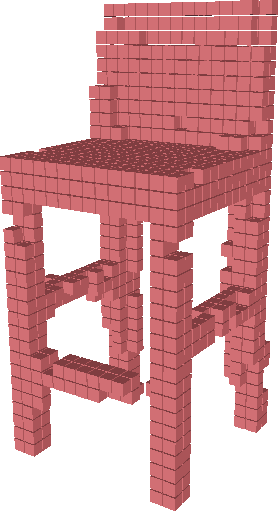}
&\includegraphics[width=0.035\textwidth]{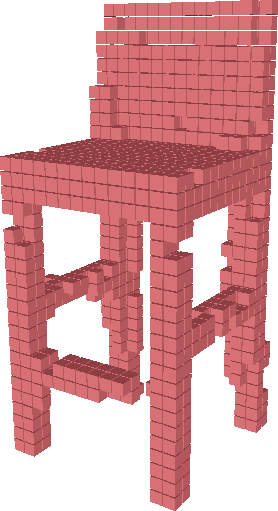}
&\includegraphics[width=0.06\textwidth]{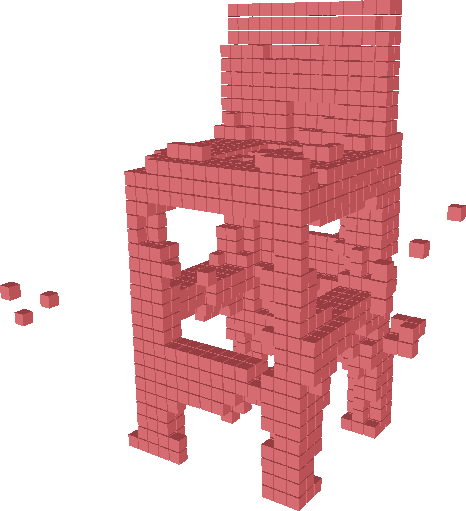}
&\includegraphics[width=0.06\textwidth]{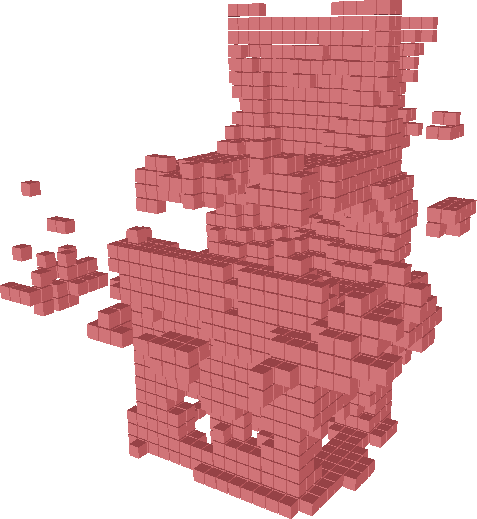}
&\includegraphics[width=0.06\textwidth]{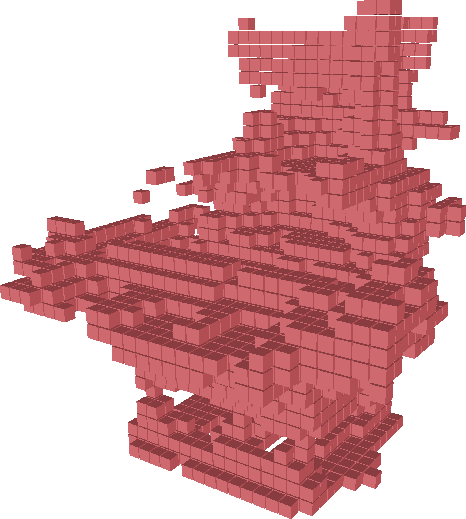}
&\includegraphics[width=0.06\textwidth]{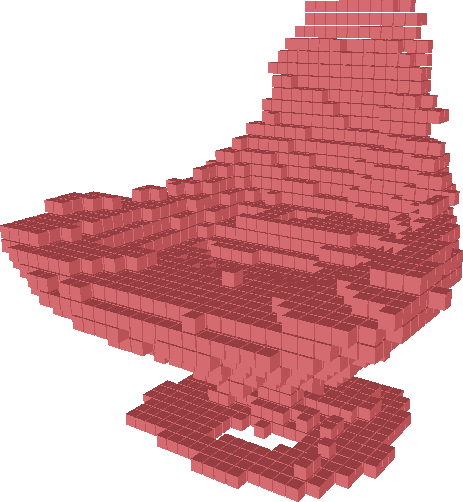}
\\
\raisebox{.45\height}{\includegraphics[width=0.07\textwidth]{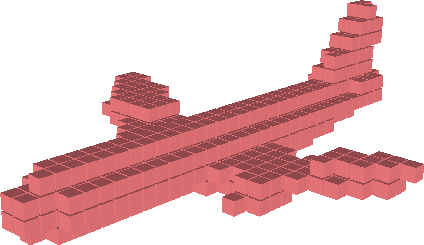}}
&\includegraphics[width=0.06\textwidth]{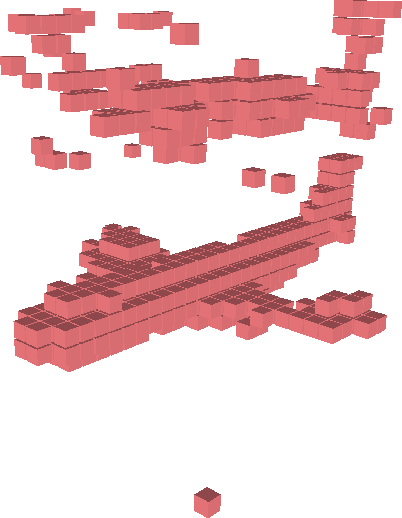}
&\includegraphics[width=0.06\textwidth]{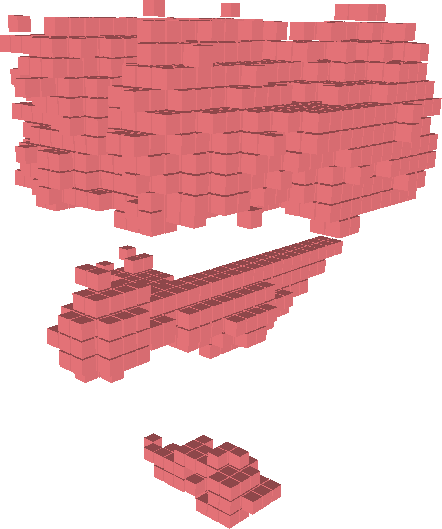}
&\includegraphics[width=0.06\textwidth]{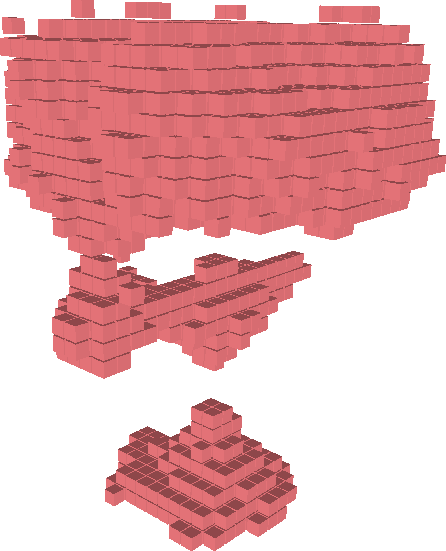}
&\includegraphics[width=0.06\textwidth]{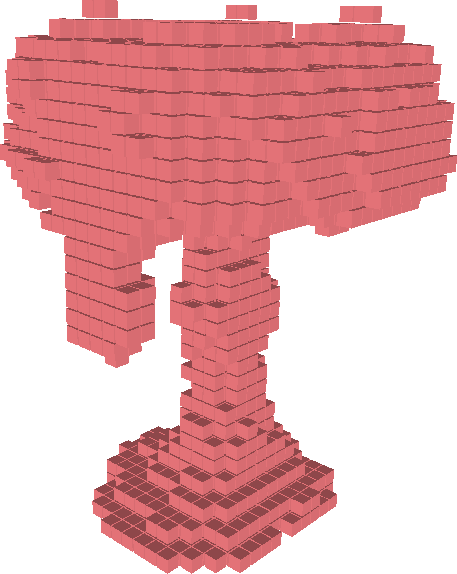}
&\includegraphics[width=0.06\textwidth]{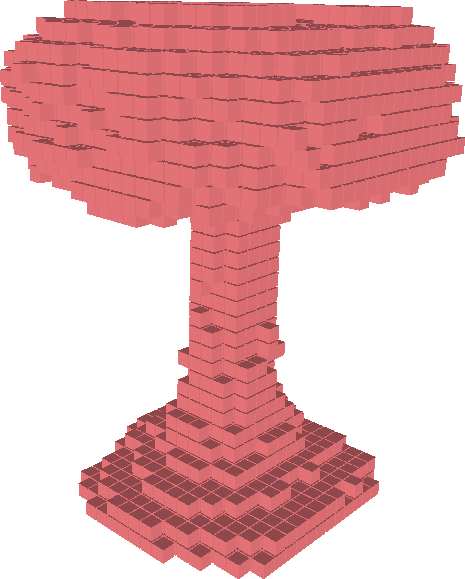}
\\
\includegraphics[width=0.03\textwidth]{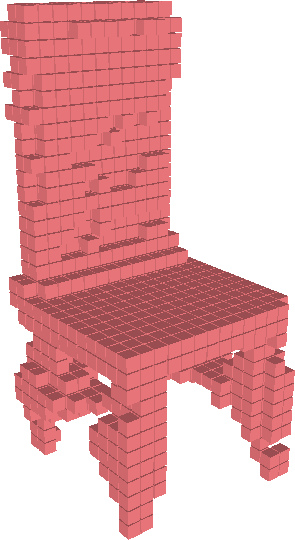}
&\includegraphics[width=0.06\textwidth]{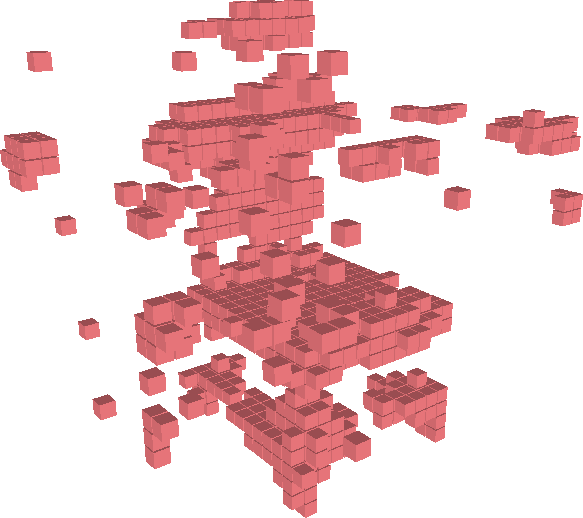}
&\includegraphics[width=0.06\textwidth]{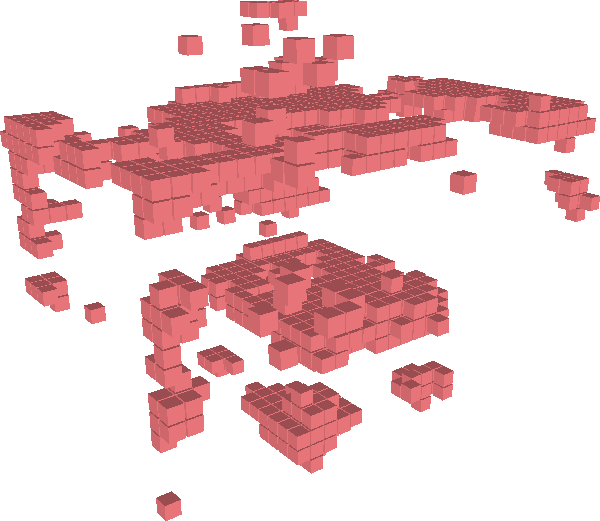}
&\includegraphics[width=0.06\textwidth]{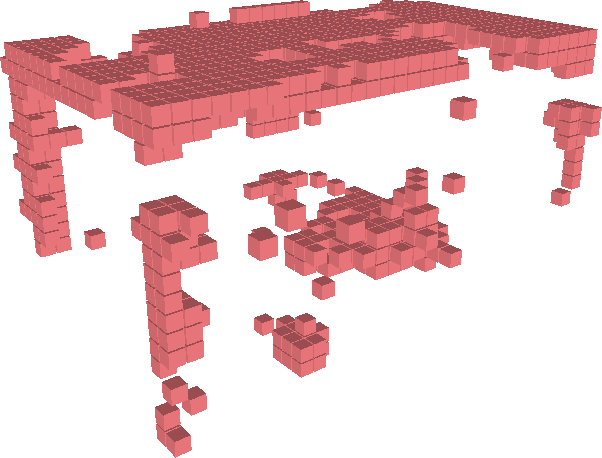}
&\includegraphics[width=0.06\textwidth]{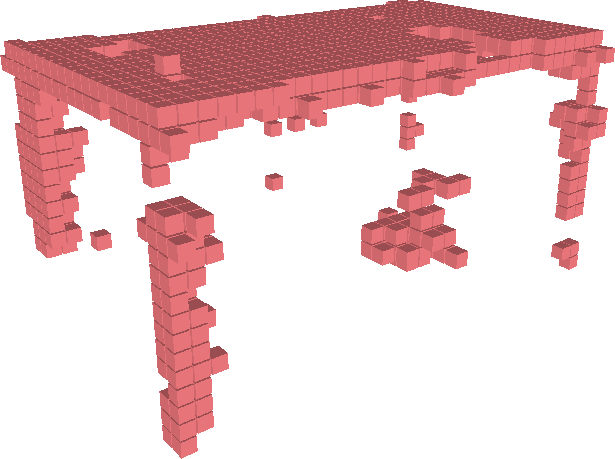}
&\includegraphics[width=0.06\textwidth]{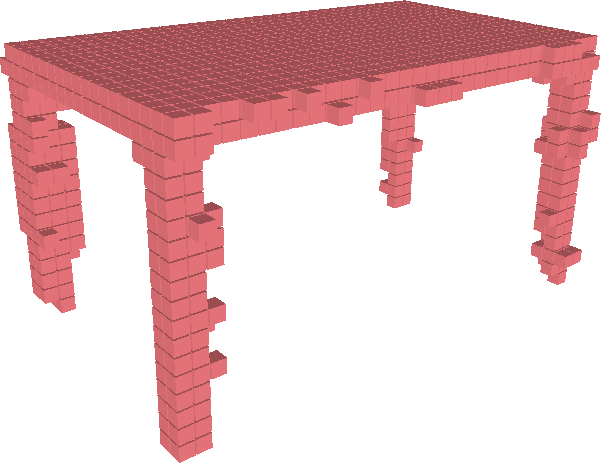}
\\
    \end{tabular}
    \caption{Shape interpolation results.}
\label{fig:interpolation}
\end{figure}

Previous works in embedding representation learning~\cite{3dgan,tl} have shown the phenomena of the capability of shape transformation by performing arithmetic on the latent vectors. Our 3D-ED-GAN also learns a latent vector $z$. To this end, we randomly chose two different instances and fed it into the encoder to produce two encoded vectors $z'$ and $z''$ and feed the interpolated vector $z'''=\gamma z'+(1-\gamma)z''$ ($0<\gamma<1$) to the decoder to produce volumes. The results for the interpolation are shown in Figure~\ref{fig:interpolation}. We observe smooth transitions in the generated object domain with gradually increasing $\gamma$.

\section{Conclusion and Future Work}

In this paper, we present a convolutional encoder-decoder generative adversarial network to inpaint corrupted 3D objects. A long-term recurrent convolutional network is further introduced, where the 3D volume is treated as a sequence of 2D images, to save GPU memory and complete high-resolution 3D volumetric data. Experimental results on both real-world and synthetic scans show the effectiveness of our method.

Since our model is easy to fit into GPU memory compared with other 3D CNN methods~\cite{vconvdae,ssn}. A potential direction is to complete more complex 3D structures, such as indoor scenes~\cite{ssn,dai2017scannet}, with much higher resolutions. Another interesting future avenue is to utilize our model on other 3D representations like 3D mesh, distance field etc.

{\small
\bibliographystyle{ieee}
\bibliography{egbib}

\begin{thebibliography}{10}\itemsep=-1pt

\bibitem{tensorflow}
M.~Abadi, A.~Agarwal, P.~Barham, E.~Brevdo, Z.~Chen, C.~Citro, G.~S. Corrado,
  A.~Davis, J.~Dean, M.~Devin, et~al.
\newblock Tensorflow: Large-scale machine learning on heterogeneous systems,
  2015.
\newblock {\em Software available from tensorflow. org}, 1, 2015.

\bibitem{marr}
A.~Bansal, B.~Russell, and A.~Gupta.
\newblock Marr {R}evisited: 2{D}-3{D} model alignment via surface normal
  prediction.
\newblock In {\em CVPR}, 2016.

\bibitem{generativediscriminative}
A.~Brock, T.~Lim, J.~Ritchie, and N.~Weston.
\newblock Generative and discriminative voxel modeling with convolutional
  neural networks.
\newblock {\em arXiv preprint arXiv:1608.04236}, 2016.

\bibitem{scenelabelinglstm}
W.~Byeon, T.~M. Breuel, F.~Raue, and M.~Liwicki.
\newblock Scene labeling with lstm recurrent neural networks.
\newblock In {\em CVPR}, June 2015.

\bibitem{shapenet}
A.~X. Chang, T.~Funkhouser, L.~Guibas, P.~Hanrahan, Q.~Huang, Z.~Li,
  S.~Savarese, M.~Savva, S.~Song, H.~Su, J.~Xiao, L.~Yi, and F.~Yu.
\newblock {ShapeNet: An Information-Rich 3D Model Repository}.
\newblock Technical Report arXiv:1512.03012, Stanford University --- Princeton
  University --- Toyota Technological Institute at Chicago, 2015.

\bibitem{r2n2}
C.~B. Choy, D.~Xu, J.~Gwak, K.~Chen, and S.~Savarese.
\newblock 3d-r2n2: A unified approach for single and multi-view 3d object
  reconstruction.
\newblock In {\em ECCV}, 2016.

\bibitem{dai2017scannet}
A.~Dai, A.~X. Chang, M.~Savva, M.~Halber, T.~Funkhouser, and M.~Nie{\ss}ner.
\newblock Scannet: Richly-annotated 3d reconstructions of indoor scenes.
\newblock {\em arXiv preprint arXiv:1702.04405}, 2017.

\bibitem{3depn}
A.~Dai, C.~R. Qi, and M.~Nie{\ss}ner.
\newblock Shape completion using 3d-encoder-predictor cnns and shape synthesis.
\newblock 2017.

\bibitem{lapgan}
E.~L. Denton, S.~Chintala, R.~Fergus, et~al.
\newblock Deep generative image models using a laplacian pyramid of adversarial
  networks.
\newblock In {\em NIPS}, 2015.

\bibitem{LRCNvideo}
J.~Donahue, L.~Anne~Hendricks, S.~Guadarrama, M.~Rohrbach, S.~Venugopalan,
  K.~Saenko, and T.~Darrell.
\newblock Long-term recurrent convolutional networks for visual recognition and
  description.
\newblock In {\em CVPR}, 2015.

\bibitem{tl}
R.~Girdhar, D.~F. Fouhey, M.~Rodriguez, and A.~Gupta.
\newblock Learning a predictable and generative vector representation for
  objects.
\newblock In {\em ECCV}, 2016.

\bibitem{gan}
I.~Goodfellow, J.~Pouget-Abadie, M.~Mirza, B.~Xu, D.~Warde-Farley, S.~Ozair,
  A.~Courville, and Y.~Bengio.
\newblock Generative adversarial nets.
\newblock In {\em NIPS}, 2014.

\bibitem{prelu}
K.~He, X.~Zhang, S.~Ren, and J.~Sun.
\newblock Delving deep into rectifiers: Surpassing human-level performance on
  imagenet classification.
\newblock In {\em ICCV}, 2015.

\bibitem{bn}
S.~Ioffe and C.~Szegedy.
\newblock Batch normalization: Accelerating deep network training by reducing
  internal covariate shift.
\newblock In {\em ICML}, 2015.

\bibitem{recurrenttranslation}
N.~Kalchbrenner and P.~Blunsom.
\newblock Recurrent continuous translation models.
\newblock In {\em EMNLP}, 2013.

\bibitem{adam}
D.~Kingma and J.~Ba.
\newblock Adam: A method for stochastic optimization.
\newblock In {\em ICLR}, 2015.

\bibitem{VAE}
D.~P. Kingma and M.~Welling.
\newblock Auto-encoding variational bayes.
\newblock In {\em ICLR}, 2014.

\bibitem{contextencoder}
D.~Pathak, P.~Kr\"ahenb\"uhl, J.~Donahue, T.~Darrell, and A.~Efros.
\newblock Context encoders: Feature learning by inpainting.
\newblock In {\em CVPR}, 2016.

\bibitem{qimultiview}
C.~R. Qi, H.~Su, M.~Nie{\ss}ner, A.~Dai, M.~Yan, and L.~Guibas.
\newblock Volumetric and multi-view cnns for object classification on 3d data.
\newblock In {\em CVPR}, 2016.

\bibitem{dcgan}
A.~Radford, L.~Metz, and S.~Chintala.
\newblock Unsupervised representation learning with deep convolutional
  generative adversarial networks.
\newblock In {\em ICLR}, 2016.

\bibitem{vconvdae}
A.~Sharma, O.~Grau, and M.~Fritz.
\newblock Vconv-dae: Deep volumetric shape learning without object labels.
\newblock {\em arXiv preprint arXiv:1604.03755}, 2016.

\bibitem{ssn}
S.~Song, F.~Yu, A.~Zeng, A.~X. Chang, M.~Savva, and T.~Funkhouser.
\newblock Semantic scene completion from a single depth image.
\newblock In {\em CVPR}, 2017.

\bibitem{sumultiview}
H.~Su, S.~Maji, E.~Kalogerakis, and E.~Learned-Miller.
\newblock Multi-view convolutional neural networks for 3d shape recognition.
\newblock In {\em ICCV}, 2015.

\bibitem{pixelrnn}
A.~van~den Oord, N.~Kalchbrenner, and K.~Kavukcuoglu.
\newblock Pixel recurrent neural networks.
\newblock In {\em ICML}, 2016.

\bibitem{reseg}
F.~Visin, M.~Ciccone, A.~Romero, K.~Kastner, K.~Cho, Y.~Bengio, M.~Matteucci,
  and A.~Courville.
\newblock Reseg: A recurrent neural network-based model for semantic
  segmentation.
\newblock In {\em CVPR Workshops}, June 2016.

\bibitem{3dgan}
J.~Wu, C.~Zhang, T.~Xue, W.~T. Freeman, and J.~B. Tenenbaum.
\newblock Learning a probabilistic latent space of object shapes via 3d
  generative-adversarial modeling.
\newblock In {\em NIPS}, 2016.

\bibitem{modelnet}
Z.~Wu, S.~Song, A.~Khosla, F.~Yu, L.~Zhang, X.~Tang, and J.~Xiao.
\newblock 3d shapenets: A deep representation for volumetric shapes.
\newblock In {\em CVPR}, 2015.

\bibitem{honglak}
X.~Yan, J.~Yang, E.~Yumer, Y.~Guo, and H.~Lee.
\newblock Perspective transformer nets: Learning single-view 3d object
  reconstruction without 3d supervision.
\newblock In {\em NIPS}, 2016.

\bibitem{neuralpatch}
C.~Yang, X.~Lu, Z.~Lin, E.~Shechtman, O.~Wang, and H.~Li.
\newblock High-resolution image inpainting using multi-scale neural patch
  synthesis.
\newblock In {\em CVPR}, 2017.

\bibitem{lstm}
W.~Zaremba, I.~Sutskever, and O.~Vinyals.
\newblock Recurrent neural network regularization.
\newblock {\em arXiv preprint arXiv:1409.2329}, 2014.

\end{thebibliography}
}

\end{document}